\documentclass[twoside,english,3p]{elsarticle}
\usepackage[T1]{fontenc}
\usepackage{geometry}
\usepackage{amssymb}
\usepackage{amsmath}

\usepackage{amsthm}
\usepackage{stmaryrd}
\usepackage{graphicx}
\usepackage{booktabs}
\usepackage{multirow}
\usepackage{makecell}
\usepackage{xurl}
\usepackage{esint}
\usepackage{algorithm}
\usepackage{algpseudocode}
\usepackage{rotating}
\usepackage{adjustbox}
\usepackage{nomencl}
\usepackage{multicol}
\makenomenclature
\makeatletter
\theoremstyle{plain}

\theoremstyle{boldremark} % remark bold

\ifx\proof\undefined

\providecommand{\proofname}{Proof}
\fi

\renewenvironment{proof}[1][\proofname]{%
\par\pushQED{\qed}\normalfont%
\topsep6\p@\@plus6\p@\relax
\trivlist\item[\hskip\labelsep\bfseries#1\@addpunct{.}]%
\ignorespaces
}{%
\popQED\endtrivlist\@endpefalse
}

\journal{Elsevier}

\usepackage{hyperref}
\hypersetup{colorlinks = true, allcolors = blue}

\usepackage[nameinlink]{cleveref}

\crefname{figure}{Fig.}{Figs.}
\crefformat{equation}{Eq.~#2(#1)#3}
\crefformat{section}{Section~#2#1#3}
\AtBeginDocument{%
\let\citet\cite
}

\usepackage[labelfont=bf]{caption}
\@ifundefined{showcaptionsetup}{}{%
\PassOptionsToPackage{caption=false}{subfig}}
\usepackage{subfig}
\makeatother

\usepackage{babel}
\providecommand{\remarkname}{Remark}
\providecommand{\theoremname}{Theorem}

\begin{document}

\begin{frontmatter}{}
	
	\title{Kolmogorov–Arnold-Informed neural network: A physics-informed deep learning framework for solving forward and inverse problems based on Kolmogorov–Arnold Networks}

	\author[rvt,rvt3]{Yizheng Wang}
	
	\ead{wang-yz19@tsinghua.org.cn}
	
	\author[rvt2]{Jia Sun}
	\author[rvt,rvt4,rvt5]{Jinshuai Bai}

	\author[rvt3]{Cosmin Anitescu}
	
	\author[rvt6]{Mohammad Sadegh Eshaghi}

	\author[rvt6]{Xiaoying Zhuang}
	
	\author[rvt3]{Timon Rabczuk}

	\author[rvt]{Yinghua Liu\corref{cor1}}

	\ead{yhliu@mail.tsinghua.edu.cn}
	\cortext[cor1]{Corresponding author}
	\address[rvt]{Department of Engineering Mechanics, Tsinghua University, Beijing 100084, China}

	\address[rvt3]{Institute of Structural Mechanics, Bauhaus-Universit\"{a}t Weimar, Marienstr. 15, D-99423 Weimar, Germany}	

	\address[rvt6]{ Institute of Photonics, Department of Mathematics and Physics, Leibniz University Hannover, Germany}

	\address[rvt2]{Drilling Mechanical Department, CNPC Engineering Technology RD Company Limited, Beijing 102206, China}

	\address[rvt4]{School of Mechanical, Medical and Process Engineering, Queensland University of Technology, Brisbane, QLD 4000, Australia}

	\address[rvt5]{ARC Industrial Transformation Training Centre—Joint Biomechanics, Queensland University of Technology, Brisbane, QLD 4000, Australia}

\begin{abstract}

AI for partial differential equations (PDEs) has garnered significant attention, particularly with the emergence of Physics-informed neural networks (PINNs). The recent advent of Kolmogorov-Arnold Network (KAN) indicates that there is potential to revisit and enhance the previously MLP-based PINNs. Compared to MLPs, KANs offer interpretability and require fewer parameters. PDEs can be described in various forms, such as strong form, energy form, and inverse form. While mathematically equivalent, these forms are not computationally equivalent, making the exploration of different PDE formulations significant in computational physics. Thus, we propose different PDE forms based on KAN instead of MLP, termed Kolmogorov-Arnold-Informed Neural Network (KINN) for solving forward and inverse problems. We systematically compare MLP and KAN in various numerical examples of PDEs, including multi-scale, singularity, stress concentration, nonlinear hyperelasticity, heterogeneous, and complex geometry problems. Our results demonstrate that KINN significantly outperforms MLP regarding accuracy and convergence speed for numerous PDEs in computational solid mechanics, except for the complex geometry problem. This highlights KINN's potential for more efficient and accurate PDE solutions in AI for PDEs.

\end{abstract}

\printnomenclature

%Graphical abstract
\begin{graphicalabstract}
	\includegraphics[scale=0.45]{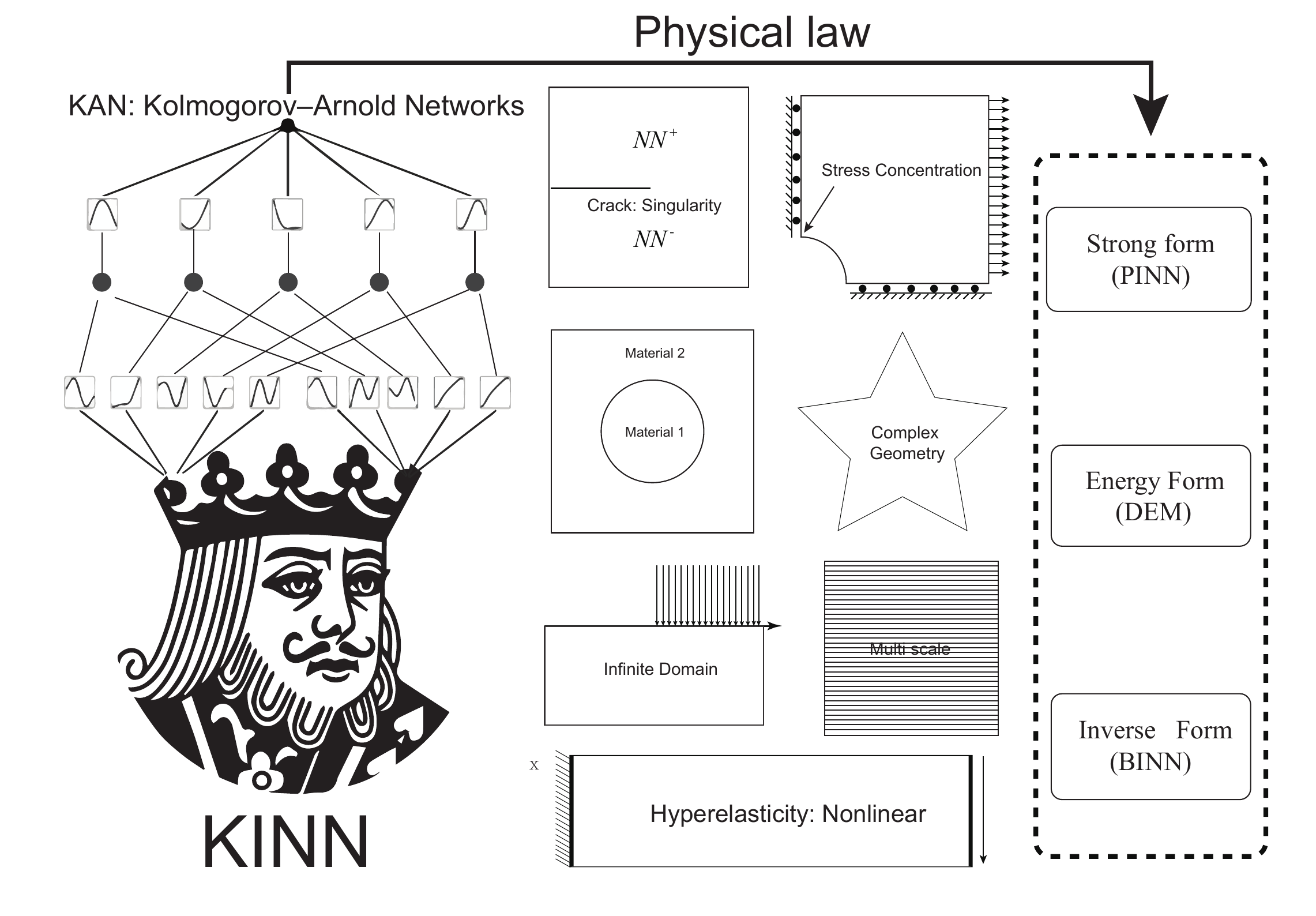}
\end{graphicalabstract}


\end{frontmatter}{}

\section*{Nomenclature}
\begin{multicols}{2}
	\begin{description}
		\item[BINN]  Boundary-integral type neural networks (inverse form of PDEs)
		\item[BINN\_MLP] Use MLP in BINN
		\item[CENN] Deep Energy Method with subdomains
		\item[CPINNs] The strong form of PINNs with subdomains
		\item[CPINNs\_MLP] Use MLP in CPINNs
		\item[DEM] Deep Energy Method (energy form of PDEs)
		\item[DEM\_MLP] Use MLP in DEM
		\item[FEM] Finite Element Method
		\item[FNO] Fourier Neural Operator
		\item[IGA] Isogeometric Analysis
		\item[KAN] Kolmogorov-Arnold Network
		\item[KINN\_BINN] Use KAN in BINN
		\item[KINN\_CPINNs] Use KAN in CPINNs
		\item[KINN\_DEM] Use KAN in DEM
		\item[KINN\_PINNs] Use KAN in PINNs
		\item[MLP] Fully connected neural network
		\item[Mont] Monte Carlo integration
		\item[NURBS] Non-Uniform Rational B-Splines
		\item[NTK] Neural Tangent Kernel
		\item[PDEs] Partial differential equations
		\item[PINO] Physics-Informed Neural Operator
		\item[PINNs]  Physics-Informed Neural Networks (strong form of PDEs)
		\item[PINNs\_MLP] Use MLP in PINNs
		\item[RBF] Radial Basis Function
		\item[Simp] Simpson’s rule
		\item[Trap] Trapezoidal rule
		\item[\_penalty] Use penalty function to satisfy boundary conditions
		\item[\_rbf] Use RBF as distance function to satisfy boundary conditions in advance
	\end{description}
\end{multicols}

\section{Introduction}
A multitude of physical phenomena rely on partial differential equations (PDEs) for modeling. Thus, solving PDEs is essential for gaining an understanding of the behavior of both natural and engineered systems. However, once the boundary and initial conditions become complex, the exact solution of PDEs is often difficult to obtain \cite{loss_is_minimum_potential_energy}. At this point, various numerical methods are employed to solve PDEs for achieving approximate solutions, such as commonly used finite element methods \cite{finite_element_book,hughes2012finite,bathe2006finite,reddy2019introduction}, mesh-free methods \cite{zhang2016material,liu2003mesh,rabczuk2004cracking,rabczuk2007three,rabczuk2019extended,nguyen2008meshless}, finite difference methods \cite{leveque2007finitedifferentialmethod}, finite volume methods \cite{darwish2016finitevolumemethod}, and boundary element methods \cite{brebbia2012boundary}.

Recently, AI for PDEs, an important direction of AI for science, refers to a class of algorithms that use deep learning to solve PDEs. There are three important approaches to AI for PDEs: Physics-Informed Neural Networks (PINNs) \citet{PINN_original_paper}, operator learning \citet{DeepOnet}, and Physics-Informed Neural Operator (PINO) \citet{li2024physics}. We will review these three approaches.
The first approach in AI for PDEs is PINNs \citet{PINN_original_paper}. Since the same PDE can have different forms of expression, each with varying accuracy and efficiency, different forms of PINNs have been developed based on these expressions. These forms include PINNs in strong form \citet{PINN_original_paper}, PINNs in weak form (hp-VPINNs) \citet{hp-VPINN}, PINNs in energy form (DEM: Deep Energy Method) \citet{loss_is_minimum_potential_energy}, and PINNs in inverse form (BINN: Boundary Element Method) \citet{sun2023binn}. It is important to emphasize that although these different mathematical descriptions of PDEs are equivalent, they are not computationally equivalent. Therefore, exploring different PDE descriptions is significant in computational physics.
The second approach in AI for PDEs is operator learning, represented by DeepONet \citet{DeepOnet} and FNO (Fourier Neural Operator) \citet{li2020fourier}. Initially, proposed operator learning methods are purely data-driven, making them well-suited for problems with big data \cite{lu2022comprehensive}, especially weather prediction \cite{bi2023accurate}. Notably, unlike the initially proposed PINNs, which can only solve specific problems and require re-solving when boundary conditions, geometries, or materials change, operator learning methods learn a series of mappings, i.e., a family of PDEs \cite{li2020fourier}. 
Operator learning can quickly provide solutions even when the above conditions change \citet{kovachki2023neural}.
However, recent theories suggest that some operator learning algorithms exhibit aliasing errors between the discrete representations and the continuous operators \cite{bartolucci2024representation}.
The third approach is PINO \citet{li2024physics}, which combines physical equations with operator learning. By incorporating physical equations during the training of operator learning, traditional operator learning can achieve higher accuracy \citep{goswami2022physics,wang2021learning}. Additionally, PINO can utilize operator learning to first obtain a good approximate solution and then refine it using PDEs, greatly accelerating the computation of PDEs \citep{wang2023dcm}. So why AI for PDEs can be successful?

AI for PDEs has succeeded mainly due to the neural network's strong approximation capability. The universal approximation theorem shows that a fully connected neural network (MLP), given enough hidden neurons and appropriate weight configurations, can approximate any continuous function with arbitrary precision \cite{super_approximation,hornik1989multilayer}. The universal approximation theorem is the cornerstone of MLP's application in AI for PDEs. However, another important theory, the Kolmogorov-Arnold representation theorem, also approximates any multivariate continuous function \cite{kolmogorov1961representation}. This theorem tells us that any multivariate function can be decomposed into a finite combination of univariate functions \cite{braun2009constructive}. Based on the Kolmogorov-Arnold representation theorem, Hecht-Nielsen proposed the Kolmogorov network as approximate functions \cite{hecht1987kolmogorov}, but with only two layers. However, the nature of the univariate functions in the original Kolmogorov-Arnold representation theorem is often poor, making the Kolmogorov network with shallow layers nearly ineffective in practical applications \cite{poggio2020theoretical}.

Recently, Liu et al. proposed KAN (Kolmogorov-Arnold Network) \cite{liu2024kan}, which deepens the shallow Kolmogorov network to create univariate functions with good properties. KAN is very similar to MLP, with the main difference being that KAN's activation functions need to be learned. In the original KAN, B-splines are used for activation function construction due to their excellent fitting ability \cite{liu2024kan}. Later, some studies modified the B-splines used as activation functions in KAN, replacing B-splines with Chebyshev orthogonal polynomials \cite{ss2024chebyshev,shukla2024comprehensive}, radial basis functions \cite{li2024kolmogorov}, and wavelet transforms \cite{bozorgasl2024wav}. Function approximation theory in numerical analysis tells us that different weight functions form different orthogonal polynomials with various advantages, such as Legendre polynomials, Laguerre polynomials, Hermite orthogonal polynomials, and Chebyshev polynomials \cite{hildebrand1987introduction}. In theory, these orthogonal polynomials can all replace B-splines in KAN. Similarly, other traditional approximation functions can also replace B-splines, such as radial basis functions, wavelet transforms, and the renowned NURBS (Non-Uniform Rational B-Splines) in IGA (Isogeometric Analysis) \cite{hughes2005isogeometric}. Therefore, theoretically, almost all traditional approximation functions can replace B-splines in KAN, making this work very extensive. However, the core of KAN is the introduction of a framework for composite functions that can learn activation functions.

As a result, we propose KINN for solving forward and inverse problems, which is the KAN version of different forms of PDEs (strong form, energy form, and inverse form) in this work. Due to the extensive work on using different approximation functions instead of B-splines, we use the original B-spline version of KAN to directly compare it with MLP in different forms of PDEs. We systematically compare whether accuracy and efficiency are improved. Our motivation is simple: MLP has many parameters and lacks interpretability, with spectral bias problems \cite{wang2021learning}, making it less accurate and interpretable when used as the approximate function for different forms of PINNs. However, KAN has fewer parameters than MLP, and because KAN's activation functions are B-splines, KAN's function construction is more aligned with the essence of numerical algorithms for solving PDEs. Therefore, it makes sense to believe that combining KAN with various forms of PINNs to replace MLP will have better results.

The outline of the paper is as follows. \Cref{sec:Preparatory-knowledge} introduces the mathematical descriptions of different PDEs, divided into strong form, energy form, and inverse form. \Cref{sec:method} introduces the traditional KAN and our proposed KINN.  \Cref{sec:Result} is numerical experiments and divided into three parts:
\begin{enumerate}
\item Problems that MLP cannot solve but KAN can: multi-scale problems, i.e., high and low-frequency mixed problems.

\item Problems where KAN is more accurate than MLP: crack singularity, stress concentration (plate with a circular hole), nonlinear hyperelastic problems, and heterogeneous problems. Furthermore, for extreme heterogeneous inverse problems, KAN exhibits higher accuracy compared to MLP.

\item Problems where KAN performs worse than MLP: complex boundary problems.
\end{enumerate}
Finally, in Section \ref{sec:Conclusion}, we summarize the current advantages and limitations of KINN and provide suggestions for the future research direction of KAN in solving PDEs.

\section{Preparatory knowledge\label{sec:Preparatory-knowledge}}

In this section, we provide an overview of the different forms of AI for PDEs. Although the same PDEs
can have different formulations, they all aim to solve the same underlying PDEs. The reason for researching these
different formulations is that each form offers distinct advantages regarding computational efficiency
and accuracy. Therefore, we introduce several different approaches to solving PDEs, including the strong
form (PINNs: Physics-Informed Neural Networks), the energy form (DEM: Deep Energy Method), and the inverse
form (BINNs: Boundary-Integral Neural Networks).

\subsection{Introduction to the strong form of PDEs \label{subsec:Introduction_PINNs}}

We begin our discussion from the PDEs of boundary value problems, considering the following equations:
\begin{equation}
	\begin{cases}
		\boldsymbol{P}(\boldsymbol{u}(\boldsymbol{x}))=\boldsymbol{f}(\boldsymbol{x}) & \boldsymbol{x}\in\Omega\\
		\boldsymbol{B}(\boldsymbol{u}(\boldsymbol{x}))=\boldsymbol{g}(\boldsymbol{x}) & \boldsymbol{x}\in\Gamma
	\end{cases},\label{eq:original_form}
\end{equation}
where $\boldsymbol{P}$ and $\boldsymbol{B}$ (they could be nonlinear) are the domain operator and boundary operator of the differential
equations, respectively, and $\Omega$ and $\Gamma$ represent the domain and boundary, respectively.
We use the weighted residual method to transform these equations into their weighted residual form:
\begin{equation}
	\begin{cases}
		\int_{\Omega}[\boldsymbol{P}(\boldsymbol{u}(\boldsymbol{x}))-\boldsymbol{f}(\boldsymbol{x})]\cdot\boldsymbol{w}(\boldsymbol{x})d\Omega=0 & \boldsymbol{x}\in\Omega\\
		\int_{\Gamma}[\boldsymbol{B}(\boldsymbol{u}(\boldsymbol{x}))-\boldsymbol{g}(\boldsymbol{x})]\cdot\boldsymbol{w}(\boldsymbol{x})d\Gamma=0 & \boldsymbol{x}\in\Gamma
	\end{cases},\label{eq:original_form_weighted}
\end{equation}
where $\boldsymbol{w}(\boldsymbol{x})$ is the weight function. The equations in their original form
and weighted residual form are equivalent if $\boldsymbol{w}(\boldsymbol{x})$ is arbitrary. For numerical
convenience, we often predefine the form of $\boldsymbol{w}(\boldsymbol{x})$ and get the residual form
of the PDEs:

\begin{equation}
	\boldsymbol{w}(\boldsymbol{x})=\begin{cases}
		\boldsymbol{P}(\boldsymbol{u}(\boldsymbol{x}))-\boldsymbol{f}(\boldsymbol{x}) & \boldsymbol{x}\in\Omega\\
		\boldsymbol{B}(\boldsymbol{u}(\boldsymbol{x}))-\boldsymbol{g}(\boldsymbol{x}) & \boldsymbol{x}\in\Gamma
	\end{cases}.\label{eq:strong_form_weight}
\end{equation}
\Cref{eq:original_form_weighted} is transformed into the integration form:
\begin{equation}
	\begin{cases}
		\int_{\Omega}[\boldsymbol{P}(\boldsymbol{u}(\boldsymbol{x}))-\boldsymbol{f}(\boldsymbol{x})]\cdot[\boldsymbol{P}(\boldsymbol{u}(\boldsymbol{x}))-\boldsymbol{f}(\boldsymbol{x})]d\Omega=0 & \boldsymbol{x}\in\Omega\\
		\int_{\Gamma}[\boldsymbol{B}(\boldsymbol{u}(\boldsymbol{x}))-\boldsymbol{g}(\boldsymbol{x})]\cdot[\boldsymbol{B}(\boldsymbol{u}(\boldsymbol{x}))-\boldsymbol{g}(\boldsymbol{x})]d\Gamma=0 & \boldsymbol{x}\in\Gamma
	\end{cases}.\label{eq:original_form_delta}
\end{equation}
Next, we approximate these integrals in \Cref{eq:original_form_delta} numerically, leading to the strong
form of PINNs:

\begin{equation}
\mathcal{L}_{PINNs}=\frac{\lambda_{r}}{N_{r}}\Sigma_{i=1}^{N_{r}}|\boldsymbol{P}(\boldsymbol{u}(\boldsymbol{x}_{i};\boldsymbol{\theta}))-\boldsymbol{f}(\boldsymbol{x}_{i})|{}^{2}+\frac{\lambda_{b}}{N_{b}}\Sigma_{i=1}^{N_{b}}|\boldsymbol{B}(\boldsymbol{u}(\boldsymbol{x}_{i};\boldsymbol{\theta}))-\boldsymbol{g}(\boldsymbol{x}_{i})|{}^{2}.
\end{equation}
We optimize the above loss function to obtain the neural network approximation of the field variable
$\boldsymbol{u}(\boldsymbol{x};\boldsymbol{\theta})$:
\begin{equation}
\boldsymbol{u}(\boldsymbol{x};\boldsymbol{\theta})=\underset{\boldsymbol{\theta}}{\arg\min}\mathcal{L}_{PINNs}.
\end{equation}

Thus, mathematically, the strong form of PINNs \citep{PINN_original_paper} essentially involves choosing
the specific weight function $\boldsymbol{w}(\boldsymbol{x})$ as \Cref{eq:strong_form_weight} as the
residual form of the PDEs.

\subsection{Introduction to the energy form of PDEs\label{subsec:Introduction_DEM}}

In this chapter, we introduce the energy form of PDEs \citep{loss_is_minimum_potential_energy}.
We consider the weight function $\boldsymbol{w}(\boldsymbol{x})$ in \Cref{eq:original_form_weighted}
as $\delta\boldsymbol{u}$, which leads to the Garlerkin form.  \Cref{eq:original_form_weighted} can be written as:
\begin{align}
	\int_{\Omega}[\boldsymbol{P}(\boldsymbol{u}(\boldsymbol{x}))-\boldsymbol{f}(\boldsymbol{x})]\cdot\delta\boldsymbol{u}d\Omega=0 & ,\boldsymbol{x}\in\Omega.\label{eq:galerkin_form}
\end{align}
Boundary conditions have not been introduced, because we will introduce Neumann boundary conditions and eliminate Dirichlet boundary conditions in the subsequent Gaussian integration formula. For simplicity, we consider a specific Poisson equation to illustrate the energy form:
\begin{equation}
	\begin{cases}
		-\triangle(u(\boldsymbol{x}))=f(\boldsymbol{x}) & \boldsymbol{x}\in\Omega\\
		u(\boldsymbol{x})=\bar{u}(\boldsymbol{x}) & \boldsymbol{x}\in\Gamma^{u}\\
		\frac{\partial u(\boldsymbol{x})}{\partial n}=\bar{t}(\boldsymbol{x}) & \boldsymbol{x}\in\Gamma^{t}
	\end{cases},\label{eq:poisson_equation}
\end{equation}
where $\Gamma^{u}$ and $\Gamma^{t}$ are the Dirichlet and Neumann boundary conditions, respectively.
For the Poisson equation, the Garlerkin form of \Cref{eq:poisson_equation} can be expressed as:
\begin{align}
	\int_{\Omega}[-\triangle(u(\boldsymbol{x}))-f(\boldsymbol{x})]\cdot\delta ud\Omega=0 & ,\boldsymbol{x}\in\Omega.\label{eq:galerkin_form_poisson}
\end{align}
Using the Gaussian integration formula, we can transform the above equation to:
\begin{equation}
	\int_{\Omega}(-u_{,ii}-f)\delta ud\Omega=\int_{\Omega}u_{,i}\delta u_{,i}d\Omega-\int_{\Gamma}u_{,i}n_{i}\delta ud\Gamma-\int_{\Omega}f\delta ud\Omega=0.\label{eq:gaussian_poisson}
\end{equation}
By incorporating the boundary conditions from \Cref{eq:poisson_equation} into \Cref{eq:gaussian_poisson},
we obtain the Garlerkin weak form:
\begin{equation}
	\int_{\Omega}(-u_{,ii}-f)\delta ud\Omega=\int_{\Omega}u_{,i}\delta u_{,i}d\Omega-\int_{\Gamma^{t}}\bar{t}\delta ud\Gamma-\int_{\Omega}f\delta ud\Omega=0.\label{eq:weak form}
\end{equation}

Since $u(\boldsymbol{x})$ is given on $\Gamma^{u}$, the corresponding variation $\delta\boldsymbol{u}=0$
on $\Gamma^{u}$. Here, we observe an interesting phenomenon: we must satisfy $u(\boldsymbol{x})=\bar{u}(\boldsymbol{x})$
on $\Gamma^{u}$ in advance, which involves constructing an admissible function. This is crucial for
DEM (Deep energy form), and DEM essentially refers to a numerical algorithm that utilizes neural networks to solve the energy form of PDEs. Additionally, \Cref{eq:weak form} includes the domain PDEs and the boundary conditions on $\Gamma^{t}$.
Therefore, solving \Cref{eq:weak form} is equivalent to solving \Cref{eq:poisson_equation}.

We can further use the variational principle to write \Cref{eq:weak form} as:

\begin{equation}
	\delta\mathcal{L}=\int_{\Omega}u_{,i}\delta u_{,i}d\Omega-\int_{\Gamma^{t}}\bar{t}\delta ud\Gamma-\int_{\Omega}f\delta ud\Omega\label{eq:first_vari}
\end{equation}
\begin{equation}
	\mathcal{L}=\frac{1}{2}\int_{\Omega}u_{,i}u_{,i}d\Omega-\int_{\Gamma^{t}}\bar{t}ud\Gamma-\int_{\Omega}fud\Omega\label{eq:energy}
\end{equation}
$\mathcal{L}$ represents the potential energy. 
\Cref{eq:first_vari} is equivalent to \Cref{eq:poisson_equation},
and we can observe that $\delta^{2}\mathcal{L}>0$ (excluding
zero solutions), indicating that we can solve for $u(\boldsymbol{x})$ by minimizing the energy:

\begin{equation}
	u(\boldsymbol{x})=\underset{u}{\arg\min}\mathcal{L}\label{eq:minimum_energy}.
\end{equation}
The essence of DEM is to approximate $u(\boldsymbol{x})$ using a neural network $u(\boldsymbol{x};\boldsymbol{\theta})$,
and then optimize \Cref{eq:minimum_energy}:
\begin{equation}
	u(\boldsymbol{x};\boldsymbol{\theta})=\underset{\boldsymbol{\theta}}{\arg\min}\mathcal{L}_{DEM}=\underset{u}{\arg\min}\{\frac{1}{2}\int_{\Omega}u(\boldsymbol{x};\boldsymbol{\theta})_{,i}u(\boldsymbol{x};\boldsymbol{\theta})_{,i}d\Omega-\int_{\Gamma^{t}}\bar{t}u(\boldsymbol{x};\boldsymbol{\theta})d\Gamma-\int_{\Omega}fu(\boldsymbol{x};\boldsymbol{\theta})d\Omega\}\label{eq:DEM}.
\end{equation}
Therefore, the core of DEM lies in the integration of the domain energy and boundary energy, as well
as the construction of the admissible function. Integration strategies can use numerical analysis methods,
such as simple Monte Carlo integration, or more accurate methods like Gaussian integration or Simpson's
Rule.

Here, we emphasize the construction of the admissible function. We use the concept of a distance network
for the construction of the admissible function:
\begin{equation}
u(\boldsymbol{x})=u_{p}(\boldsymbol{x};\boldsymbol{\theta}_{p})+D(\boldsymbol{x})*u_{g}(\boldsymbol{x};\boldsymbol{\theta}_{g})\label{eq:admissible},
\end{equation}
where $u_{p}(\boldsymbol{x};\boldsymbol{\theta}_{p})$ is the particular solution network that fits the
Dirichlet boundary condition, such that it outputs $\bar{u}(\boldsymbol{x})$ when the input points are
on $\Gamma^{u}$, and outputs any value elsewhere. The parameters $\boldsymbol{\theta}_{p}$ are optimized
by:
\begin{equation}
\boldsymbol{\theta}_{p}=\underset{\boldsymbol{\theta}_{p}}{\arg\min}MSE(u_{p}(\boldsymbol{x};\boldsymbol{\theta}_{p}),\bar{u}(\boldsymbol{x})),\boldsymbol{x}\in\Gamma^{u},
\end{equation}
where $D(x)$ is the distance network, which we approximate using radial basis functions \citep{wang2022cenn,bai2023physics2}.
Other fitting functions can also be used. The effect is to output the minimum distance to the Dirichlet
boundary:
\begin{equation}
D(\boldsymbol{x})=\min_{\boldsymbol{y}\in\Gamma^{u}}\sqrt{(\boldsymbol{x}-\boldsymbol{y})\cdot(\boldsymbol{x}-\boldsymbol{y})},
\end{equation}
where $u_{g}(\boldsymbol{x};\boldsymbol{\theta}_{g})$ is a standard neural network. When using the minimum
potential energy principle, we only optimize $\boldsymbol{\theta}_{g}$:
\begin{equation}
u(\boldsymbol{x};\boldsymbol{\theta}_{p},\boldsymbol{\theta}_{g})=\underset{\boldsymbol{\boldsymbol{\theta}_{g}}}{\arg\min}\mathcal{L}_{DEM}.
\end{equation}

Please note that not all PDEs have an energy form, but most have a corresponding weak form. The requirement that PDEs have the energy form is that PDEs need to satisfy the linear self-adjoint operator with $\delta^{2}\mathcal{L}>0$ (where $\mathcal{L}$ is the functional) mathematically. The
proof is provided in \ref{sec:PINN_DEM}.

\subsection{Introduction to the inverse form of PDEs \label{subsec:Introduction_BINN}}

BINN (Boundary-Integral Type Neural Networks) essentially refers to a numerical algorithm that utilizes neural networks to solve the inverse form of PDEs. The inverse form of PDEs is often represented by boundary integral equations. Mathematically, boundary integral equations are derived from the weighted residual method shown in \Cref{eq:original_form_weighted} using Gaussian integration. This process transforms differential operators from the trial function to the test function and uses the fundamental solution of the differential equation to convert all unknowns to the boundary.

To illustrate boundary integral equations, we start from \Cref{eq:poisson_equation} and transform the strong form into a weighted residual form:
\begin{equation}
\int_{\Omega}[-\Delta u(\boldsymbol{x}) \cdot w(\boldsymbol{x}) - f(\boldsymbol{x}) \cdot w(\boldsymbol{x})] \, d\Omega = 0.
\end{equation}
Using Gaussian integration twice, we can transfer the Laplacian operator \(\Delta\) to the weight function \(w(\boldsymbol{x})\), incorporating the boundary conditions to obtain the inverse form:
\begin{equation}
	\int_{\Omega}[-u\cdot(\triangle w)-f\cdot w]d\Omega+\int_{\Gamma^{u}}\bar{u}\frac{\partial w}{\partial\boldsymbol{n}}d\Gamma+\int_{\Gamma^{t}}u\frac{\partial w}{\partial\boldsymbol{n}}d\Gamma-\int_{\Gamma^{u}}\frac{\partial u}{\partial\boldsymbol{n}}wd\Gamma-\int_{\Gamma^{t}}\bar{t}wd\Gamma=0\label{eq:inverse_form},
\end{equation}
where $\boldsymbol{n}$ is the outer normal vector of the boundary. If we choose the weight function as the fundamental solution of the differential equation:
\begin{equation}
	\Delta w(\boldsymbol{x}; \boldsymbol{y}) = \delta(\boldsymbol{x} - \boldsymbol{y}) \label{eq:fundamental_solution},
\end{equation}
where \(\delta(\boldsymbol{x} - \boldsymbol{y})\) is the Dirac delta function, the solution to the above equation is denoted as \(w(\boldsymbol{x}; \boldsymbol{y}) = u^{f}(\boldsymbol{x}; \boldsymbol{y}) = -\ln(r)/(2\pi)\), with \(r = \sqrt{(\boldsymbol{x} - \boldsymbol{y}) \cdot (\boldsymbol{x} - \boldsymbol{y})}\). Substituting \Cref{eq:fundamental_solution} into \Cref{eq:inverse_form}, we get:
\begin{equation}
	\begin{alignedat}{1}c(\boldsymbol{y})u(\boldsymbol{y})+\int_{\Gamma^{u}}\frac{\partial u(\boldsymbol{x})}{\partial\boldsymbol{n}}u^{f}(\boldsymbol{x};\boldsymbol{y})d\Gamma-\int_{\Gamma^{t}}u(\boldsymbol{x})\frac{\partial u^{f}(\boldsymbol{x};\boldsymbol{y})}{\partial\boldsymbol{n}}\Gamma=\\
		-\int_{\Omega}f(\boldsymbol{x})\cdot u^{f}(\boldsymbol{x};\boldsymbol{y})d\Omega+\int_{\Gamma^{u}}\bar{u}(\boldsymbol{x})\frac{\partial u^{f}(\boldsymbol{x};\boldsymbol{y})}{\partial\boldsymbol{n}}\Gamma-\int_{\Gamma^{t}}\bar{t}(\boldsymbol{x})u^{f}(\boldsymbol{x};\boldsymbol{y})d\Gamma
	\end{alignedat}
	\label{eq:solution_inverse}
\end{equation}
where \(c(\boldsymbol{y})\) is determined by boundary continuity, being 0.5 on a smooth boundary, 1 inside the domain, and 0 outside the domain. It is evident that, when \(\boldsymbol{y}\) on the boundary, all unknowns in \Cref{eq:solution_inverse} are located on the boundary:
\begin{equation}
	\begin{aligned}
		\frac{\partial u(\boldsymbol{x})}{\partial n}, & \quad \boldsymbol{x} \in \Gamma^{u} \\
		u(\boldsymbol{x}), & \quad \boldsymbol{x} \in \Gamma^{t}.
	\end{aligned}
\end{equation}
We use a neural network \(\phi(\boldsymbol{x}; \boldsymbol{\theta})\) to approximate these unknowns \(u(\boldsymbol{x})\) on \(\Gamma^{t}\) and \(\partial u(\boldsymbol{x})/\partial n\) on \(\Gamma^{u}\). Note that \(c(\boldsymbol{y}) u(\boldsymbol{y})\) is chosen as:
\begin{equation}
\begin{cases}
	c(\boldsymbol{y}) u(\boldsymbol{y}) = c(\boldsymbol{y}) \bar{u}(\boldsymbol{y}) & \boldsymbol{y} \in \Gamma^{u}, \\
	c(\boldsymbol{y}) u(\boldsymbol{y}) = c(\boldsymbol{y}) \phi(\boldsymbol{y}; \boldsymbol{\theta}) & \boldsymbol{y} \in \Gamma^{t}.
\end{cases}
\end{equation}
The essence of BINN is to solve \Cref{eq:solution_inverse}, and the loss function for BINN is:
\begin{equation}
	\begin{aligned}
		\mathcal{L}_{BINN} = & \frac{1}{N} \sum_{i=1}^{N_{s}} |R(\boldsymbol{y}_{i}; \boldsymbol{\theta})|, \quad \boldsymbol{y}_{i} \in \Gamma \\
		R(\boldsymbol{y}; \boldsymbol{\theta}) = & c(\boldsymbol{y}) u(\boldsymbol{y}) + \int_{\Gamma^{u}} \frac{\partial \phi(\boldsymbol{x}; \boldsymbol{\theta})}{\partial n} u^{f}(\boldsymbol{x}; \boldsymbol{y}) \, d\Gamma - \int_{\Gamma^{t}} \phi(\boldsymbol{x}; \boldsymbol{\theta}) \frac{\partial u^{f}(\boldsymbol{x}; \boldsymbol{y})}{\partial n} \, d\Gamma \\
		& + \int_{\Omega} f(\boldsymbol{x}) \cdot u^{f}(\boldsymbol{x}; \boldsymbol{y}) \, d\Omega - \int_{\Gamma^{u}} \bar{u}(\boldsymbol{x}) \frac{\partial u^{f}(\boldsymbol{x}; \boldsymbol{y})}{\partial n} \, d\Gamma + \int_{\Gamma^{t}} \bar{t}(\boldsymbol{x}) u^{f}(\boldsymbol{x}; \boldsymbol{y}) \, d\Gamma,
	\end{aligned}\label{eq:BINN_loss_funciton}
\end{equation}
where \(N_{s}\) is the number of source points, i.e., the points at which the loss function is evaluated. 
If the interior source term $f(\boldsymbol{x})=0$, \Cref{eq:BINN_loss_funciton} will be more concise and only involve boundary integrals. Note that the $u^f(\boldsymbol{x},\boldsymbol{y})$ and $\partial u^{f}(\boldsymbol{x;y})\slash\partial \boldsymbol{n}$ will be singular when $\boldsymbol{x}\rightarrow\boldsymbol{y}$. 
To be specific,
integrals related to $u^f(\boldsymbol{x},\boldsymbol{y})$ are weakly singular integrals, while integrals related to $\partial u^{f}(\boldsymbol{x;y})\slash\partial \boldsymbol{n}$ are Cauchy-principle integrals \cite{1958longman}.
Thus, singular integrals must be treated carefully to ensure accuracy. In the present work, a piece-wise integration strategy same as \cite{sun2023binn} is adopted, where regularization techniques are involved to remove the singularity.  A brief demonstration is presented in \ref{sec:app_integral}.

\section{Method} \label{sec:method}

The purpose of the previous chapter was to gain a better understanding of the different numerical formats for solving PDEs. With the basic concepts established, our idea is quite simple: we propose replacing the traditional MLP used in AI for PDEs with KAN (Kolmogorov–Arnold Networks) \cite{liu2024kan} in different PDEs forms (strong, energy, and inverse form), subsequently named KINN. In the following sections, we introduce the KAN network and KINN.

\subsection{Kolmogorov–Arnold Networks}

In KAN, the weights to be optimized depend on the number of input neurons \(l_{i}\) and output neurons \(l_{o}\) in each layer of KAN. We denote the activation function for the layer of KAN as \(\phi_{ij}\), where \(i \in \{1, 2, \cdots, l_{o}\}\) and \(j \in \{1, 2, \cdots, l_{i}\}\). Each element of the spline activation function \(\phi_{ij}\) is determined by the number of grids size \(G\) and the order of the B-splines \(r\). The specific expression is given by:
\begin{equation}
\phi_{ij}(\boldsymbol{X})=\left[\begin{array}{cccc}
	\sum_{m=1}^{G_{1}+r_{1}}c_{m}^{(1,1)}B_{m}(x_{1}) & \sum_{m=1}^{G_{2}+r_{2}}c_{m}^{(1,2)}B_{m}(x_{2}) & \cdots & \sum_{m=1}^{G_{l_{i}}+r_{l_{i}}}c_{m}^{(1,l_{i})}B_{m}(x_{l_{i}})\\
	\sum_{m=1}^{G_{1}+r_{1}}c_{m}^{(2,1)}B_{m}(x_{1}) & \sum_{m=1}^{G_{2}+r_{2}}c_{m}^{(2,2)}B_{m}(x_{2}) & \cdots & \sum_{m=1}^{G_{l_{i}}+r_{l_{i}}}c_{m}^{(2,l_{i})}B_{m}(x_{l_{i}})\\
	\vdots & \vdots & \ddots & \vdots\\
	\sum_{m=1}^{G_{1}+r_{1}}c_{m}^{(l_{o},1)}B_{m}(x_{1}) & \sum_{m=1}^{G_{2}+r_{2}}c_{m}^{(l_{o},2)}B_{m}(x_{2}) & \cdots & \sum_{m=1}^{G_{l_{i}}+r_{l_{i}}}c_{m}^{(l_{o},l_{i})}B_{m}(x_{l_{i}})
\end{array}\right]
\end{equation}
where \(G_{j}\) is the number of grids in the \(j\)-th input direction, and \(r_{j}\) is the order of the B-splines in the \(j\)-th input direction, with \(j \in \{1, 2, \cdots, l_{i}\}\). The coefficients \(c_{m}^{(i,j)}\) are the B-spline coefficients, whose number is determined by the grid \(G_{j}\) and the order \(r_{j}\), totaling \(G_{j}+r_{j}\). Note that the grid division and order in each input direction are independent and can be selected individually. \(B_{m}\) is the  basis function of the B-spline.

To enhance the fitting ability of the activation function, we introduce \(S_{ij}\), a matrix of the same size as \(\phi_{ij}\), given by:
\begin{equation}
S_{ij}=\left[\begin{array}{cccc}
	s_{11} & s_{12} & \cdots & s_{1l_{i}}\\
	s_{21} & s_{22} & \cdots & s_{2l_{i}}\\
	\vdots & \vdots & \ddots & \vdots\\
	s_{l_{o}1} & s_{l_{o}2} & \cdots & s_{l_{o}l_{i}}
\end{array}\right].
\end{equation}
The role of \(\boldsymbol{S}\) is to adjust the magnitude of the activation function \(\boldsymbol{\phi}\), i.e., \(\boldsymbol{\phi} = \boldsymbol{\phi} \odot \boldsymbol{S}\), where \(\odot\) denotes element-wise multiplication. Additionally, we have  nonlinear activation and linear matrix multiplication operations, and the final output is given by:
\begin{equation}
\boldsymbol{Y} = [\sum_{\text{column}}\boldsymbol{\phi}(\boldsymbol{X}) \odot \boldsymbol{S}] + \boldsymbol{W} \cdot \sigma(\boldsymbol{X}), \label{eq:function_each_kan}
\end{equation}
where \(\boldsymbol{W}\) is the linear matrix operation, and \(\sigma\) is the nonlinear activation function. The inclusion of \(\sigma\) ensures smooth function fitting; without it, relying solely on B-splines could lead to a rough approximation. \(\boldsymbol{S}\) and \(\boldsymbol{W}\) act as scaling factors, similar to normalization in machine learning. The term \(\boldsymbol{W} \cdot \sigma(\boldsymbol{X})\) is a residual term, akin to the ResNet approach \cite{he2016deep}. The code of
KAN is adapted from \url{https://github.com/Blealtan/efficient-kan} and \url{https://github.com/KindXiaoming/pykan}

Assuming that the grid division and B-spline order in each input neuron direction are consistent, the number of $c_{m}^{(i,j)}$ are all $G+r$. The trainable parameters are shown in \Cref{tab:trainable_para_KAN}. 

\begin{table}
	\caption{The trainable parameter in KAN\label{tab:trainable_para_KAN}}
	\begin{adjustbox}{max width=\textwidth}
	\begin{centering}
		\begin{tabular}{|c|c|c|c|}
			\hline 
			Trainable parameters & Variable & Number & Description\tabularnewline
			\hline 
			$c_{m}^{(i,j)}$ & spline\_weight & $l_{o}*l_{i}*(G+r)$ & Coefficients of B-spline in activation function $\boldsymbol{\phi(\boldsymbol{X})}$\tabularnewline
			\hline 
			$W_{ij}$ & base\_weight & $l_{i}*l_{o}$ & Linear transformation for nonlinear activation function $\sigma(\boldsymbol{X})$ \tabularnewline
			\hline 
			$S_{ij}$ & spline\_scaler & $l_{i}*l_{o}$ & Scaling factors of activation function $\boldsymbol{\phi(\boldsymbol{X})}$ \tabularnewline
			\hline 
		\end{tabular}
		\par\end{centering}
	\end{adjustbox}
\end{table}

\subsection{KINN}

\begin{figure}
	\begin{centering}
		\includegraphics[scale=0.6]{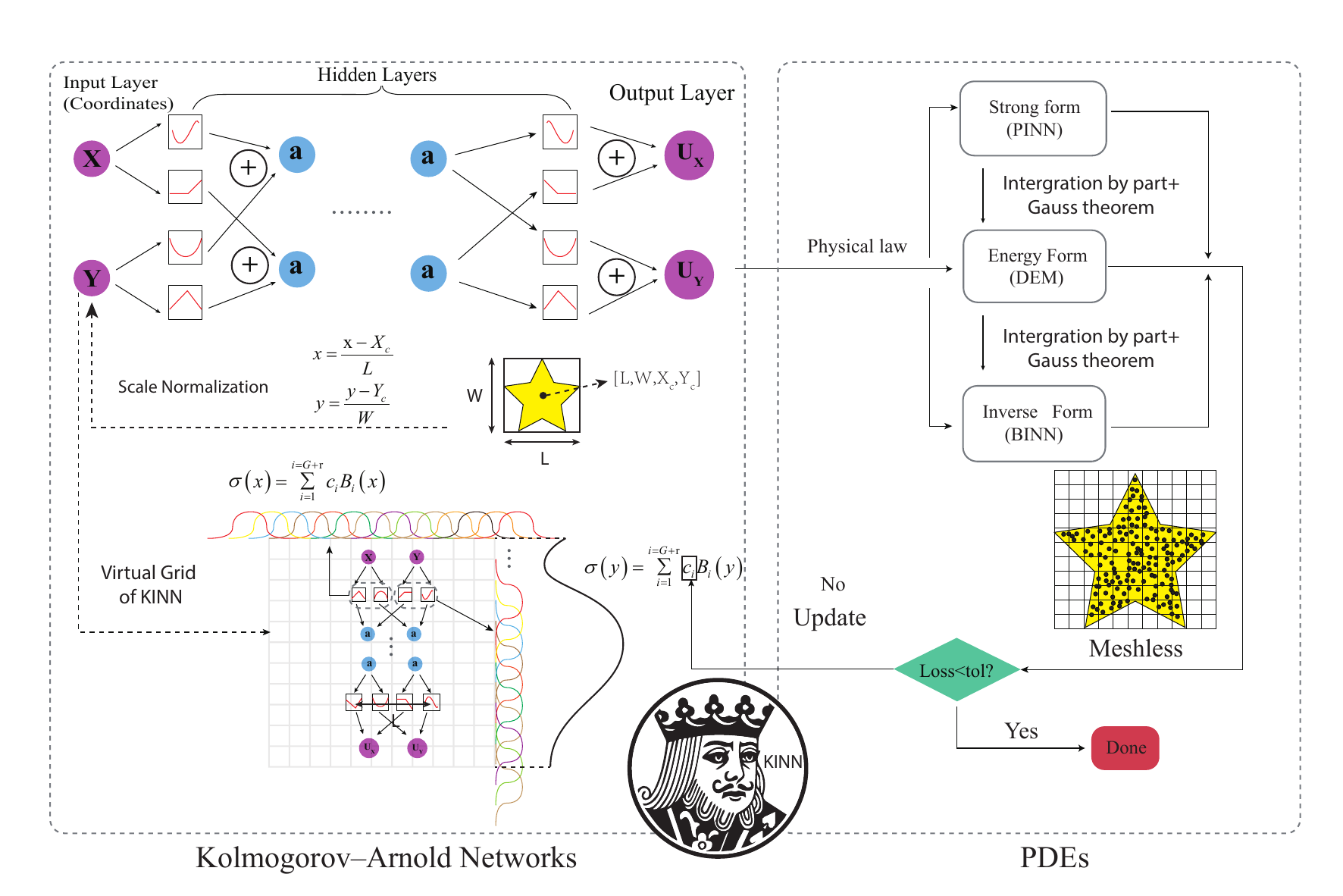}
		\par\end{centering}
	\caption{Schematic of KINN: The idea of KINN is to replace MLP with Kolmogorov–Arnold Networks (KAN) in different PDEs forms (strong, energy, and inverse forms). In KAN, the main training parameters are the undetermined coefficients \(c_{i}\) of the B-splines in the activation function. KINN establishes the loss function based on different numerical formats of PDEs and optimizes the \(c_{i}\). The virtual Grid is determined by the grid size in KAN. \label{fig:Schematic-of-KINN:}}
\end{figure}

\Cref{fig:Schematic-of-KINN:} illustrates the core idea of KINN \footnote{KINN refers to various PDEs forms (strong: PINNs, energy: DEM, and inverse: BINN) for solving PDEs based on KAN, so sometimes we use KAN directly to represent the various PDEs forms.}, which uses KAN instead of MLP in the different forms of PDEs (strong, energy, and inverse form). Notably, as the output range can easily exceed the grid range in B-splines after a single KAN layer (we must set the grid range in advance), we apply a tanh activation function to the output of the KAN layer to constrain it within \([-1,1]\) and better utilize the capabilities of the B-splines in this range. Thus, the output of each KAN layer undergoes an additional tanh activation:
\begin{equation}
\boldsymbol{Y}^{\text{new}} = \tanh(\boldsymbol{Y}) = \tanh\left\{[\sum_{\text{column}}\boldsymbol{\phi}(\boldsymbol{X}) \odot \boldsymbol{S}] + \boldsymbol{W} \cdot \sigma(\boldsymbol{X})\right\}.
\end{equation}
However, we do not apply tanh to the final layer since the output does not necessarily lie within \([-1,1]\). If the simulation domain is outside the \([-1,1]\) grid range, we perform a scale normalization for input $\boldsymbol{X}$. For instance, we can find the smallest bounding box enclosing the geometry to be simulated, denoted as \([L, W, X_{c}, Y_{c}]\) as shown in \Cref{fig:Schematic-of-KINN:}, and then  normalize the input as follows:
\begin{equation}
	 x^{s} = \frac{x - X_{c}}{L/2}; \quad y^{s} = \frac{y - Y_{c}}{W/2}. 
\end{equation}

Once the grid size, grid range, and order of the B-splines are determined, the KAN approximation function is fixed, forming a virtual grid approximation function as shown in \Cref{fig:Schematic-of-KINN:}. Next, we select the appropriate form of PDEs, such as strong form, energy form, or inverse form, as detailed in  \Cref{sec:Preparatory-knowledge}. After determining the PDE form, we randomly sample points within the domain and on the boundary to compute the corresponding loss functions, as the meshless random sampling in the strong form of PINNs. Finally, we optimize the trainable parameters in KAN, as shown in \Cref{tab:trainable_para_KAN}.

KINN shares similarities with finite element methods (FEM) and isogeometric analysis (IGA). The approximation function of KAN is like composition of functions based on NURBS in IGA or shape functions in FEM, as proven in  \ref{sec:Similarities-KAN_FEM}. Additionally, the highest derivative order of the PDEs that KINN can solve depends on the selected activation function, the number of layers, and the order of B-splines, as analyzed in  \ref{sec:The-highest-order}. Below, we present numerical results of solving PDEs using KINN. We compare the results of PDEs in strong form, energy form, and inverse form between MLP and the corresponding KINN. All computations are carried out on a single Nvidia RTX 4060 Ti GPU with 16GB memory.

\section{Result\label{sec:Result}}

In this chapter, we demonstrate the performance of KINN in solving different PDE formats, including the strong, energy, and inverse forms. To compare the performance of MLP and KAN in solving PDEs, we ensure that all methods use the same optimization method and learning rate for both KAN and MLP unless otherwise specified. The only difference is the type of neural network used.

\subsection{KAN can, MLP cannot}

To demonstrate the advantages of KAN, we present a function-fitting example that MLP fails to achieve. Wang et al. \cite{NTK_PINN} found that MLPs perform poorly on multi-scale PDE problems. According to the Neural Tangent Kernel (NTK) theory \cite{jacot2018neural}, MLPs are ineffective in fitting high-frequency components of functions. This is because the eigenvalues of the NTK matrix for high-frequency components are smaller than those for low-frequency components. To illustrate this point, we first introduce the NTK matrix:
\begin{equation}
	\begin{aligned}
		\boldsymbol{K}_{ntk}(\boldsymbol{x},\boldsymbol{x}^{'}) & = -\lim_{\eta \rightarrow 0} \frac{f(\boldsymbol{x};\boldsymbol{\theta} - \eta \frac{\partial f}{\partial \boldsymbol{\theta}}|_{\boldsymbol{x}^{'}}) - f(\boldsymbol{x}; \boldsymbol{\theta})}{\eta} \\
		& = -\lim_{\eta \rightarrow 0} \frac{f(\boldsymbol{x}; \boldsymbol{\theta}) - \eta (\frac{\partial f}{\partial \boldsymbol{\theta}}|_{\boldsymbol{x}^{'}}) \circ (\frac{\partial f}{\partial \boldsymbol{\theta}}|_{\boldsymbol{x}}) + O(\eta \frac{\partial f}{\partial \boldsymbol{\theta}}|_{\boldsymbol{x}^{'}}) - f(\boldsymbol{x}; \boldsymbol{\theta})}{\eta} = (\frac{\partial f}{\partial \boldsymbol{\theta}}|_{\boldsymbol{x}}) \cdot (\frac{\partial f}{\partial \boldsymbol{\theta}}|_{\boldsymbol{x}^{'}})
	\end{aligned}
	\label{eq:NTK_matrix}
\end{equation}
where \(f\) is the function to be fitted, such as the displacement field in mechanics. The elements of the NTK matrix are computed for all training points. If there are \(N\) points, the NTK matrix \(\boldsymbol{K}_{ntk}\) is an \(N \times N\) non-negative definite matrix. The NTK matrix is calculated using the last term in \Cref{eq:NTK_matrix}, by computing the gradients at points \(\boldsymbol{x}\) and \(\boldsymbol{x}^{'}\) and then taking their dot product. The limit form in \Cref{eq:NTK_matrix} describes the NTK matrix, which measures the change at point \(\boldsymbol{x}\) when performing gradient descent at point \(\boldsymbol{x^{'}}\). As neural network parameters are interdependent, the NTK matrix can be used to measure the convergence of neural network algorithms (a larger change implies faster convergence).

Next, we consider the least squares loss function:
\begin{equation}
\begin{aligned}
	\mathcal{L} & = \frac{1}{2n} \sum_{i=1}^{n} [f(\boldsymbol{x}_{i}; \boldsymbol{\theta}) - y_{i}]^{2} = \frac{1}{n} \sum_{i=1}^{n} l_{i}  \\
	l_{i} & = \frac{1}{2} [f(\boldsymbol{x}_{i}; \boldsymbol{\theta}) - y_{i}]^{2} \label{eq:square_loss}
\end{aligned}
\end{equation}
The gradient of \Cref{eq:square_loss} with respect to the parameters is:
\begin{equation}
	\begin{aligned}\frac{\partial\mathcal{L}}{\partial\boldsymbol{\theta}} & =\frac{1}{n}\sum_{i=1}^{n}\{[f(\boldsymbol{x}_{i};\boldsymbol{\theta})-y_{i}]\frac{\partial f(\boldsymbol{x}_{i};\boldsymbol{\theta})}{\partial\boldsymbol{\theta}}\}=\frac{1}{n}\sum_{i=1}^{n}\frac{\partial l_{i}}{2\partial\boldsymbol{\theta}}\\
		\frac{\partial l_{i}}{\partial\boldsymbol{\theta}} & =[f(\boldsymbol{x}_{i};\boldsymbol{\theta})-y_{i}]\frac{\partial f(\boldsymbol{x}_{i};\boldsymbol{\theta})}{\partial\boldsymbol{\theta}}
	\end{aligned}
	\label{eq:every_loss_grad}
\end{equation}
Considering the dynamic gradient and substituting \Cref{eq:every_loss_grad}:
\begin{equation}
	\begin{aligned}
		\frac{df(\boldsymbol{x}_{i}; \boldsymbol{\theta})}{dt} & = \lim_{\eta \rightarrow 0} \frac{f(\boldsymbol{x}_{i}; \boldsymbol{\theta} - \eta \frac{\partial l_{i}}{\partial \boldsymbol{\theta}}|_{\boldsymbol{x}_{i}}) - f(\boldsymbol{x}_{i}; \boldsymbol{\theta})}{\eta} = -\frac{\partial l_{i}}{\partial \boldsymbol{\theta}}|_{\boldsymbol{x}_{i}} \cdot \frac{\partial f}{\partial \boldsymbol{\theta}}|_{\boldsymbol{x}_{i}} \\
		& = -(\frac{\partial f}{\partial \boldsymbol{\theta}}|_{\boldsymbol{x}_{i}} \cdot \frac{\partial f}{\partial \boldsymbol{\theta}}|_{\boldsymbol{x}_{i}}) [f(\boldsymbol{x}_{i}; \boldsymbol{\theta}) - y_{i}]
	\end{aligned}
	\label{eq:dynamic_grad}
\end{equation}
Comparing \Cref{eq:dynamic_grad} and \Cref{eq:NTK_matrix}, we obtain the vector form of \Cref{eq:dynamic_grad}:
\begin{align}
	\frac{d\boldsymbol{f}(\boldsymbol{X}; \boldsymbol{\theta})}{dt} & = -\boldsymbol{K}_{ntk} \cdot [\boldsymbol{f}(\boldsymbol{X}; \boldsymbol{\theta}) - \boldsymbol{Y}] \nonumber \\
	\boldsymbol{f}(\boldsymbol{X}; \boldsymbol{\theta}) & = [f(\boldsymbol{x}_{1}; \boldsymbol{\theta}), f(\boldsymbol{x}_{2}; \boldsymbol{\theta}), \cdots, f(\boldsymbol{x}_{n}; \boldsymbol{\theta})]^{T} \label{eq:ODE} \\
	\boldsymbol{Y} & = [y_{1}, y_{2}, \cdots, y_{n}]^{T} \nonumber
\end{align}
Since a first-order linear ordinary differential equation system has an analytical solution, we obtain the particular solution of \Cref{eq:ODE}:
\begin{equation}
	\boldsymbol{f}(\boldsymbol{X}; \boldsymbol{\theta}) = (\boldsymbol{I} - e^{-t\boldsymbol{K}_{ntk}}) \cdot \boldsymbol{Y} \label{eq:f_analytical}
\end{equation}
Next, we perform an eigenvalue decomposition on \(e^{-\boldsymbol{K}_{ntk}t}\). Since \(\boldsymbol{K}_{ntk}\) is a real symmetric matrix, it can be diagonalized as \(\boldsymbol{K}_{ntk} = \boldsymbol{Q} \cdot \boldsymbol{\lambda} \cdot \boldsymbol{Q}^{T}\), where \(\boldsymbol{Q} = [\boldsymbol{q}_{1}, \boldsymbol{q}_{2}, \cdots, \boldsymbol{q}_{n}]\). \(\boldsymbol{q}_{i}\) are the eigenvectors of \(\boldsymbol{K}_{ntk}\). Therefore:
\begin{equation}
	e^{-t\boldsymbol{K}_{ntk}} = e^{-\boldsymbol{Q} \cdot \boldsymbol{\lambda} \cdot \boldsymbol{Q}^{T}t} = \boldsymbol{Q} \cdot e^{-\boldsymbol{\lambda}t} \cdot \boldsymbol{Q}^{T} \label{eq:eigvalue}
\end{equation}
Substituting \Cref{eq:eigvalue} into \Cref{eq:f_analytical}, we get:
\begin{equation}
	\begin{aligned}
		\boldsymbol{f}(\boldsymbol{X}; \boldsymbol{\theta}) & = (\boldsymbol{I} - \boldsymbol{Q} \cdot e^{-\boldsymbol{\lambda}t} \cdot \boldsymbol{Q}^{T}) \cdot \boldsymbol{Y} \\
		\boldsymbol{Q}^{T}[\boldsymbol{f}(\boldsymbol{X}; \boldsymbol{\theta}) - \boldsymbol{Y}] & = e^{-\boldsymbol{\lambda}t} \cdot \boldsymbol{Q}^{T} \cdot \boldsymbol{Y}
	\end{aligned}
\end{equation}
\begin{align}
	\left[\begin{array}{c}
		\boldsymbol{q}_{1}^{T} \\
		\boldsymbol{q}_{2}^{T} \\
		\vdots \\
		\boldsymbol{q}_{n}^{T}
	\end{array}\right] & \left[\begin{array}{c}
		f(\boldsymbol{x}_{1}; \boldsymbol{\theta}) - y_{1} \\
		f(\boldsymbol{x}_{2}; \boldsymbol{\theta}) - y_{2} \\
		\vdots \\
		f(\boldsymbol{x}_{n}; \boldsymbol{\theta}) - y_{n}
	\end{array}\right] = \left[\begin{array}{cccc}
		e^{-\lambda_{1}t} \\
		& e^{-\lambda_{2}t} \\
		& & \ddots \\
		& & & e^{-\lambda_{n}t}
	\end{array}\right] \cdot \boldsymbol{Q}^{T} \cdot \boldsymbol{Y} \label{eq:res}
\end{align}
From \Cref{eq:res}, we observe that the residual \(f(\boldsymbol{x}_{i}; \boldsymbol{\theta}) - y_{i}\) is inversely proportional to the eigenvalues of \(\boldsymbol{K}_{ntk}\). The more evenly distributed the eigenvalues of \(\boldsymbol{K}_{ntk}\), the better the neural network's fitting ability. However, the eigenvalues of fully connected MLPs are generally concentrated, and the large eigenvalue eigenvectors are usually low-frequency. This causes MLPs to prefer low frequencies. Although Fourier feature embeddings can partially mitigate the low-frequency preference, they introduce an additional hyperparameter \(\sigma\). The above derivations and analyses are largely based on \cite{NTK_PINN}.

Naturally, we ask whether KAN can fundamentally solve the "spectral bias" of MLP.

\begin{figure}
	\begin{centering}
		\includegraphics[scale=0.7]{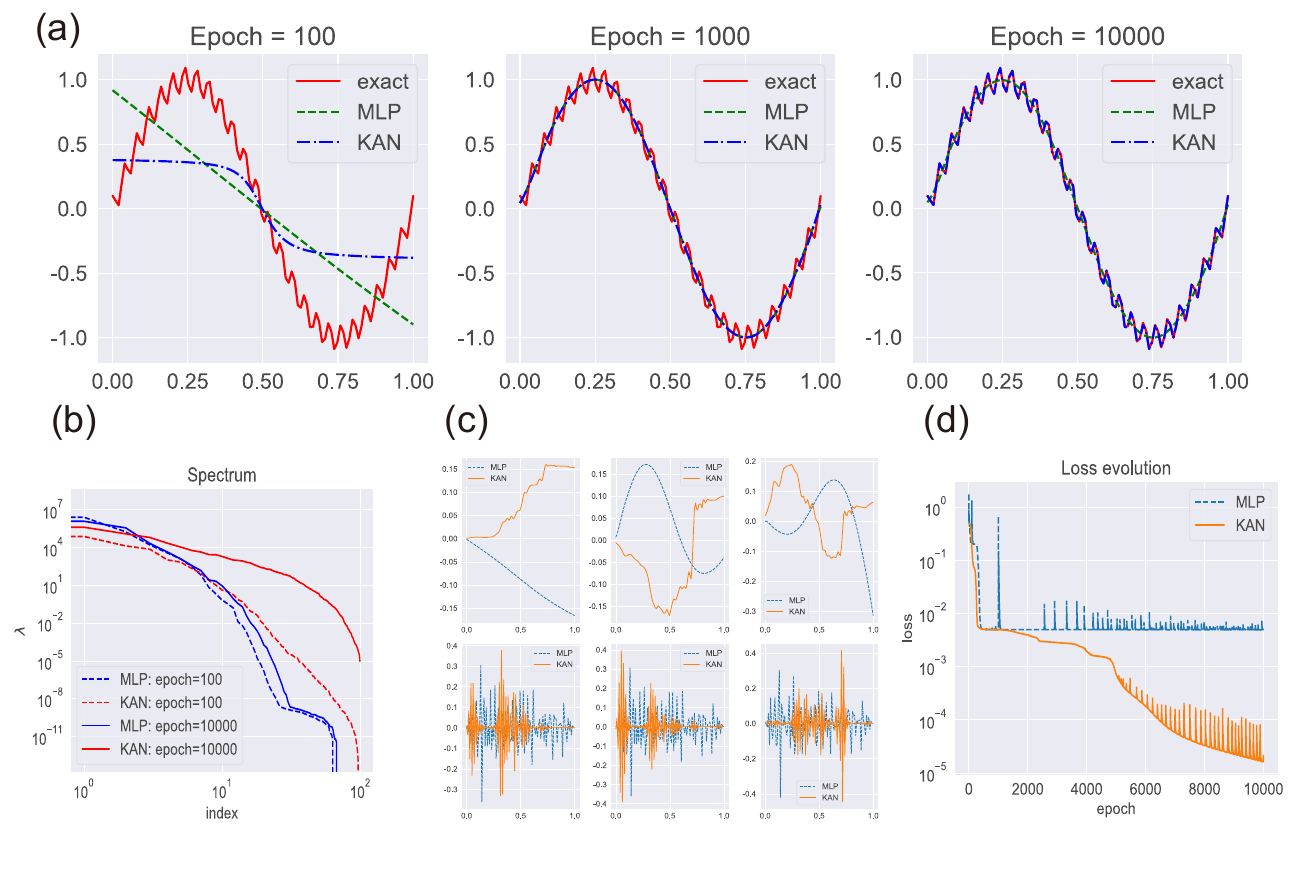}
		\par\end{centering}
	\caption{Validation of the "spectral bias" of MLP and KAN in fitting \(u = \sin(2\pi x) + 0.1\sin(50\pi x)\). (a) Exact solutions, MLP predictions, and KAN predictions at epochs 100, 1000, and 10000, (b) Eigenvalue distributions of MLP and KAN, (c) The First row is the eigenvectors of the largest eigenvalue sorted from largest to smallest, and the second row is the eigenvectors of the three smallest eigenvalues, (d) Evolution of the loss functions of MLP and KAN.\label{fig:MLP_KAN_function_appro}}
\end{figure}

To validate our hypothesis, we consider the following 1D Poisson problem and its analytical solution:
\begin{equation}
	\begin{cases}
		\frac{\partial^{2}u(x)}{\partial x^{2}} = -4\pi^{2} \sin(2\pi x) - 250\pi^{2} \sin(50\pi x), & x \in [0,1] \\
		u(0) = u(1) = 0 \\
		u = \sin(2\pi x) + 0.1\sin(50\pi x)
	\end{cases}
\end{equation}
To simplify, we first consider function fitting. If function fitting fails, solving PDEs will certainly fail. To objectively validate, we solve the same problem as in \cite{NTK_PINN}, referencing the code at \url{https://github.com/PredictiveIntelligenceLab/MultiscalePINNs}. We fit \(u = \sin(2\pi x) + 0.1\sin(50\pi x)\) using MLP and KAN. As shown in \Cref{fig:MLP_KAN_function_appro}a, we uniformly distribute 100 points in \([0,1]\). The structure of KAN is {[}1,5,5,5,1{]} with grid size=10 while the structure of MLP is {[}1,100,100,100,100,1{]}. We set the same learning rate (0.001) and optimizer (Adam) for objective comparison. \Cref{fig:MLP_KAN_function_appro}b demonstrates that KAN has a more evenly distributed eigenvalue spectrum compared to MLP. Additionally, \Cref{fig:MLP_KAN_function_appro}c shows the eigenvectors of KAN contain both high and low frequencies.
The eigenvalues of the NTK matrix represent the convergence rate under the corresponding eigenvectors, as shown in \Cref{eq:res}. Specifically, the eigenvalues indicate the convergence speed of the residual in the eigenvalue space. The eigenvectors in \Cref{fig:MLP_KAN_function_appro}c represent the basis modes fitted by the neural network. In summary, the eigenvalues and eigenvectors inform us of the convergence speed in the corresponding basis modes.
 KAN converges faster and with higher accuracy for problems with both low and high frequencies. 
\Cref{fig:MLP_KAN_function_appro}c shows that in the feature vector space, KAN exhibits similar patterns to MLP in the low-frequency range. However, KAN simultaneously carries high-frequency information within the low-frequency range. The fact that the eigenvector of KAN under the largest eigenvalue includes both high and low frequencies indicates that KAN can rapidly converge on a mix of high and low frequencies, unlike MLP, which tends to favor low frequencies.
This is because KAN's structure, utilizing B-splines, is more suitable for the multi-scale function approximation.

Thus, our results show that compared to MLP, KAN almost does not have the problem of "spectral bias" and can solve some problems that MLP cannot. The following numerical examples demonstrate the advantage of KAN in solving high and low-frequency mixed PDEs.

Since the above is a purely data-driven example, it is to demonstrate KAN's fitting ability for high and low-frequency problems. If MLP cannot fit a function approximation problem, it is unlikely to fit PDE-solving problems. As this work emphasizes the capabilities of KINN for solving PDEs, we solve the same heat conduction problem in \citet{NTK_PINN}:

\begin{equation}
	\begin{aligned}
		u_{t} = \frac{1}{(F\pi)^{2}} u_{xx}, & \quad x \in [0,1], t \in [0,1] \\
		u(x,0) = \sin(F\pi x), & \quad x \in [0,1] \\
		u(0,t) = u(1,t) = 0, & \quad t \in [0,1],
	\end{aligned}
	\label{eq:heat_multi}
\end{equation}
where $F$ is the frequency. The analytical solution we set for high and low-frequency mix is:

\begin{equation}
	u(x,t) = e^{-t} \sin(F\pi x)
\end{equation}

\begin{figure}
	\begin{centering}
		\includegraphics[scale=0.7]{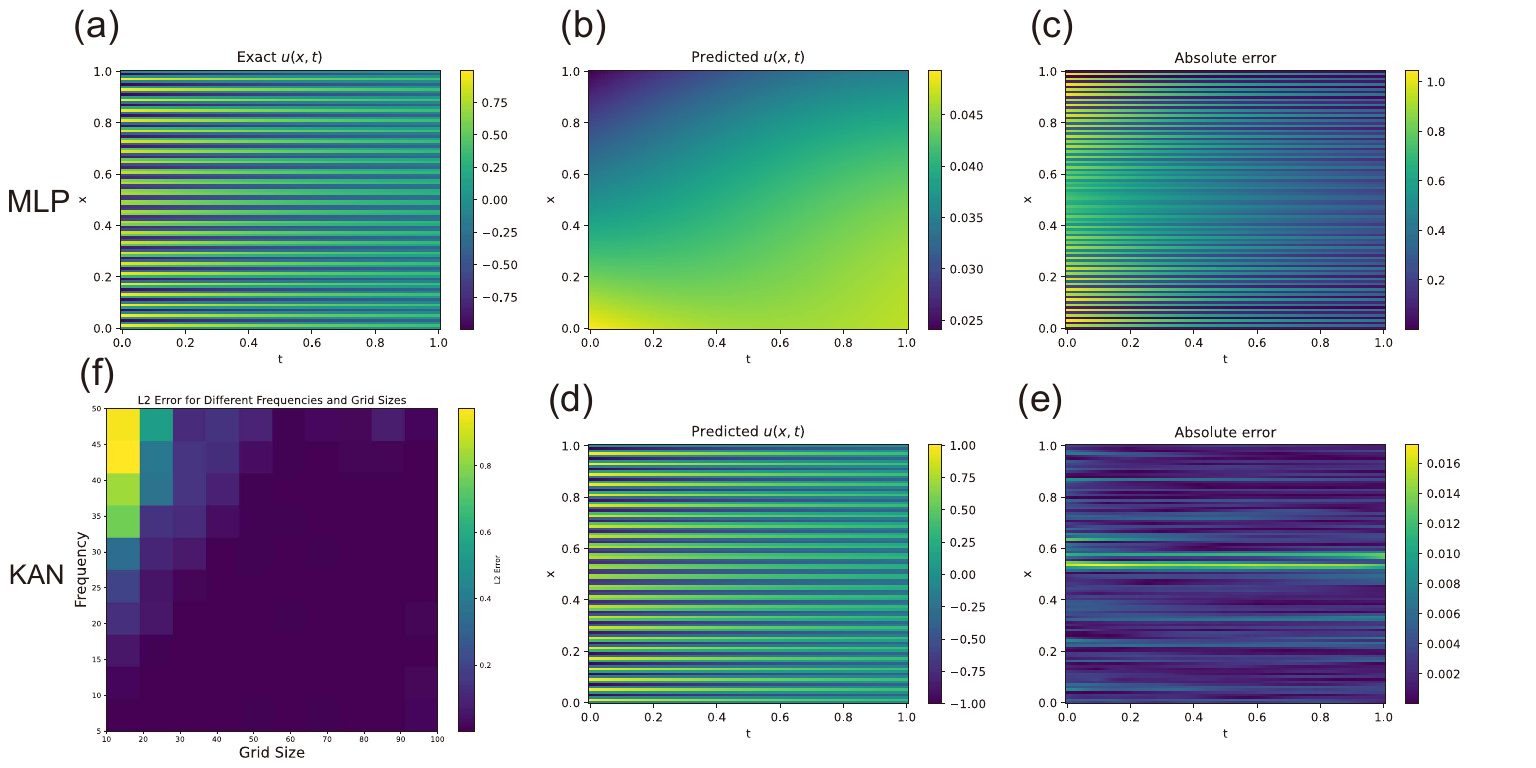}
		\par\end{centering}
	\caption{MLP and KAN fitting the high and low-frequency mixed heat conduction problem. (a) Exact solution for frequency $F=50$. (b) MLP prediction for frequency $F=50$ after 3000 iterations. (c) The absolute error of MLP for frequency $F=50$ after 3000 iterations. (d) KAN prediction for frequency $F=50$ after 3000 iterations. (e) The absolute error of KAN for frequency $F=50$ after 3000 iterations. (f) Relative error of KAN under different grid sizes and frequencies, with 3000 iterations and network structure [2,5,1].\label{fig:KAN_MLP_heat}}
\end{figure}

The structure of MLP is [2,30,30,30,30,1], and the structure of KAN is [2,5,5,1]. Apart from the different network structures, all other settings are the same. Note that the grid size of KAN is 100 and the order is 3. \Cref{fig:KAN_MLP_heat} shows the performance of MLP and KAN in solving \Cref{eq:heat_multi} (with $F=50$). The results are after 3000 epochs, where we can see that MLP completely fails to solve this problem, but KAN achieves good results. This further highlights that KAN's "spectral bias" is far less severe than that of MLP.

However, during our numerical experiments, we found that if KAN's grid size is too small, it also fails to solve the problem with high frequency $F$. Therefore,  \Cref{fig:KAN_MLP_heat}(f) shows the relative error $\mathcal{L}_{2}$ of KAN under different frequencies $F$ and grid sizes, with 3000 iterations and a network structure of [2,5,1]. The results indicate that KAN's "spectral bias" is related to the grid size. Nevertheless, compared to MLP, KAN's "spectral bias" is far less severe.

Due to the less severe "spectral bias" of KAN compared to MLP, KAN has greater potential for addressing multi-scale and multi-frequency issues, such as turbulence in fluid mechanics, than MLP-based approaches. This is a promising research direction for the future: using KAN to solve turbulence issues in fluid mechanics could achieve higher accuracy than MLP-based methods.

\subsection{KAN are more accurate than MLP in solving PDEs}

\subsubsection{Problems with singularities}

\begin{figure}
	\begin{centering}
		\includegraphics[scale=0.7]{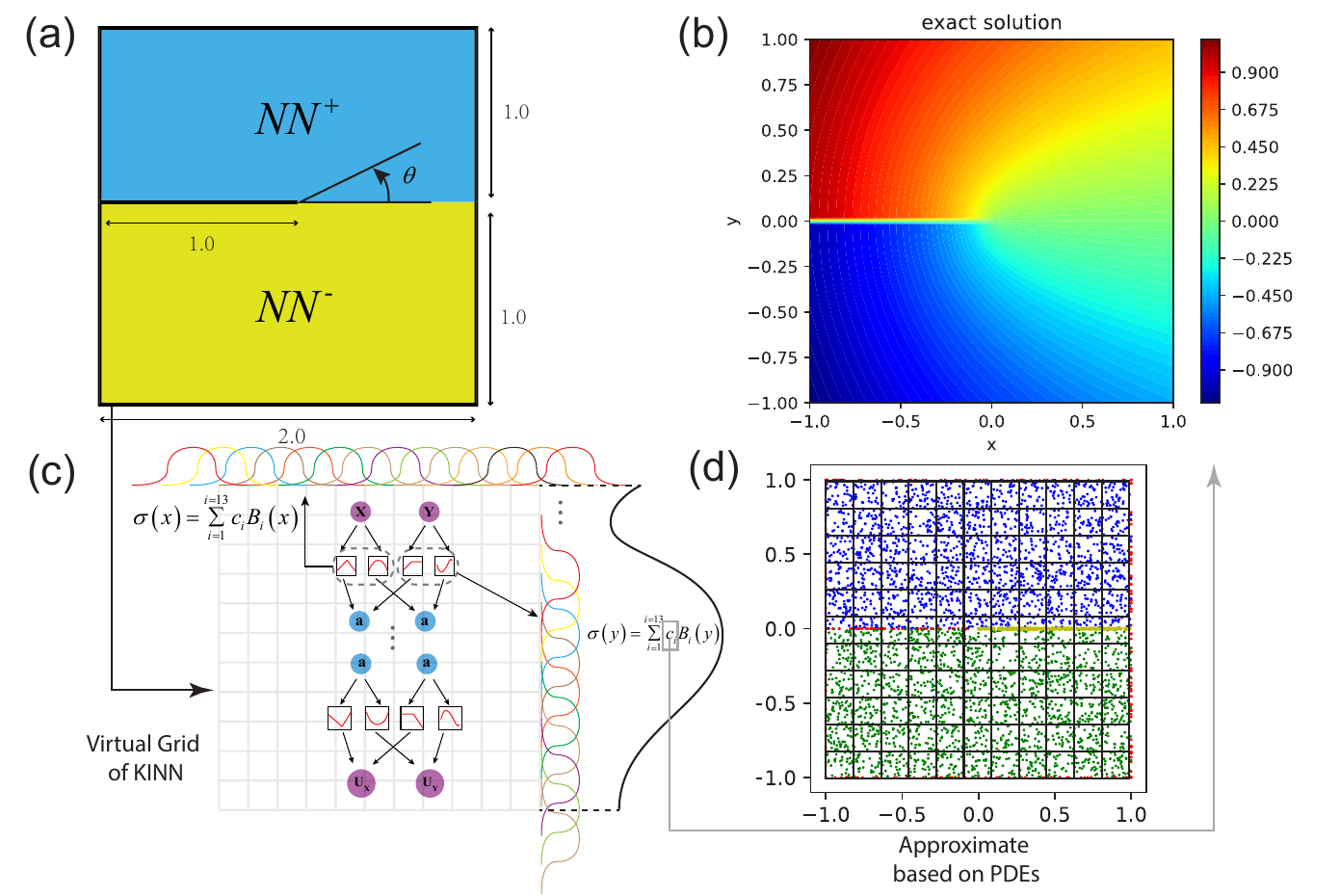}
		\par\end{centering}
	\caption{Introduction to mode III crack: (a) The structure of the mode III crack, with a size of $\text{[-1,1]}^{2}$ in a square region. $\theta$ is from $[-\pi, \pi]$. The blue and yellow regions represent two neural networks since the displacement is discontinuous at the crack ($x<0,y=0$). Therefore, two neural networks are needed to fit the displacement above and below the crack. (b) The analytical solution for this problem: $u=r^{\frac{1}{2}} \sin(\theta/2)$, where $r$ is the distance from the coordinate $\boldsymbol{x}$ to the origin $\text{x=y=0}$, and $\theta$ is the angle in the polar coordinate system, with $x>0,y=0$ as the reference. Counterclockwise is considered a positive angle. (c) The grid distribution in KINN, with order=3 and grid size=10, is uniformly distributed in both the x and y directions. (d) The meshless random sampling points for PINN, DEM, and KINN. The red points are essential boundary displacement points (256 points), the blue points are for the upper region neural network (2048 points), the green points are for the lower region neural network (2048 points), and the yellow points are the interface sampling points (1000 points). \label{fig:Crack_info}}
\end{figure}

Mode III crack as shown in \Cref{fig:Crack_info}a is a benchmark problem used to verify algorithms for solving singular problems \cite{wang2022cenn}. The PDEs for the mode III crack are as follows:
\begin{equation}
\begin{cases}
	\Delta(u(\boldsymbol{x}))=0 & \boldsymbol{x} \in \Omega\\
	\bar{u}(\boldsymbol{x})=r^{\frac{1}{2}} \sin(\frac{1}{2} \theta) & \boldsymbol{x} \in \Gamma
\end{cases}.
\end{equation}
where \(\Delta\) is the Laplacian operator, \(\boldsymbol{x}\) is the coordinate point, and \(\Omega\) and \(\Gamma\) represent the square domain ($[-1,1]^{2}$) and the boundary, respectively. The analytical solution \(r^{\frac{1}{2}} \sin(\theta/2)\) is applied as the boundary condition. Here, \(r\) is the distance of  \(\boldsymbol{x}\) from the origin point (0,0), and \(\theta\) is the angle in the polar coordinate system, with \(x>0, y=0\) as the reference.  Counterclockwise is considered a positive angle. The range of $\theta$ is $[-\pi, +\pi]$, which causes two different values of $u$ to appear at the same coordinate at the crack ($x<0, y=0$). Therefore, this problem is strongly discontinuous, shown in \Cref{fig:Crack_info}b.

To demonstrate the performance of KINN, we compare the results of KINN with different solving formats, including traditional strong form by PINNs, energy form by DEM, and inverse form by BINN. The comparison is made based on the accuracy and efficiency of solving the same problem. Since the displacement field is discontinuous at the crack, we use CPINNs \cite{CPINN} for the strong form of PINNs.
Displacement continuity and displacement derivative continuity conditions are incorporated into CPINN.
 The loss function for CPINNs is given by:
\begin{equation}
	\begin{split}
		\mathcal{L}_{cpinns} &= \lambda_{1} \sum_{i=1}^{N_{pde+}} |\Delta(u^{+}(\boldsymbol{x}_{i}))|^{2} + \lambda_{2} \sum_{i=1}^{N_{pde-}} |\Delta(u^{-}(\boldsymbol{x}_{i}))|^{2} + \lambda_{3} \sum_{i=1}^{N_{b+}} |u^{+}(\boldsymbol{x}_{i}) - \bar{u}(\boldsymbol{x}_{i})|^{2} + \lambda_{4} \sum_{i=1}^{N_{b-}} |u^{-}(\boldsymbol{x}_{i}) - \bar{u}(\boldsymbol{x}_{i})|^{2} \\
		&+ \lambda_{5} \sum_{i=1}^{N_{I}} |u^{-}(\boldsymbol{x}_{i}) - u^{+}(\boldsymbol{x}_{i})|^{2}  + \lambda_{6} \sum_{i=1}^{N_{I}} |\boldsymbol{n} \cdot (\nabla u^{+}(\boldsymbol{x}_{i}) - \nabla u^{-}(\boldsymbol{x}_{i}))|^{2}
	\end{split} \label{eq:cpinns_loss}
\end{equation}
where \(\{\lambda_{i}\}_{i=1}^{6}=\{1,1,50,50,10,10\}\) are the hyperparameters for CPINN. \(N_{pde+}\) and \(N_{pde-}\) are the total number of sampled points in the upper and lower domains, respectively. \(N_{b+}\) and \(N_{b-}\) are the total number of sampled points on the upper and lower boundaries, respectively. \(N_{I}\) is the total number of interface points ($x>0,y=0$). \Cref{fig:Crack_info}d shows the sampling method for CPINNs and DEM is the same, with 4096 points in the domain (upper and lower domains combined), 256 points on each boundary, and 1000 points on the interface. We use an MLP fully connected neural network with a structure of [2,30,30,30,30,1], optimized with Adam and a learning rate of 0.001. CPINN divides the region into two areas along the crack: \(y \geq 0\) and \(y \leq 0\). \(u^{+}\) and \(u^{-}\) represent different neural networks for the upper and lower regions, respectively, as shown in  \Cref{fig:Crack_info}. Each neural network has 2911 trainable parameters, and both regions use the same neural network structure and optimization method.

The loss function for DEM is:
\begin{equation}
\mathcal{L}_{DEM} = \int_{\Omega} \frac{1}{2} (\nabla u) \cdot (\nabla u) d\Omega.
\end{equation}
Since DEM is based on the principle of minimum potential energy, \(u\) must satisfy the essential boundary condition in advance:
\begin{equation}
u(x) = u_{p}(\boldsymbol{x}; \boldsymbol{\theta}_{p}) + RBF(\boldsymbol{x}) \cdot u_{g}(\boldsymbol{x}; \boldsymbol{\theta}_{g}),
\end{equation}
where \(u_{p}\) represents the particular network, and \(\boldsymbol{\theta}_{p}\) denotes its parameters. The term RBF refers to the radial basis function, while \(u_{g}\) signifies the generalized network, with \(\boldsymbol{\theta}_{g}\) being its parameters. The RBF points are uniformly distributed with an 11x11 grid, using Gaussian kernels:
\begin{equation}
RBF(\boldsymbol{x}) = \sum_{i=1}^{121} w_{i} \exp(-\gamma |\boldsymbol{x} - \boldsymbol{x}_{i}|)
\end{equation}
where \(\gamma\) is a hyperparameter, set to 0.5, \(|\boldsymbol{x} - \boldsymbol{x}_{i}|\) is the distance between \(\boldsymbol{x}\) and \(\boldsymbol{x}_{i}\), and \(w_{i}\) is the trainable parameters of RBF. During the training process, we keep the parameters of the particular network and RBF fixed and only train the generalized neural network. It is important to highlight that the RBF equals zero at the essential boundary, and only the particular network is active at the essential boundary. The details of DEM are in \Cref{subsec:Introduction_DEM}. All network configurations and optimization methods are the same as CPINNs, with 2911 trainable parameters.

The crack problem involving a strong discontinuity across the crack surface is not feasible for BINN without using multiple networks. For simplicity, the upper-half crack plane (\(y>0\)) is considered with full Dirichlet boundary conditions in BINN.  To calculate the boundary integrals, a strategy demonstrated in our previous work \cite{sun2023binn} is employed in BINN. In this approach, the Gauss quadrature rule is adopted piecewise for the integrals without singularity, while regularization techniques are implemented to calculate the singular integrals. In this example, 80 source points with 800 integration points are allocated uniformly along the boundary. Then BINN is applied to solve the problem using MLP and KAN, respectively. The structures of MLP and KAN in KINN are [2,30,30,30,1] and  [2,5,5,5, 1] respectively. The details of BINN are in \Cref{subsec:Introduction_DEM}.

\begin{figure}
	\begin{centering}
		\includegraphics{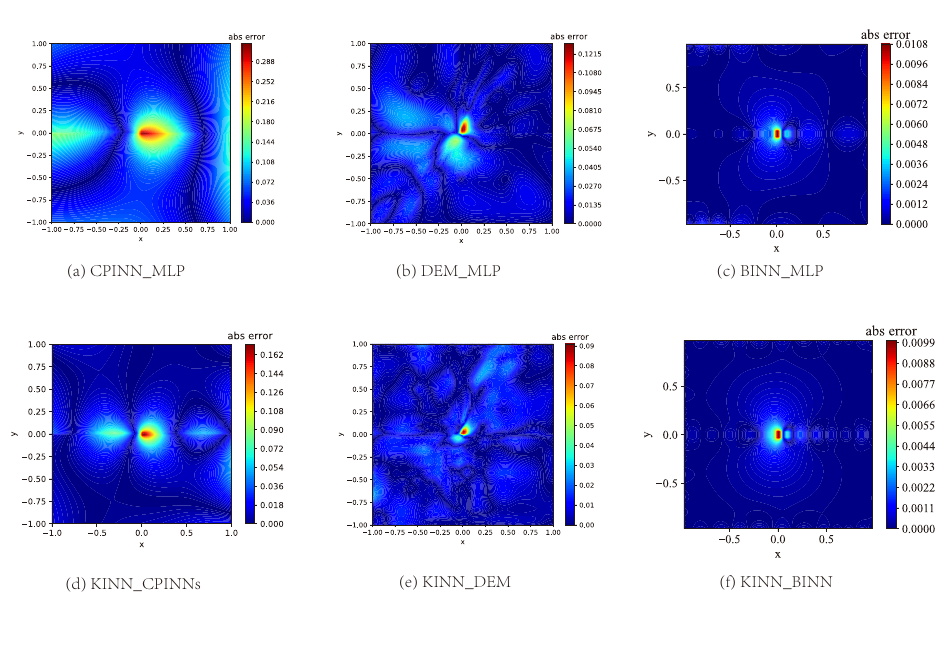}
		\par\end{centering}
	\caption{The absolute error by CPINN, DEM, BINN, and KINN in mode III crack: CPINN\_MLP, DEM\_MLP, BINN\_MLP respectively (first row), KINN\_CPINN, KINN\_DEM, KINN\_BINN respectively (second row). \label{fig:crack_abs_error}}
\end{figure}

The structure of KAN in KINN is [2,5,5,1], with only 600 parameters. We replace the MLP neural network in the above PINNs, DEM with KAN, forming the corresponding KINN versions (KAN in BINN is [2,5,5,5,1]).  \Cref{fig:crack_abs_error} shows the absolute error contour plots for CPINN, DEM, BINN, and KINN. The results indicate that KINN improves accuracy in terms of error distribution and maximum absolute error compared to various MLP-based numerical algorithms. Additionally, we quantify the accuracy and efficiency of KINN and traditional MLP algorithms.  \Cref{tab:The-comparison-crack} shows that with fewer trainable parameters, KINN achieves lower relative \(\mathcal{L}_{2}\) error:
\begin{equation}
\phi_{\mathcal{L}_{2}} = \frac{\int_{\Omega} |\phi_{pred} - \phi_{exact}|^{2} d\Omega}{\int_{\Omega} |\phi_{exact}|^{2} d\Omega}.
\end{equation}

The error for BINN is the lowest, primarily because BINN directly uses the fundamental solution of the problem as the weight function, which is quite close to the analytical solution of the crack. Additionally, BINN directly imposes Dirichlet boundary conditions at the interface (\(x>0, y=0\)), and uses some technology for the numerical integration in \Cref{eq:BINN_loss_funciton}. DEM and PINN, compared to BINN, implicitly contain more information about the equations. Although KINN has fewer trainable parameters, its computation time is currently about three times longer than the corresponding MLP version over 1000 epochs. This is because KAN is implemented by ourselves and does not have the optimized AI libraries like MLP, which have undergone extensive optimization. This indicates that there is still room for optimization in KAN within KINN, but it also highlights the great potential of KAN in solving PDE problems. This is because solving PDE problems essentially does not require an excessive number of parameters for simulation. Unlike fields such as computer vision and natural language processing, which require complex neural networks, solving PDEs is more akin to function fitting. The difference is that PDE solving does not provide the function directly but implicitly gives the function to be simulated through the mathematical equations of PDEs. Most function-fitting problems are not as complex as AI tasks in computer vision, so overly complex neural networks are not necessary. Therefore, the final convergence accuracy of KINN is higher than that of the MLP.

\begin{table}
	\caption{The comparison between different PINNs, DEM, and BINN algorithms based on MLP or KAN. Parameters mean the trained parameters in the architecture of NNs. Relative error represents the \(\mathcal{L}_{2}\) error at convergence. Grid size is the number of grids in the KAN network. Grid range is the initial range of the KAN grid. Order is the order of the B-splines in KAN. The architecture of NNs is the structure of the corresponding neural network. Parameters are the trainable parameters of the corresponding network. The time is the duration for 1000 epochs of the corresponding algorithms.\label{tab:The-comparison-crack}}
	\begin{adjustbox}{max width=\textwidth}
	\centering{}%
	\begin{tabular}{cccccccc}
		\toprule 
		\multirow{1}{*}{Algorithms} & Relative error & Grid size & Grid range & Order & Architecture of NNs & Parameters & Time (Second, 1000 Epoch)\tabularnewline
		\midrule 
		CPINNs\_MLP & 0.062535 &  &  &  & {[}2, 30,30,30,30,1{]} & 2911 & 11.15\tabularnewline
		KINN\_CPINNs & 0.029879 & 10 & {[}-1,1{]} & 3 & {[}2, 5,5, 1{]} & 600 & 39.94\tabularnewline
		DEM\_MLP & 0.024292 &  &  &  & {[}2, 30,30,30,30,1{]} & 2911 & 8.77\tabularnewline
		KINN\_DEM & 0.011166 & 10 & {[}-1,1{]} & 3 & {[}2, 5,5, 1{]} & 600 & 17.04\tabularnewline
		BINN\_MLP & 0.000872 &  &  &  & {[}2, 30,30,30,30,1{]} & 2911 & 3.23\tabularnewline
		KINN\_BINN & 0.000828 & 10 & {[}-1,1{]} & 3 & {[}2, 5,5, 5,1{]} & 975 & 14.17\tabularnewline
		\bottomrule
	\end{tabular}
	\end{adjustbox}
\end{table}

\begin{figure}
	\begin{centering}
		\includegraphics{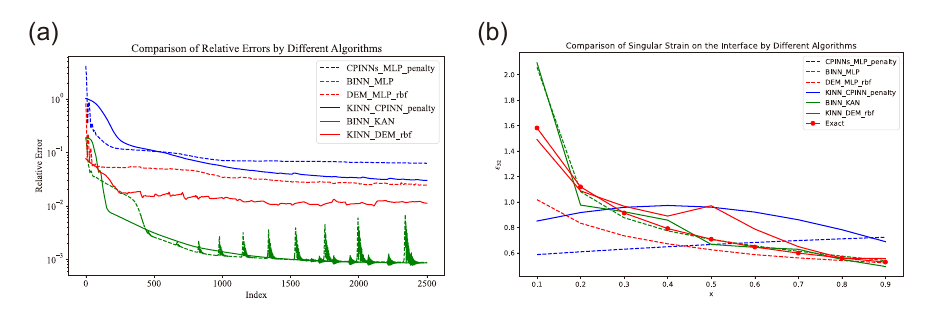}
		\par\end{centering}
	\caption{The error evolution and singular displacement derivative at the interface for KINN and corresponding MLP algorithms in mode III crack: (a) Evolution of the relative \(\mathcal{L}_{2}\) error. (b) Comparison of the singular displacement derivative \(\partial u/\partial y\) (\(\varepsilon_{z\theta}\)) at \(x>0, y=0\).
Penalty refers to the boundary conditions of the PDEs applied through the penalty function, as shown in the terms corresponding to	$\lambda_{3},\lambda_{4},\lambda_{5},\lambda_{6}$ in \Cref{eq:cpinns_loss}.
		\label{fig:crack_error_evolution}}
\end{figure}

To more clearly show the training process of KINN and the corresponding algorithms,  \Cref{fig:crack_error_evolution}a shows the evolution of the relative \(\mathcal{L}_{2}\) error with the number of iterations. We can see that KINN converges faster and has higher accuracy compared to the corresponding MLP versions. The singularity at the crack tip is the most important characteristic of the crack problem. Considering the strain:
\begin{equation}
\varepsilon_{z\theta}|_{interface} = \frac{1}{r} \frac{\partial u}{\partial \theta} = \frac{1}{2 \sqrt{r}} \cos(\frac{\theta}{2})|_{\theta=0} = \frac{1}{2 \sqrt{r}},
\end{equation}
where the interface is the line \(x>0, y=0\). $\theta$ is from $[-\pi, \pi]$ shown in \Cref{fig:crack_error_evolution}a. We compared PINNs, DEM, BINN, and the corresponding KINN versions. \Cref{fig:crack_error_evolution}b shows that KINN-DEM performs the best, but there is a jump in the result at \(x=0.5\). Note that the results of the six algorithms are all uniformly sampled.

\begin{figure}
	\begin{centering}
		\includegraphics{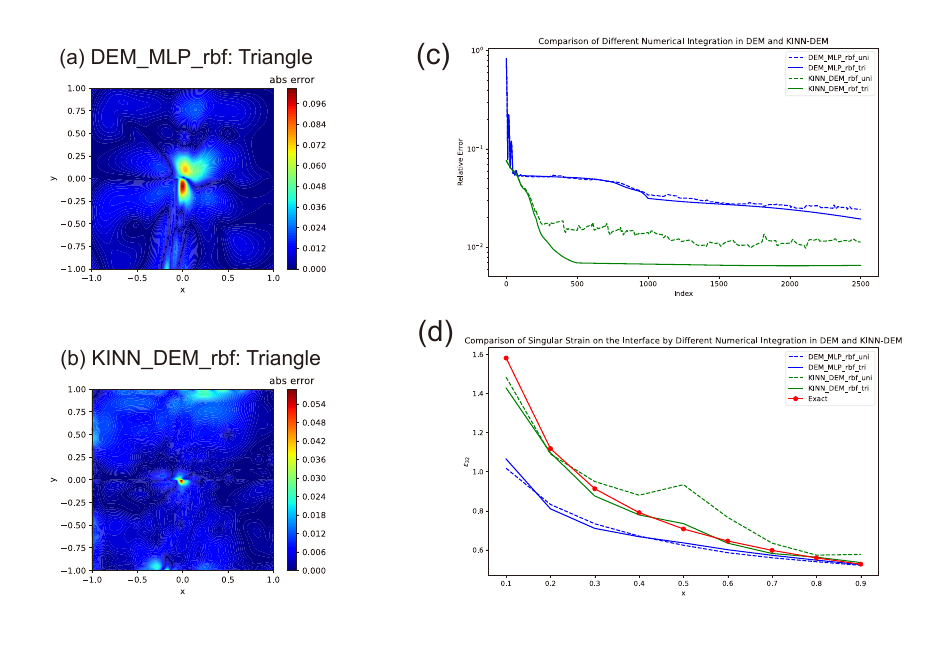}
		\par\end{centering}
	\caption{Comparison of uniform Monte Carlo integration and triangular integration in KINN-DEM and KINN for mode III crack: (a) Absolute error contour plot for DEM with uniform Monte Carlo integration. (b) Absolute error contour plot for KINN-DEM with triangular grid integration. (c) Evolution of relative \(\mathcal{L}_{2}\) error for DEM with Monte Carlo and triangular integration. (d) Singular displacement derivative with Monte Carlo and triangular integration at the interface.\label{fig:crack_integration}}
\end{figure}

Since the numerical integration scheme can affect the accuracy of DEM \cite{PINN_hyperelasticity}, we changed the integration scheme from Monte Carlo integration to triangular numerical integration, using 79,202 triangular integration points. Note that the previous results were all uniformly sampled and used Monte Carlo integration to approximate the loss function in DEM. Triangular integration involves dividing the domain into triangular grids, finding the centroid of each triangle, and calculating the numerical integration based on the area of the triangles:
\begin{equation}
\int_{\Omega} f(\boldsymbol{x}) d\Omega = \sum_{i=1}^{N} f(\boldsymbol{x}_{i}) S_{i},
\end{equation}
where \(N\) is the total number of triangular grids, \(\boldsymbol{x}_{i}\) is the centroid of the \(i\)-th triangular grid, and \(S_{i}\) is the area of the corresponding triangular grid. Note that dividing the grid causes DEM to lose some meshless advantages, but the adaptability to complex geometries and simplicity of triangular grids make them easy to implement in DEM. \Cref{fig:crack_integration} compares the triangular grid and uniform Monte Carlo integration. We can see that the absolute error contour plot for the triangular grid has smaller errors compared to Monte Carlo integration.  Most importantly, we find that the triangular grid in DEM fits the singularity problem very well, especially in the KINN-DEM form, where the triangular grid almost perfectly matches the analytical solution.

\begin{figure}
	\begin{centering}
		\includegraphics{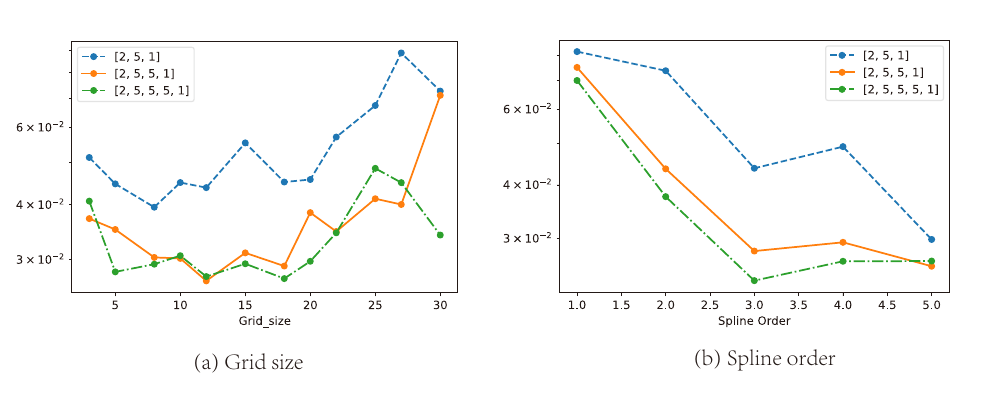}
		\par\end{centering}
	\caption{Relative \(\mathcal{L}_{2}\) error for KINN-CPINNs in mode III crack in different architectures of KAN  with different grid sizes (a) and different B-spline orders (b).\label{fig:crack_scaling_law}}
\end{figure}

Considering that KINN has some hyperparameters, such as grid size, B-spline order, and the depth and width of the KAN network, we study the impact of these hyperparameters on KINN. We use KINN-CPINNs here because the strong form by PINNs can solve almost any PDEs, making them more versatile than the energy form by DEM and the inverse form by BINN.  \Cref{fig:crack_scaling_law} shows that with different grid sizes, the relative \(\mathcal{L}_{2}\) error in KINN-CPINNs reaches a minimum and then increases as the grid size continues to increase. This is consistent with the conclusions in the original KAN paper \citet{liu2024kan}. Excessive grid size leads to overfitting and an unsmooth fitted function. The grid size should be proportional to the complexity of the fitting function in PDE problems, meaning that the grid size should increase with the complexity of the function. Additionally, we fixed the grid size at 10 and adjusted the order of the B-spline. We found that increasing the spline order reaches a plateau, where further increases do not improve model accuracy. This is similar to the element order in FEM. Unfortunately, KINN does not currently exhibit grid convergence like FEM, where the grid size can be increased indefinitely. We believe that exploring this aspect further is an important direction for future research on KINN.

\subsubsection{Stress concentration}

\begin{figure}
	\begin{centering}
		\includegraphics{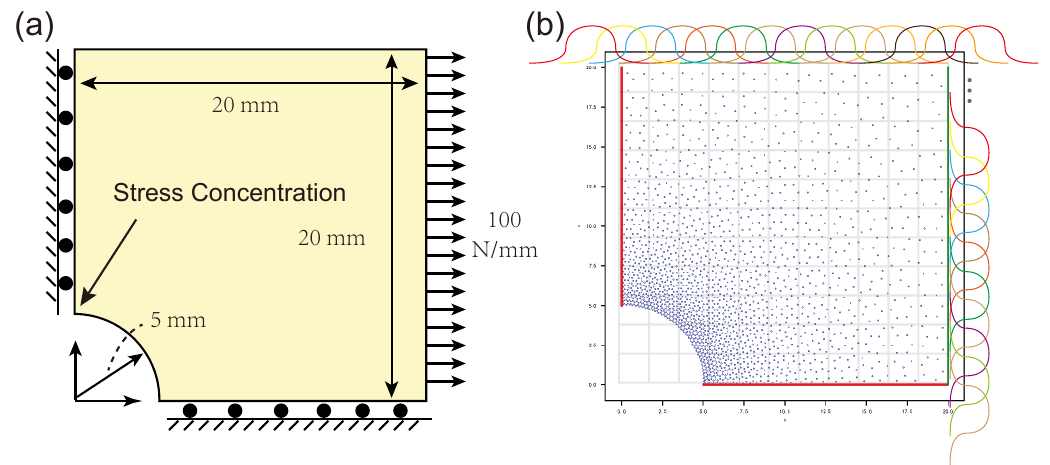}
		\par\end{centering}
	\caption{The introduction of plate with central hole: (a) The geometry of the plate with a central hole, with a hole radius of 5 mm and a square plate with an edge length of 20 mm. The right side is subjected to a uniform load of 100 N/mm. (b) The grid distribution in KINN, with order=3 and grid size=10, is uniformly distributed in both the x and y directions. \label{fig:plate_hole_intro}}
\end{figure}

The plate with a central hole is a commonly used benchmark in solid mechanics, as shown in  \Cref{fig:plate_hole_intro}. This problem is not only common in engineering but also features stress concentration near the hole. The geometry of the plate with a central hole has an original size of 40x40 mm, with a hole at the center and a radius of 5 mm. The left and right sides are subjected to a force of \(t_{x}=100 \, \text{N/mm}\), with an elastic modulus \(E=1000 \, \text{MPa}\) and a Poisson's ratio \(\nu=0.3\). Due to symmetry, we simulate only the upper right quarter, with an edge length \(L=20 \, \text{mm}\). The boundary conditions are \(u_{x}=0\) at \(x=0\) and \(u_{y}=0\) at \(y=0\), with other boundaries subjected to force boundary conditions (free boundary conditions). We use a plane stress model, as shown in  \Cref{fig:plate_hole_intro}a.

As in the previous example, we use the strong form by PINNs, energy form by DEM, and inverse form by BINN to solve this problem, and then compare the results by MLP with the corresponding versions of KINN. For PINNs and DEM, we use the same collocation scheme (uniform distribution) with 75,298 points in the domain and 1,000 points on each boundary (a total of 5 boundaries). The loss function for PINNs is:
\begin{equation}
	\begin{split}
		\mathcal{L}_{pinns} &= \lambda_{1} \sum_{i=1}^{N_{pde}} \left[ |\sigma(\boldsymbol{x}_{i})_{xx,x} + \sigma(\boldsymbol{x}_{i})_{xy,y}|^{2} + |\sigma(\boldsymbol{x}_{i})_{yx,x} + \sigma(\boldsymbol{x}_{i})_{yy,y}|^{2} \right] \\
		&+ \lambda_{2} \sum_{i=1}^{N_{left}} |u_{x}(\boldsymbol{x}_{i})|^{2} + \lambda_{3} \sum_{i=1}^{N_{left}} |\sigma_{xy}(\boldsymbol{x}_{i})|^{2} \\
		&+ \lambda_{4} \sum_{i=1}^{N_{down}} |u_{y}(\boldsymbol{x}_{i})|^{2} + \lambda_{5} \sum_{i=1}^{N_{down}} |\sigma_{xy}(\boldsymbol{x}_{i})|^{2} \\
		&+ \lambda_{6} \sum_{i=1}^{N_{up}} \left[ |\sigma_{yy}(\boldsymbol{x}_{i})|^{2} + |\sigma_{xy}(\boldsymbol{x}_{i})|^{2} \right] \\
		&+ \lambda_{7} \sum_{i=1}^{N_{right}} \left[ |\sigma_{xx}(\boldsymbol{x}_{i}) - t_{x}|^{2} + |\sigma_{xy}(\boldsymbol{x}_{i})|^{2} \right] \\
		&+ \lambda_{8} \sum_{i=1}^{N_{circle}} \left[ |\sigma_{xx}(\boldsymbol{x}_{i}) n_{x} + \sigma_{xy}(\boldsymbol{x}_{i}) n_{y}|^{2} + |\sigma_{yx}(\boldsymbol{x}_{i}) n_{x} + \sigma_{yy}(\boldsymbol{x}_{i}) n_{y}|^{2} \right],
	\end{split}
\end{equation}
\begin{equation}
\sigma_{ij}  =\frac{E}{1+\upsilon}\varepsilon_{ij}+\frac{E\upsilon}{(1+\upsilon)(1-2\upsilon)}\varepsilon_{kk}\delta_{ij}
\end{equation}
\begin{equation}
\varepsilon_{ij} = \frac{1}{2} (u_{i,j} + u_{j,i})
\end{equation}
We can observe that the loss function of PINNs in the strong form is very complex and has numerous hyperparameters. Although there are some techniques for adjusting hyperparameters \citet{ill_gradient,NTK_PINN,NTK_to_get_hyperparameter_of_PINN}, it is a challenge to get the optimal hyperparameters. Therefore, we manually set the hyperparameters $\{\lambda_{i}\}_{i=1}^{8}=\{10,2000,1,1,2000,1,1,1\}$. We use a neural network to approximate the displacement field:
\begin{equation}
	\boldsymbol{u}(\boldsymbol{x}) \approx \boldsymbol{u} \left( \frac{\boldsymbol{x}}{L}; \boldsymbol{\theta} \right),
\end{equation}
where \(\boldsymbol{\theta}\) are the neural network parameters.

For DEM, the loss function is:
\begin{equation}
	\begin{split}
		\mathcal{L}_{DEM} &= \int_{\Omega} \Psi \, d\Omega - \int_{\Gamma^{right}} t_{x} u_{x} \, d\Gamma, \\
		\Psi &= \frac{1}{2} \sigma_{ij} \varepsilon_{ij}, \\
		u_{x} &= x \cdot u_{x} \left( \frac{\boldsymbol{x}}{L}; \boldsymbol{\theta} \right), \\
		u_{y} &= y \cdot u_{y} \left( \frac{\boldsymbol{x}}{L}; \boldsymbol{\theta} \right),
	\end{split}
\end{equation}
We can see that DEM has no hyperparameters in the loss function and involves lower-order derivatives than strong form by PINNs, but it requires the displacement field to satisfy essential boundary conditions in advance. Here, we construct the admissible displacement field by multiplying the coordinates. Additionally, DEM requires the integration of internal energy and external work. As shown in  \Cref{fig:crack_integration}, triangular integration is more accurate for DEM than Monte Carlo integration, so we use triangular grid integration for this example instead of Monte Carlo integration in DEM.

\begin{figure}
	\begin{centering}
		\includegraphics[scale=0.9]{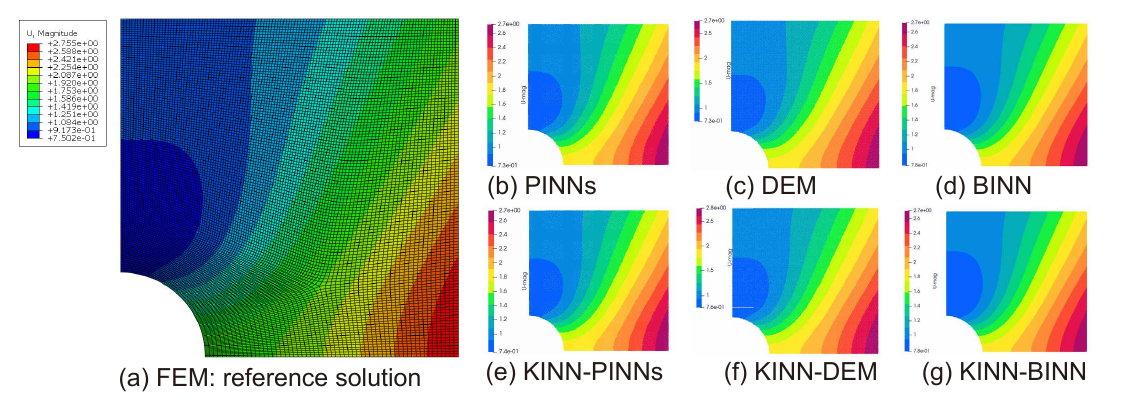}
		\par\end{centering}
	\caption{Predicted displacement solutions for PINNs, DEM, BINN, and their corresponding KINN versions: (a) FEM reference solution, (b) PINNs, (c) DEM, (d) BINN, (e) KINN-PINNs, (f) KINN-DEM, (g) KINN-BINN. \label{fig:U_mag_prediction_plate_hole}}
\end{figure}

\Cref{fig:U_mag_prediction_plate_hole} shows the predicted absolute displacement (\(u_{mag} = \sqrt{u_{x}^{2} + u_{y}^{2}}\)) for  PINNs, DEM, BINN, and the corresponding KINN versions. FEM is used as the reference solution. The FEM mesh is shown in \Cref{fig:U_mag_prediction_plate_hole}a. It uses 15,300 eight-node quadratic elements with reduced integration (CPS8R). We performed a mesh convergence analysis for the FEM solution, so it is reliable. The FEM computation time is 30.64s. Our numerical experiments demonstrated that using scale normalization techniques in \ref{sec:scale_normal} allows us to solve problems with a scale significantly greater than 1 (e.g., a plate with a hole having a characteristic size of 20).

\begin{figure}
	\begin{centering}
		\includegraphics{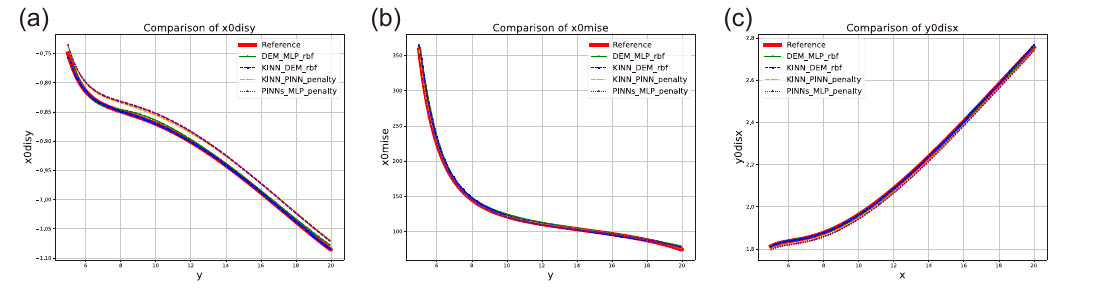}
		\par\end{centering}
	\caption{Displacement and Mises stress along \(x=0\) and \(y=0\) for PINNs, DEM, BINN, and their corresponding KINN versions: (a) \(u_{y}\) along \(x=0\), (b) Mises stress along \(x=0\), (c) \(u_{x}\) along \(y=0\). \label{fig:Plate_hole_edge}}
\end{figure}

\begin{figure}
	\begin{centering}
		\includegraphics[scale=1.1]{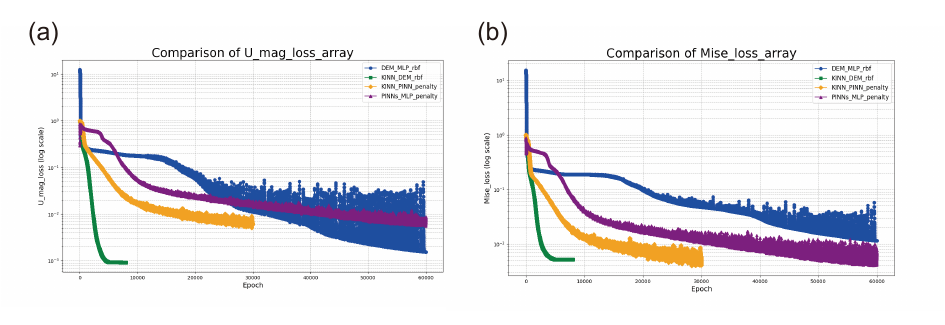}
		\par\end{centering}
	\caption{Evolution of the relative \(\mathcal{L}_{2}\) and \(\mathcal{H}_{1}\) errors for PINNs, DEM, BINN, and their corresponding KINN versions: (a) Evolution of the relative \(\mathcal{L}_{2}\) displacement error, (b) Evolution of the relative \(\mathcal{H}_{1}\) error in Mises stress. \label{fig:plate_hole_error_evolution}}
\end{figure}

Due to the stress concentration near the hole, traditional FEM requires a denser mesh near the stress concentration to ensure simulation accuracy. Traditionally, PINNs and DEM with MLP often perform poorly near the stress concentration in Monte Carlo integration. Therefore, we replace the MLP in PINNs and DEM with KAN to see if there is any improvement. The plate with a central hole exhibits the maximum \(u_{y}\) and the most significant stress concentration along the line \(x=0\), while the maximum \(u_{x}\) occurs along the line \(y=0\). Thus, we compare the different algorithms in terms of \(u_{y}\) and Mises stress shown in \Cref{eq:Von_mises} along \(x=0\) and \(u_{x}\) along \(y=0\).  \Cref{fig:Plate_hole_edge} shows that KINN-DEM achieves the highest accuracy, perfectly matching the reference solution. \Cref{fig:plate_hole_error_evolution} shows the relative errors in displacement and Mises stress, indicating that KINN-DEM achieves almost the same accuracy as FEM.
The most surprising thing is that the convergence speed of KINN-DEM is extremely fast.

In terms of efficiency, \Cref{tab:plate_hole_comparison} shows the accuracy and efficiency of various algorithms.

\begin{equation}
	Mises=\sqrt{\frac{3}{2}(\sigma_{ij}-\frac{1}{3}\sigma_{kk}\delta_{ij})(\sigma_{ij}-\frac{1}{3}\sigma_{mm}\delta_{ij})}\label{eq:Von_mises}
\end{equation}

\begin{table}
\centering
\caption{Comparison of accuracy and errors for various algorithms in the plate with a central hole \label{tab:plate_hole_comparison}}
\begin{adjustbox}{max width=\textwidth}
\begin{tabular}{ccccccccc}
	\toprule 
	\multirow{1}{*}{Algrithms} & Dis relative error ($\mathcal{L}_{2}$) & Mise relative error ($\mathcal{H}_{1}$)  & Grid size & grid\_range & order & Architecture of NNs & Parameters & Time (Second, 1000 Epoch)\tabularnewline
	\midrule 
	PINNs\_MLP & 0.00858 & 0.00649 &  &  &  & {[}2, 30,30,30,30,2{]} & 2942 & 70.65\tabularnewline
	KINN-PINNs & 0.00629 & 0.00481 & 15 & {[}0,1{]} & 3 & {[}2, 5,5,5 2{]} & 1400 & 1622\tabularnewline
	DEM\_MLP & 0.00154 & 0.0115 &  &  &  & {[}2, 30,30,30,30,2{]} & 2942 & 14.57\tabularnewline
	KINN-DEM & 0.000919 & 0.00517 & 15 & {[}0,1{]} & 3 & {[}2, 5,5,5 2{]} & 1400 & 310\tabularnewline
	BINN-MLP & 0.000545 & 0.00434 &  &  &  & {[}2, 30,30,30,30,2{]} & 2942 & 5.4\tabularnewline
	KINN- BINN & 0.000226 & 0.00211 & 10 & {[}0,20{]} & 3 & {[}2, 5,5,5 2{]} & 1400 & 21.3\tabularnewline
	\bottomrule
\end{tabular}
\end{adjustbox}
\end{table}

\subsubsection{Nonlinear PDEs}

This example aims to verify the effectiveness of KINN in solving nonlinear problems. We consider a hyperelasticity problem \cite{PINN_hyperelasticity}. Hyperelasticity problems are nonlinear due to the nonlinearity of both constitutive law and geometric equations. We consider the Neo-Hookean constitutive model for the hyperelasticity problem:
\begin{equation}
	\begin{split}
		\varPsi & = \frac{1}{2} \lambda (\ln J)^{2} - \mu \ln J + \frac{1}{2} \mu (I_{1} - 3) \\
		J & = \det(\boldsymbol{F}) \\
		I_{1} & = \operatorname{trace}(\boldsymbol{C}) \\
		\boldsymbol{C} & = \boldsymbol{F}^{T} \boldsymbol{F}
	\end{split}
\end{equation}
where \(\varPsi\) is the Neo-Hookean strain energy density function, and \(\lambda\) and \(\mu\) are the Lamé parameters:
\begin{equation}
\begin{cases}
	\lambda = \frac{\nu E}{(1 + \nu)(1 - 2\nu)} \\
	\mu = \frac{E}{2(1 + \nu)}
\end{cases}
\end{equation}
where \(E\) and \(\nu\) are the elastic modulus and Poisson's ratio, respectively. \(\boldsymbol{C}\) is the Green tensor, and \(\boldsymbol{F}\) is the deformation gradient:
\begin{equation}
\boldsymbol{F} = \frac{\partial \boldsymbol{x}}{\partial \boldsymbol{X}}
\end{equation}
where \(\boldsymbol{x}\) is the spatial coordinate, and \(\boldsymbol{X}\) is the material coordinate. The relationship between the spatial and material coordinates is \(\boldsymbol{x} = \boldsymbol{X} + \boldsymbol{u}\), where \(\boldsymbol{u}\) is the displacement field. Since this problem has an energy form, it can be solved using DEM. Notably, solving this problem in the strong form using PINNs is complex and less efficient than DEM, as the solution order of PINNs is higher than that of DEM. For detailed proof, refer to \cite{wang2022cenn}. Therefore, we only consider the energy form by DEM for this problem. The loss function of DEM is:
\begin{equation}
\mathcal{L} = \int_{\Omega} (\varPsi - \boldsymbol{f} \cdot \boldsymbol{u}) d\Omega - \int_{\Gamma^{\boldsymbol{t}}} \bar{\boldsymbol{t}} \cdot \boldsymbol{u} d\Gamma
\end{equation}
where \(\boldsymbol{f}\) is the body force, and \(\bar{\boldsymbol{t}}\) is the surface traction on the boundary \(\Gamma^{\boldsymbol{t}}\).

\begin{figure}
	\begin{centering}
		\includegraphics[scale=0.9]{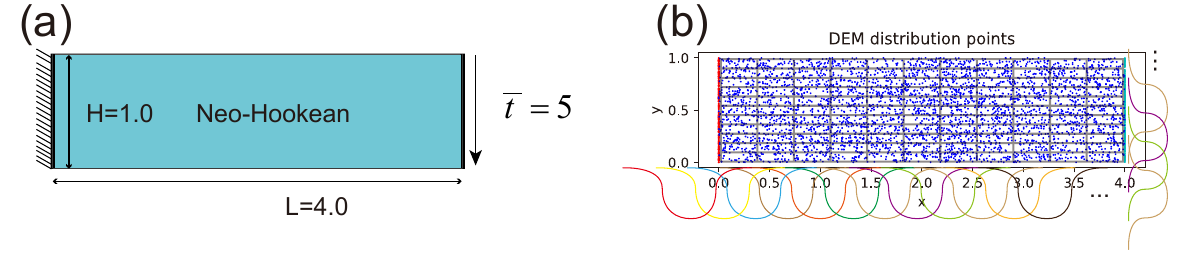}
		\par\end{centering}
	\caption{Description of the Neo-Hookean hyperelastic cantilever beam problem: (a) Dimensions of the cantilever beam, height \(H = 1.0\), length \(L = 4.0\), \(E = 1000\), and \(\nu = 0.3\). The Neo-Hookean hyperelastic constitutive model is used. The left end \(x = 0.0\) is fixed, and a uniformly distributed downward load \(\bar{t} = 5\) is applied at the right end. (b) DEM point distribution: blue points are domain points, red points are essential boundary condition points, and green points are surface traction points.\label{fig:Beam_neo_hookie}}
\end{figure}

We use MLP and KAN networks to solve the benchmark 2D cantilever beam plane stress problem, as shown in \Cref{fig:Beam_neo_hookie}. Since solving this problem in the strong form using PINNs is inconvenient, and BINN does not have a fundamental solution, we only use DEM to solve this problem.

For DEM, we adopt different integration methods, including Monte Carlo, trapezoidal rule, and Simpson’s rule. We compare these integration schemes using KINN and DEM\_MLP. We uniformly distribute 200 points in the length direction and 50 points in the width direction, resulting in 10,000 points in total.

We use the same MLP and KAN configurations as in the previous numerical examples, except for the network types. We use the LBFGS second-order optimization algorithm. \Cref{tab:hyperelasticity_comparison} shows the accuracy and efficiency between MLP and KAN. Our reference solution is obtained using FEM. Due to stress singularities at the corners \((0,0)\) and \((0,1)\), we only compare the \(L_{2}\) displacement error without considering the \(H_{1}\) stress error. We find that KINN improves accuracy under different integration schemes, and DEM requires higher precision in numerical integration. The error of Monte Carlo integration is larger than that of Simpson's and trapezoidal rule. Although Simpson's rule has higher polynomial accuracy than the trapezoidal rule, it does not significantly improve DEM accuracy. Therefore, DEM mainly requires two things: replacing MLP with KAN and using low-order numerical integration such as triangular integration or the trapezoidal rule, without necessarily using higher-order schemes like Simpson's or Gaussian integration. Importantly, DEM should not use Monte Carlo integration due to its low convergence. KAN in DEM does not need higher-order integration because KAN can adapt the local compact space shape to fit the target function, reducing the need for precise numerical integration.  Empirically, DEM with Monte Carlo integration performs poorly. Although KAN has fewer trainable parameters than MLP, KAN takes longer to train for 100 epochs.

\begin{table}
	\caption{Accuracy and error comparison of various algorithms for the hyperelastic problem\label{tab:hyperelasticity_comparison}}
	
	\centering{}%
	\begin{adjustbox}{max width=\textwidth}
	\begin{tabular}{cccccccc}
		\toprule 
		\multirow{1}{*}{Algorithms} & Displacement Relative Error ($\mathcal{L}_{2}$) & Grid Size & Grid Range & Order & NN Architecture & Parameters & Time (Seconds, 100 Epochs)\tabularnewline
		\midrule 
		DEM\_MLP\_Mont & 0.04756 &  &  &  & {[}2, 30,30,30,30,2{]} & 2942 & 37.34\tabularnewline
		KINN-DEM\_Mont & 0.03334 & 15 & {[}0,1{]} & 3 & {[}2, 5,5,5,2{]} & 1400 & 96.44\tabularnewline
		DEM\_MLP\_Trap & 0.01524 &  &  &  & {[}2, 30,30,30,30,2{]} & 2942 & 35.76\tabularnewline
		KINN-DEM\_Trap & 0.002536 & 15 & {[}0,1{]} & 3 & {[}2, 5,5,5,2{]} & 1400 & 99.01\tabularnewline
		DEM\_MLP\_Simp & 0.01753 &  &  &  & {[}2, 30,30,30,30,2{]} & 2942 & 39.87\tabularnewline
		KINN-DEM\_Simp & 0.002181 & 15 & {[}0,1{]} & 3 & {[}2, 5,5,5,2{]} & 1400 & 104.4\tabularnewline
		\bottomrule
	\end{tabular}
	\end{adjustbox}
\end{table}

\begin{figure}
	\begin{centering}
		\includegraphics[scale=0.85]{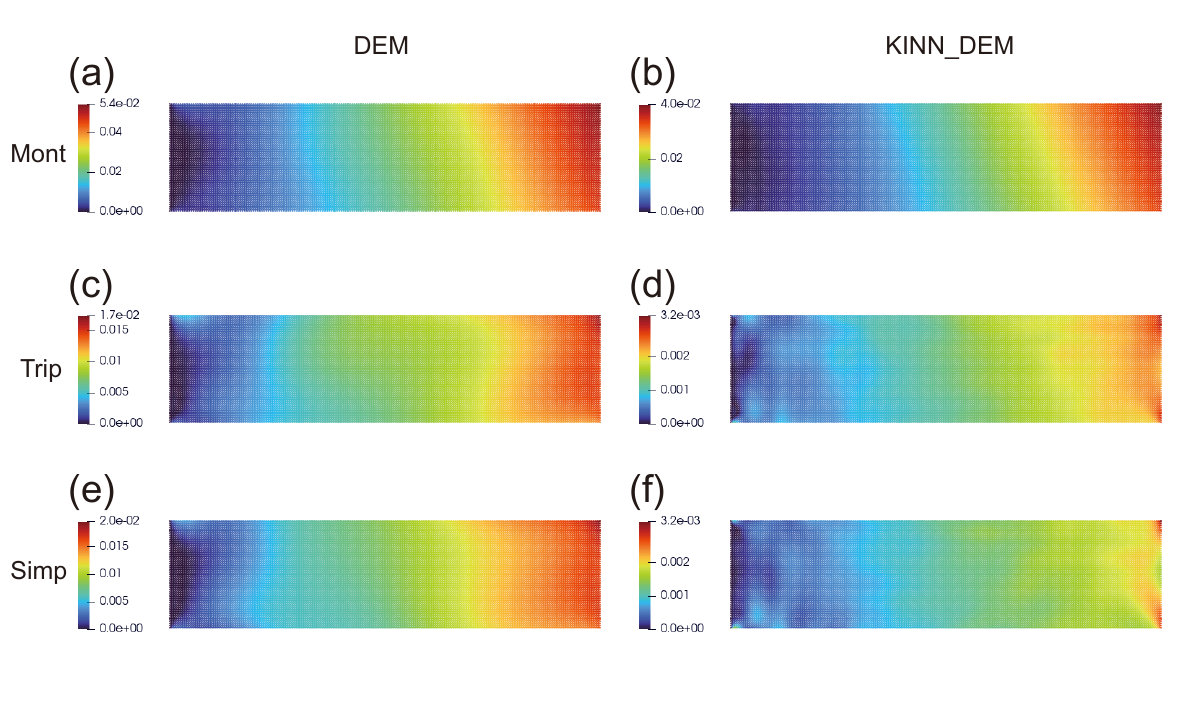}
		\par\end{centering}
	\caption{Absolute error contours of the Neo-Hookean hyperelastic cantilever beam: (a,c,e) Absolute error contours for DEM with MLP, (b,d,f) Absolute error contours for DEM with KAN. (a, b), (c,d), and (e,f) correspond to Monte Carlo, trapezoidal, and Simpson’s numerical integrations, respectively.\label{fig:hyper_abs_error}}
\end{figure}

\begin{figure}
	\begin{centering}
		\includegraphics[scale=0.5]{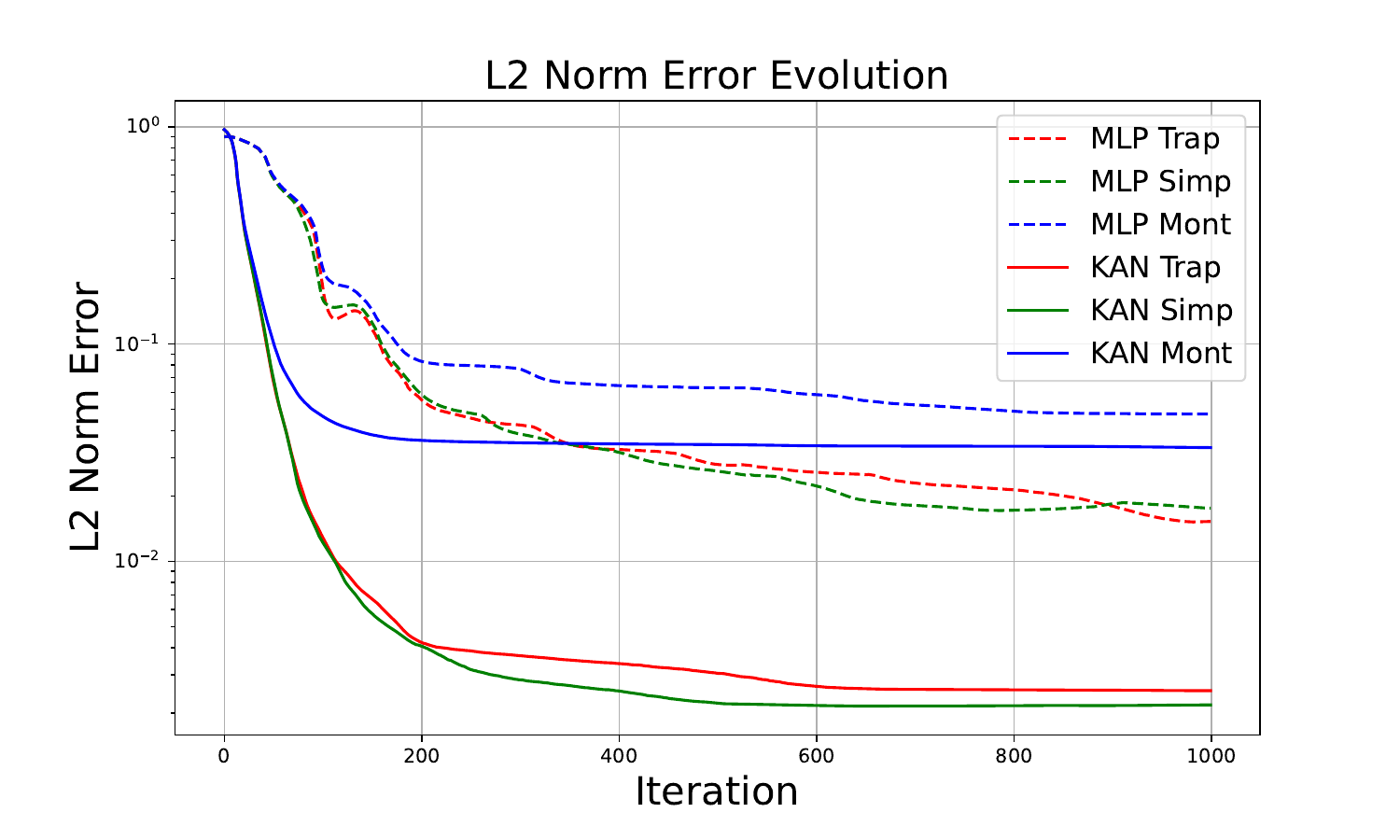}
		\par\end{centering}
	\caption{Evolution of the \(L_{2}\) relative error for the Neo-Hookean hyperelastic cantilever beam.\label{fig:hyper_evolution}}
\end{figure}

The y-direction displacement \(u_{y}\) is the most critical field function under the loading condition in the cantilever beam problem. \Cref{fig:hyper_abs_error} shows the absolute error contour plots of \(u_{y}\) for MLP and KAN in DEM. In any numerical integration method, KINN significantly reduces the maximum error and error distribution compared to MLP. \Cref{fig:hyper_evolution} shows the evolution of the \(L_{2}\) relative error, considering the absolute displacement \(u_{mag} = \sqrt{u_{x}^{2} + u_{y}^{2}}\).

\begin{figure}
	\begin{centering}
		\includegraphics[scale=0.7]{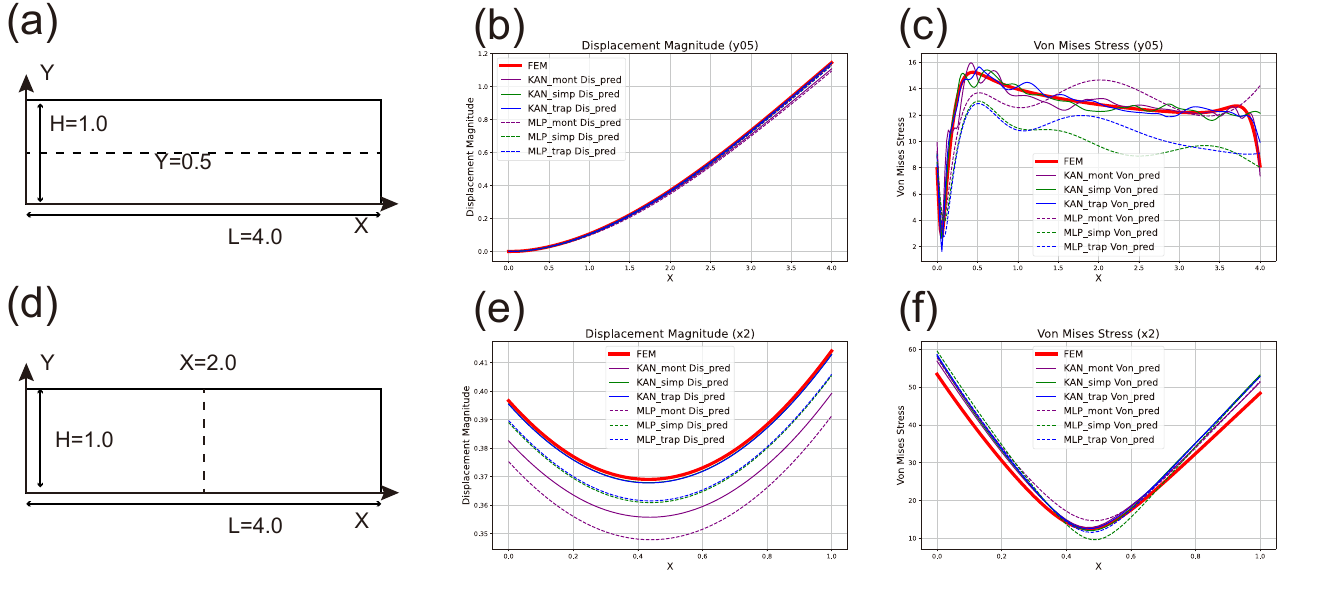}
		\par\end{centering}
	\caption{Absolute displacement and Von Mises stress at \(x = 2\) and \(y = 0.5\) for the Neo-Hookean hyperelastic cantilever beam. The first row compares the absolute displacement (a) and Von Mises stress (b) along \(y = 0.5\). The second row compares the absolute displacement (d) and Von Mises stress (e) along \(x = 2.0\).\label{fig:hyper_line}}
\end{figure}

Despite the stress singularity at the corners, we compare the absolute displacement and Von Mises stress along \(x = 2\) and \(y = 0.5\). \Cref{fig:hyper_line} shows the predictions of absolute displacement and Von Mises stress for KINN and DEM\_MLP under the three numerical integration methods. We observe that replacing MLP with KAN improves accuracy significantly for all numerical integrations.

\subsubsection{Heterogeneous problem}

\begin{figure}
	\begin{centering}
		\includegraphics[scale=0.7]{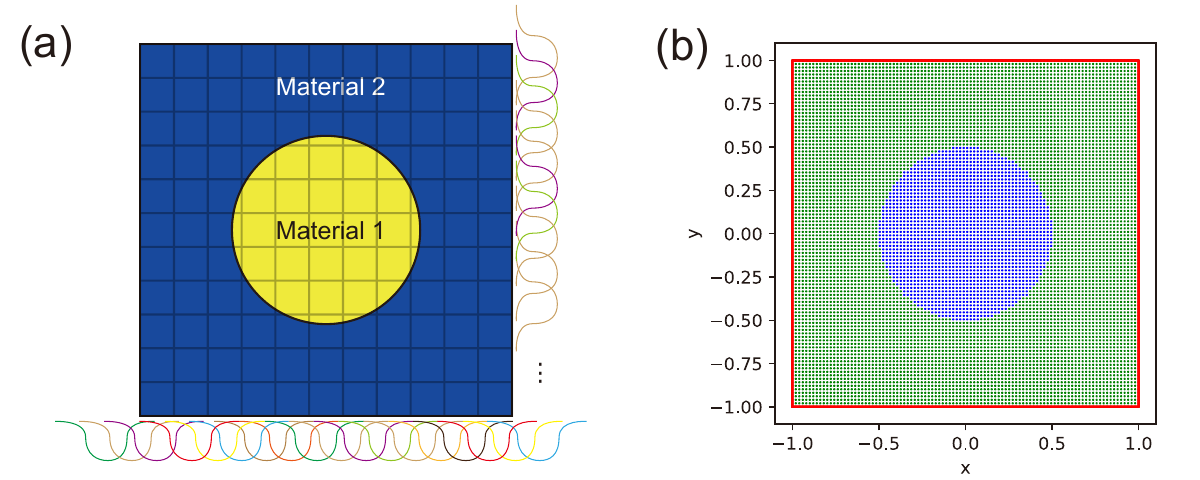}
		\par\end{centering}
	\caption{Illustration of the heterogeneous problem: (a) Two different materials, with continuous displacement but discontinuous strain at the interface. (b) Point sampling illustration, with blue and green points in different regions and red points at the Dirichlet boundary condition.\label{fig:heterogeneous_intro}}
\end{figure}

Heterogeneous problems are commonly encountered in computational mechanics, characterized by the combination of materials with different properties, as shown in \Cref{fig:heterogeneous_intro}a.  The section demonstrates whether KAN has an advantage over MLP in solving problems with materials of different properties.

The motivation for using KAN to solve heterogeneous materials is that KAN is based on spline interpolation, and splines are piecewise functions. On the other hand, since the heterogeneous problem is mathematically continuous in the original function but discontinuous in the derivative, KAN's piecewise function property theoretically has a natural advantage in solving such problems. Therefore, in this section, we experiment with KAN to solve heterogeneous material problems.

\begin{figure}
	\begin{centering}
		\includegraphics[scale=0.55]{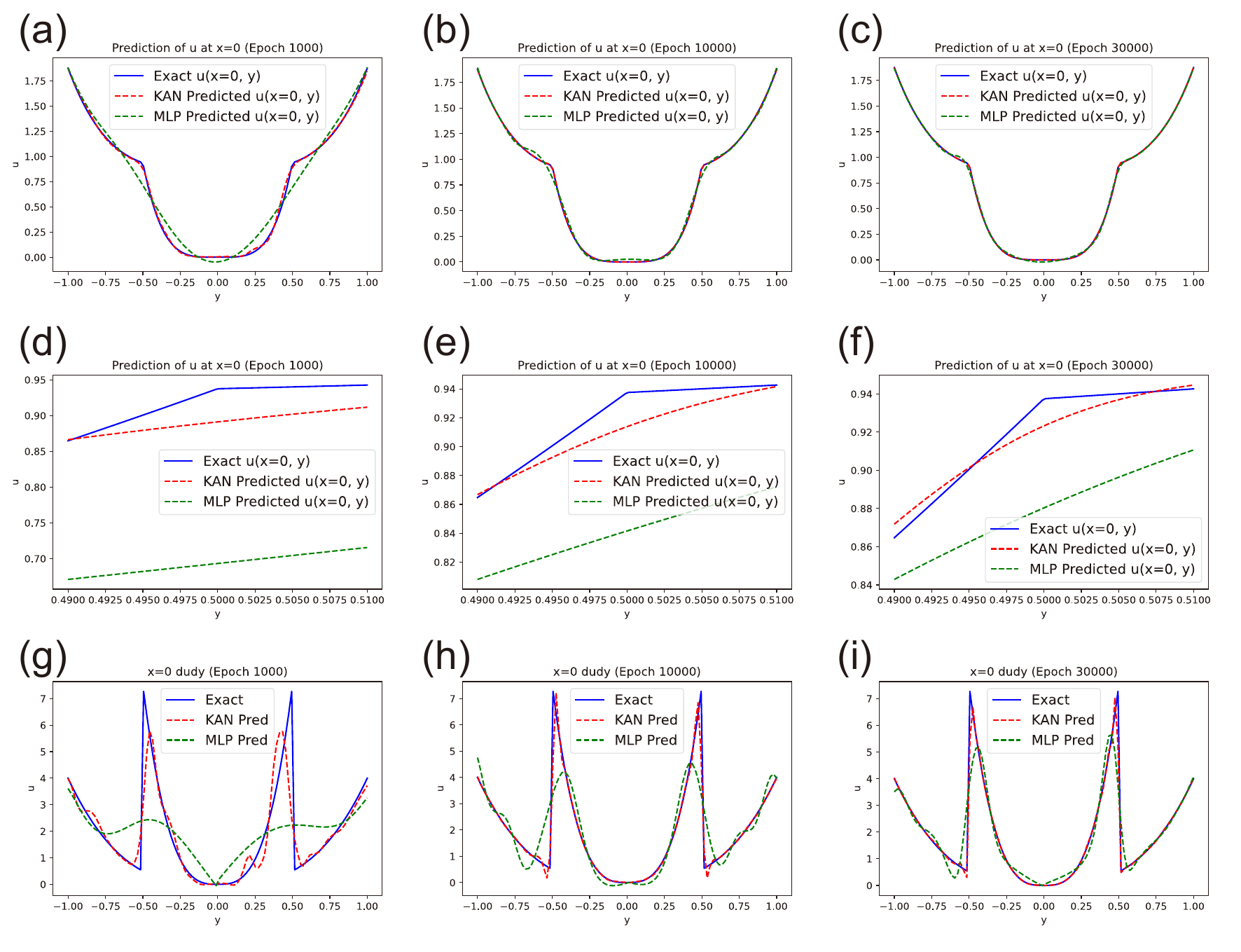}
		\par\end{centering}
	\caption{Fitting curves for the heterogeneous problem at $x=0$: MLP and KAN fitting the original function $u$ at epoch=1000 (a), 10000 (b), and 30000 (c). MLP and KAN fitting the original function $u$ near the interface at epoch=1000 (d), 10000 (e), and 30000 (f). MLP and KAN fitting the derivative $du/dy$ at epoch=1000 (g), 10000 (h), and 30000 (i).
	Analyzing \( \frac{du}{dy} \) is because of the strain discontinuity.
		\label{fig:heter_data}}
\end{figure}

First, we consider a fitting problem to see if KAN has higher accuracy and convergence speed compared to MLP. We consider:
\begin{equation}
	u(\boldsymbol{x}) = \begin{cases}
		\frac{r^{4}}{a_{1}} & r < r_{0}\\
		\frac{r^{4}}{a_{2}} + r_{0}^{4}\left(\frac{1}{a_{1}} - \frac{1}{a_{2}}\right) & r \geq r_{0}
	\end{cases}
	\label{eq:data_heter}
\end{equation}
where $r$ is the distance from point $\boldsymbol{x}$ to the origin (0,0), $a_{1} = 1/15$ and $a_{2} = 1$, $r_{0} = 0.5$, and the solution domain is the square region $[-1,1]^{2}$. Our approach is simple: fit this piecewise function using KAN and MLP to observe whether KAN has a stronger fitting ability for piecewise functions.  \Cref{fig:heter_data} shows the fitting capabilities of KAN and MLP for this piecewise function at different epochs, clearly demonstrating that KAN has a stronger fitting ability for piecewise functions compared to MLP (The relative error of KAN: $1.95 \times 10^{-6}$; MLP: $1.1 \times 10^{-4}$). Here, the network structures for both KAN and MLP are: MLP is structured as [2,100,100,1], and using the tanh activation function, KAN's structure is [2,5,5,1], with a grid size of 20, grid range within the function interval, i.e., [-1,1], and spline order of 3. Except for the neural network structure, all other parameters are the same, such as learning rate and optimizer. It is worth noting that considering the target function's $C_{0}$ continuity, we changed the MLP activation function from tanh to ReLU, which is more suited to the target function, but the accuracy still did not surpass KAN.

It is evident that KAN has the potential to be more accurate and faster for heterogeneous problems compared to MLP. Therefore, we explore its potential in solving heterogeneous PDE problems. We consider a simple Poisson problem for validation:
\begin{equation}
\begin{cases}
	a_{1}\triangle u(\boldsymbol{x}) = 16r^{2} & r < r_{0}\\
	a_{2}\triangle u(\boldsymbol{x}) = 16r^{2} & r \geq r_{0}\\
	u(\boldsymbol{x}) = \frac{r^{4}}{a_{2}} + r_{0}^{4}\left(\frac{1}{a_{1}} - \frac{1}{a_{2}}\right) & \text{Boundary: } x=\pm1,y=\pm1
\end{cases}
\end{equation}
The boundary condition is a pure Dirichlet boundary condition on the four edges of the square ($x=\pm1, y=\pm1$), with parameters the same as in \Cref{eq:data_heter}. However, here we solve the function using PDEs instead of data-driven methods. The point sampling method is shown in \Cref{fig:heterogeneous_intro}b.

It is worth noting that subdomain forms of PINNs, such as CPINNs \cite{CPINN} and the energy form of DEM, CENN \cite{wang2022cenn}, require subdomain division. Traditional PINNs and DEM without subdomain division often perform poorly in solving heterogeneous problems based on MLP, as we will demonstrate in the following numerical examples. In the subdomain form, different regions are approximated by different neural networks, allowing interface information to be added to the training process, and aiding neural network training. In the subdomain form, if we divide the subdomains along the material interface, we need to consider the following equations:
\begin{equation}
\begin{cases}
	a_{1}\triangle u^{+}(\boldsymbol{x}) = 16r^{2} & r < r_{0}\\
	a_{2}\triangle u^{-}(\boldsymbol{x}) = 16r^{2} & r \geq r_{0}\\
	\bar{u}(\boldsymbol{x}) = \frac{r^{4}}{a_{2}} + r_{0}^{4}\left(\frac{1}{a_{1}} - \frac{1}{a_{2}}\right) & \text{Boundary: } x=\pm1,y=\pm1\\
	\boldsymbol{n}\cdot(a_{1}\nabla u^{+}(\boldsymbol{x})) = \boldsymbol{n}\cdot(a_{2}\nabla u^{-}(\boldsymbol{x})) & r = r_{0}\\
	u^{+}(\boldsymbol{x}) = u^{-}(\boldsymbol{x}) & r = r_{0}
\end{cases}
\end{equation}
This is the strong form of the PDE. In CPINNs, the $u^{+}(\boldsymbol{x})$ in the region $r < r_{0}$ is approximated by one neural network, and the $u^{-}(\boldsymbol{x})$ in the region $r \geq r_{0}$ by another. CPINNs require two additional interface equations: the flux continuity condition $\boldsymbol{n}\cdot(a_{1}\nabla u^{+}(\boldsymbol{x})) = \boldsymbol{n}\cdot(a_{2}\nabla u^{-}(\boldsymbol{x}))$ and the original function continuity condition $u^{+}(\boldsymbol{x}) = u^{-}(\boldsymbol{x})$, introducing extra hyperparameters. In CENN, if we use the distance function from \Cref{eq:admissible} to construct the admissible function that satisfies the essential boundary conditions, we only need to consider:
\begin{equation}
\begin{aligned}
	\mathcal{L}_{CENN} &= \frac{1}{2}\int_{\Omega^{+}}a_{1}(\nabla u^{+}(\boldsymbol{x}))\cdot(\nabla u^{+}(\boldsymbol{x}))d\Omega + \frac{1}{2}\int_{\Omega^{-}}a_{2}(\nabla u^{-}(\boldsymbol{x}))\cdot(\nabla u^{-}(\boldsymbol{x}))d\Omega\\
	&\quad + \beta\int_{\Gamma^{inter}}[u^{+}(\boldsymbol{x})-u^{-}(\boldsymbol{x})]^{2}d\Gamma
\end{aligned}
\end{equation}
where $\Gamma^{inter}$ is the interface, i.e., $r = r_{0}$. Compared to the strong form, the interface flux continuity equation is not needed, only the original function continuity condition, as proven in \cite{wang2022cenn}.

Both CPINNs and CENN require subdomain division. The advantage of subdomain division is improved accuracy in solving heterogeneous problems by adding inductive bias. The disadvantage is increased computational cost due to additional interface information and more hyperparameters. We note that KAN has the nature of piecewise functions and theoretically is more suitable for solving heterogeneous problems, as the function in heterogeneous problems is piecewise, as shown in \Cref{fig:heter_data}. Thus, we naturally propose whether using KAN instead of MLP can achieve good results without subdomains.

\begin{figure}
	\begin{centering}
		\includegraphics[scale=0.52]{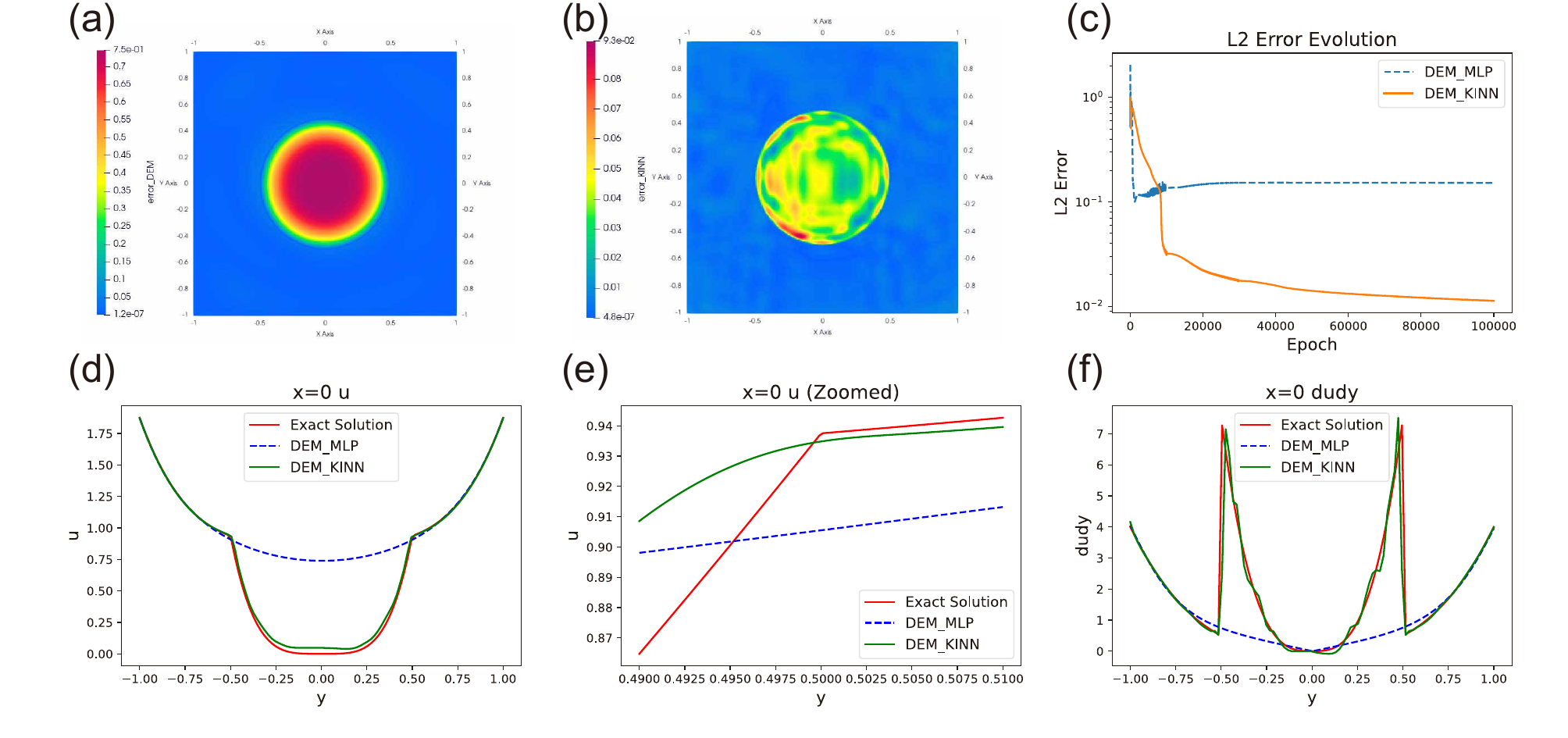}
		\par\end{centering}
	\caption{PDEs solving for heterogeneous problems, comparing DEM with MLP and KAN without subdomains. (a) Absolute error contour of DEM using MLP. (b) Absolute error contour of DEM using KAN. (c) Comparison of the relative error $\mathcal{L}_{2}$ evolution trends between DEM using MLP and KAN. (d) Comparison of the predicted $u$ and exact solution at $x=0$ between DEM using MLP and KAN. (e) Zoomed comparison of the predicted $u$ and exact solution near the interface $x=0,y=0.5$ between DEM using MLP and KAN. (f) Comparison of the predicted $du/dy$ and exact solution at $x=0$ between DEM using MLP and KAN.\label{fig:pdes_heter}}
\end{figure}

If no subdomain division is used, it is an advantage that approximating the entire region with one neural network does not require interface equations. Here we only verify the energy form of PINNs. When using one neural network to approximate the entire region, we only need to consider the following optimization function:
\begin{equation}
\mathcal{L} = \frac{1}{2}\int_{\Omega^{+}}a_{1}(\nabla u)\cdot(\nabla u)d\Omega + \frac{1}{2}\int_{\Omega^{-}}a_{2}(\nabla u)\cdot(\nabla u)d\Omega + \beta\int_{\Gamma^{u}}[u(\boldsymbol{x}) - \bar{u}(\boldsymbol{x})]^{2}d\Gamma
\end{equation}
Here we use a penalty function to satisfy the boundary conditions $\Gamma^{u}$. We conduct numerical experiments to verify, using DEM to solve this problem with both MLP and KAN, exploring whether KAN can achieve good results for heterogeneous problems without subdomain division.
In DEM, we use the triangular integration.
\Cref{fig:pdes_heter} compares DEM using KAN and MLP, clearly showing that KAN achieves good results without subdomain division for heterogeneous problems, whereas MLP does not. MLP struggles to bend the function at the interface $r=0.5$. Note that KAN is a smooth function, so it still cannot perfectly fit near the interface, as shown in \Cref{fig:pdes_heter}e. However, compared to MLP,  \Cref{fig:pdes_heter}f clearly shows that KAN can better capture the interface information.

In conclusion, the numerical experiments show that KAN has better potential than MLP for solving heterogeneous problems without subdomain division. Of course, we can still use CENN or CPINNs with subdomain division to solve this problem, and we believe the accuracy and convergence speed will be even better.

We now test the performance of KINN on heterogeneous inverse problems. We consider the steady-state heat conduction equation in a two-dimensional non-uniform medium:
\begin{equation}
	\begin{aligned}
		-k(x,y) \Delta T(x,y) = f(x,y), & \quad \{x,y\} \in [0,1]^2, \\
		T(x,y) = \cos(15\pi xy), & \quad \{x,y\} \in \Gamma,
	\end{aligned}
	\label{eq:heat_equation}
\end{equation}
where \(k(x,y)\) is the non-uniform thermal conductivity coefficient, \(T(x,y)\) is the temperature field, and \(f(x,y)\) is the heat source term. The two-dimensional area is a square \([0,1] \times [0,1]\), and \(\Gamma\) represents the boundary of the square. We employ a full inverse method, imposing boundary conditions and determining \(f\) using a pre-assumed analytical solution. We assume the analytical solution for the temperature field is \(T(x,y) = \cos(15\pi xy)\). It can be verified that \(f = \cos(15\pi xy)[(15\pi x)^2 + (15\pi y)^2]k(x,y)\).

\begin{figure}
	\centering
	\includegraphics[scale=0.55]{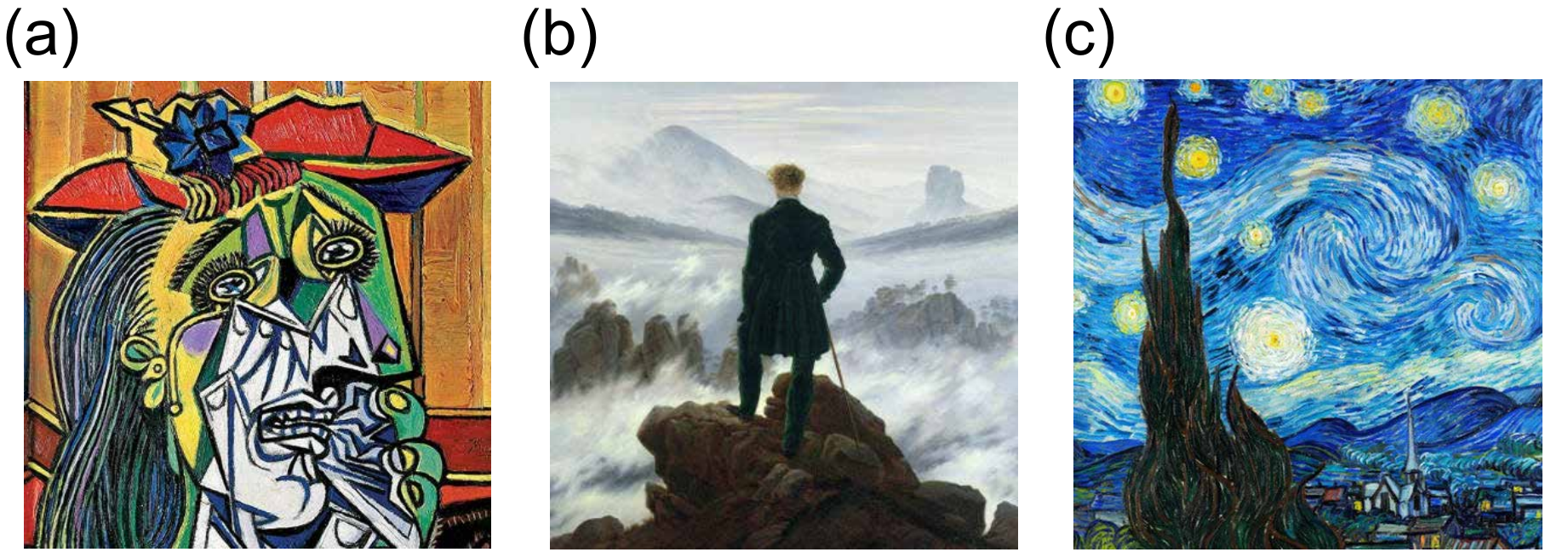}
	\caption{Three famous paintings as coefficients of thermal conductivity: (a) Picasso's "The Weeping Woman", (b) Caspar's "Wanderer above the Sea of Fog", (c) Van Gogh's "The Starry Night". \label{fig:famous_painting}}
\end{figure}

\begin{figure}
	\centering
	\includegraphics[scale=0.38]{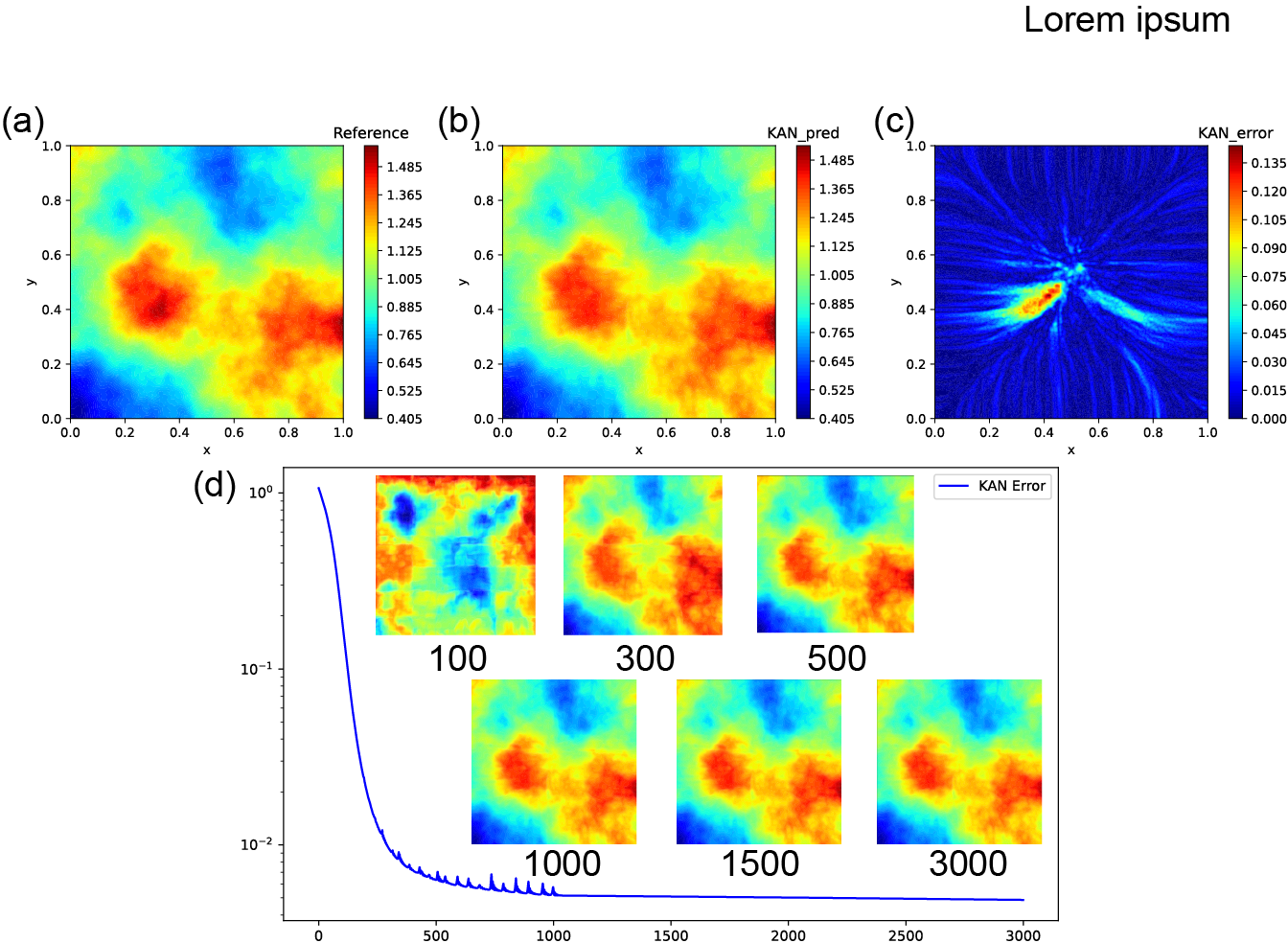}
	\caption{Performance of KINN in the heterogeneous Darcy Flow inverse problem: (a) Reference solution for the heterogeneous field $k(x,y)$; (b) Prediction solution of the inverse problem by KINN at 3000 epochs; (c) Absolute error of KINN; (d) Evolution of the KINN loss function with iterations, showing results for 100, 300, 500, 1000, 1500, and 3000 epochs.}
	\label{fig:KINN_darcy_inverse}
\end{figure}

Science is an art. Here, we employ Picasso's "The Weeping Woman", Caspar's "Wanderer above the Sea of Fog", and Van Gogh's "The Starry Night" as non-uniform \(k(x,y)\), as shown in  \Cref{fig:famous_painting}a, b, and c.
Note that actual steady-state heat conduction equation  is $-\nabla\cdot[k(x,y)\nabla T(x,y))]=f(x,y)$. Given the significant variability in $k(x, y)$ when using images as the thermal conductivities, the gradient of $k(x,y)$  can be excessively large, leading to a substantial approximation error of gradient for $k(x,y)$. Thus, the significant variability in $k(x, y)$ will distort the PINNs algorithm. Therefore, we simplify \Cref{eq:Darcy} to \Cref{eq:heat_equation} when using images as $k(x,y)$. However, we still consider the inverse problem about \Cref{eq:Darcy} when $k(x,y)$ has greater smoothness and continuity as shown in \Cref{fig:KINN_darcy_inverse}.

First, we process these paintings from RGB to 256x256 pixel grayscale values, then scale these values between 0 and 1, where the maximum is 1 and the minimum is 0. These grayscale values represent the thermal conductivity coefficients \(k(x,y)\). In the inverse problem, we assume that \(k(x,y)\) is unknown but must pre-determine the temperature field \(T(x,y)\) and \(f(x,y)\), then solve for \(k(x,y)\) using \Cref{eq:heat_equation}.

Our approach follows the methods in \cite{chen2021learning,liu2024multi}. We first establish a neural network mapping from coordinates to the \(k(x,y)\) field, \(NN_k(x,y; \boldsymbol{\theta}_k)\). Using existing temperature data points \(\{x_i, y_i, T_i, F_i\}_{i=1}^N\), \(NN_k(x,y; \boldsymbol{\theta}_k)\) outputs the thermal conductivity at these locations \(\{x_i, y_i, K_i\}_{i=1}^N\). As the thermal conductivity coefficient is determined by the neural network parameters \(\boldsymbol{\theta}_k\), the neural network initially does not satisfy the physical equation \Cref{eq:heat_equation}, requiring iterative adjustment of \(\boldsymbol{\theta}_k\) according to the loss function:
\begin{equation}
	\boldsymbol{\theta}_k = \arg\min_{\boldsymbol{\theta}_k} \sum_{i=1}^N |K_i(\boldsymbol{\theta}_k) \Delta T_i + F_i|^2,
	\label{eq:k_non_heat_equation_painting}
\end{equation}
We use finite differences to construct the differential operator since we are only given discrete points. If we were to use the AD algorithm to build the loss, it would necessitate a new neural network to approximate the temperature field \(T\), potentially increasing the approximation error. Hence, we apply finite differences directly to the raw data to construct the differential operator. Subsequently, the loss values for all data points are calculated, and the parameters in \(NN_k(x,y; \boldsymbol{\theta}_k)\) are optimized based on \Cref{eq:k_non_heat_equation_painting}.

To compare the performance of KAN and MLP in solving inverse problems, we first conduct a data-driven test to observe how MLP and KAN fit this type of highly heterogeneous problem. The network structures for KAN and MLP are shown in \Cref{tab:MLP_KAN_heterogenous}, and due to high heterogeneity, we use a larger grid size (100) for KAN to enhance its fitting capability.
Initially, we set the grid size of KAN to 15. Although the model showed some preliminary capability in image fitting, the performance is not optimal. Considering the nature of high heterogeneity, increasing the grid size (from 15 to 100) improves KAN's accuracy in fitting famous paintings. This demonstrates that the grid size for KAN needs to be matched with the complexity of the fitting problem.

\begin{figure}
	\centering
	\includegraphics[scale=0.4]{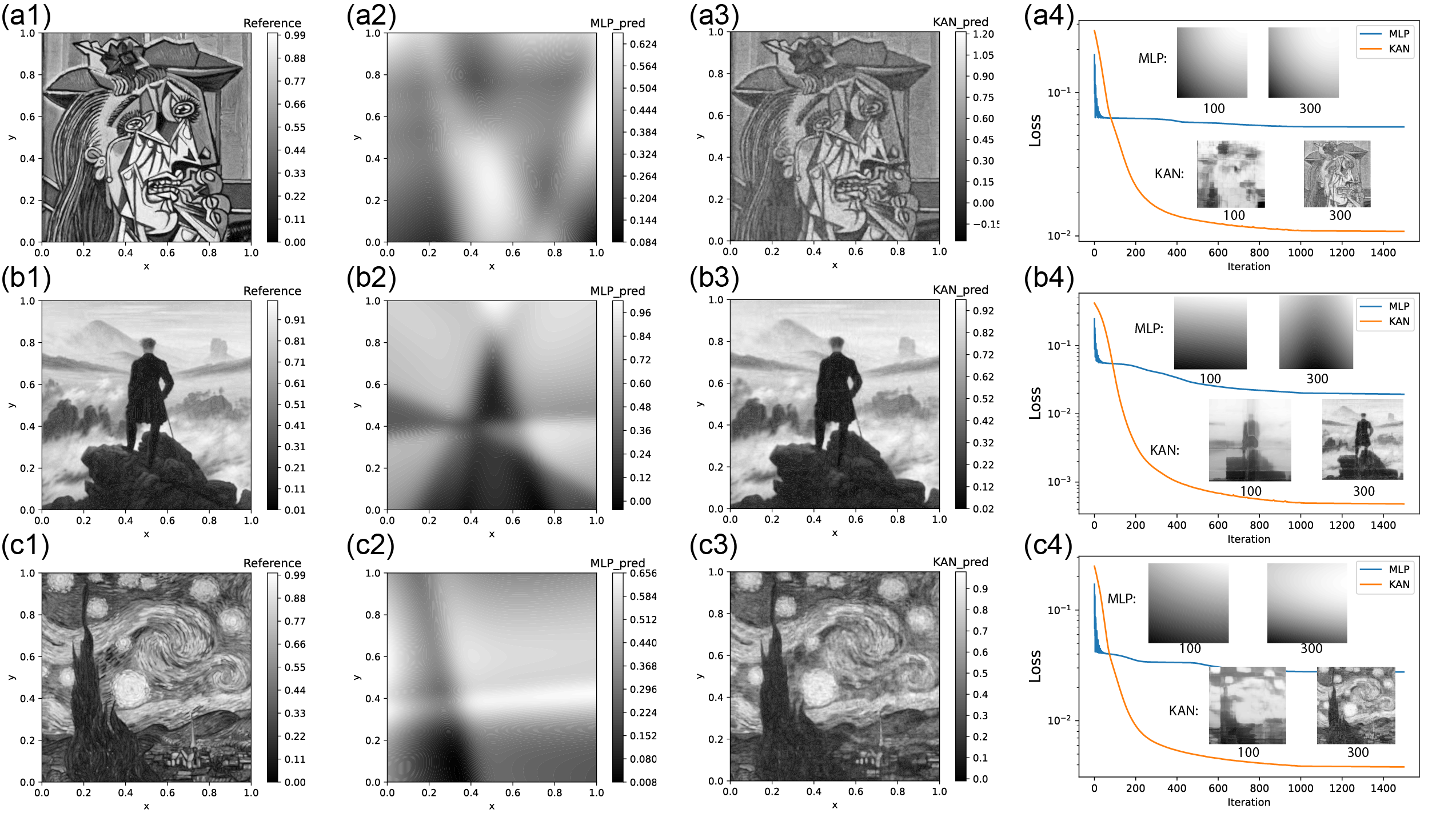}
	\caption{Data-driven fitting results of MLP and KAN for heterogeneous materials: The first row is Picasso's \textit{The Weeping Woman} (a1,2,3,4); the second row is Caspar's \textit{Wanderer above the Sea of Fog} (b1,2,3,4); the third row is Van Gogh's \textit{The Starry Night} (c1,2,3,4). The first column shows the reference solutions of thermal conductivity; the second column shows the MLP predictions at 1500 epochs; the third column shows the KAN predictions at 1500 epochs; the fourth column shows the evolution of the MISE loss function for MLP and KAN with iterations. Predictions are provided for 100 and 300 epochs.}
	\label{fig:painting_data}
\end{figure}

 \Cref{fig:painting_data} shows the fitting results for MLP and KAN on this highly heterogeneous problem. We find that KAN significantly outperforms MLP, especially in target functions with strong discontinuities (poor smoothness). MLP almost completely fails. This demonstrates KAN's greater potential for handling highly heterogeneous problems compared to MLP.

\begin{table}
	\caption{Performance of MLP and KAN on heterogeneous materials. P denotes Picasso's \textit{The Weeping Woman}, C denotes  Caspar's \textit{Wanderer above the Sea of Fog}, and V denotes Van Gogh's \textit{The Starry Night}.}
	\label{tab:MLP_KAN_heterogenous}
	\begin{adjustbox}{max width=\textwidth}
	\centering
	\begin{tabular}{cccccccc}
		\toprule
		Algorithms & Data-driven Relative Error ($\mathcal{L}_{2}$) & PDEs Relative Error ($\mathcal{L}_{2}$) & Grid Size & Grid Range & Order & Architecture of NNs & Parameters \\
		\midrule
		MLP & P: $48.6\%$; C: $21.5\%$; V: $35.0\%$ & & & & & [2, 200, 200, 1] & 41001 \\
		KAN & P: $21.1\%$; C: $3.37\%$; V: $12.9\%$ & P: $31.2\%$; C: $5.53\%$; V: $20.2\%$ & 100 & [0, 1] & 3 & [2, 15, 15, 1] & 28350 \\
		\bottomrule
	\end{tabular}
	\end{adjustbox}
\end{table}

\begin{figure}
	\centering
	\includegraphics[scale=0.36]{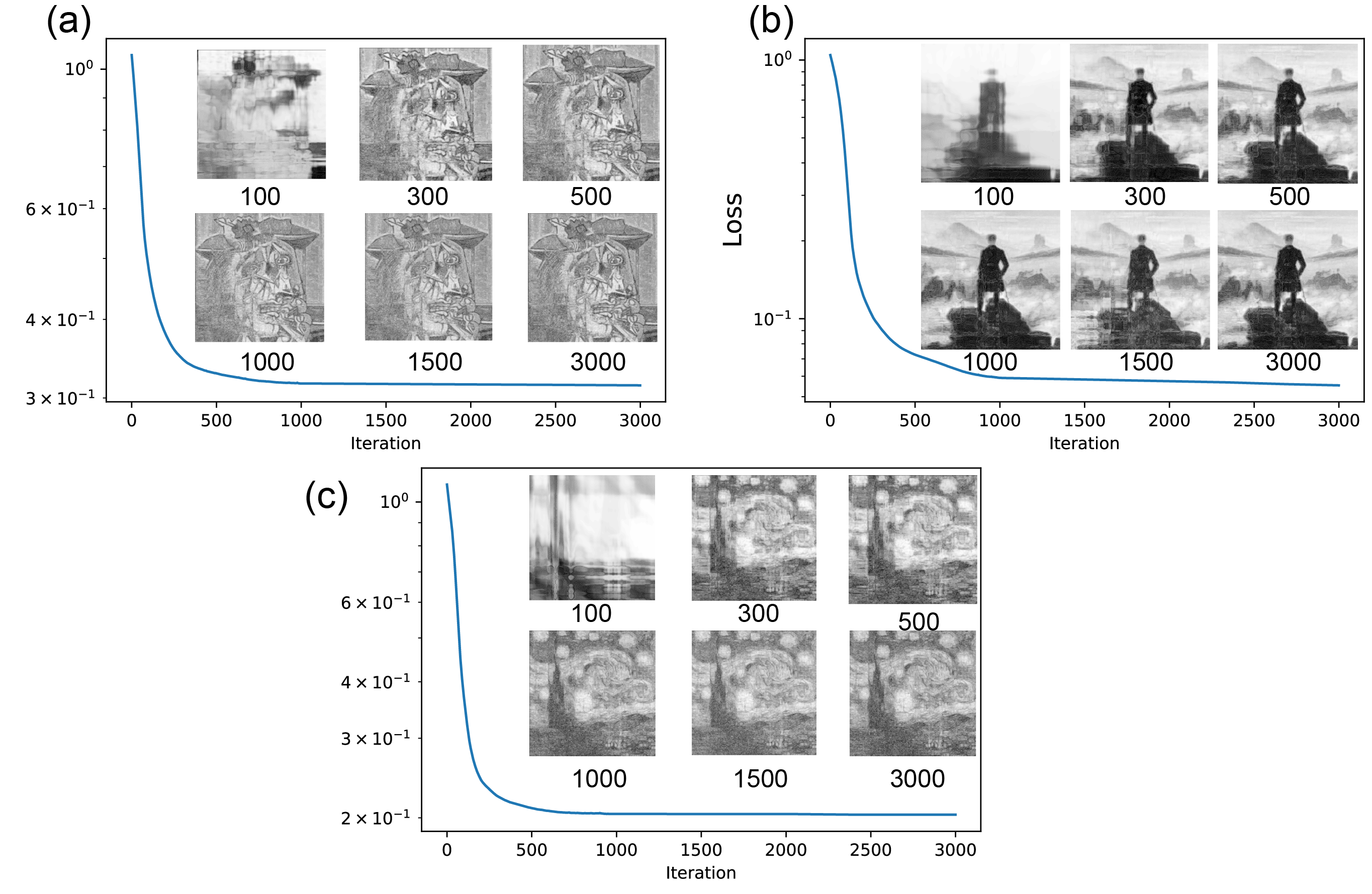}
	\caption{KINN's performance in solving PDEs for heterogeneous materials: (a) Relative error $\mathcal{L}_{2}$ for Picasso's \textit{The Weeping Woman}; (b) Relative error $\mathcal{L}_{2}$ for Caspar's \textit{Wanderer above the Sea of Fog}; (c) Relative error $\mathcal{L}_{2}$ for Van Gogh's \textit{The Starry Night}. The numbers 100, 300, 500, 1000, 1500, and 3000 represent the KINN prediction results at these iteration counts.}
	\label{fig:painting_pdes}
\end{figure}

The aforementioned results are data-driven. We now consider the inverse problem for PDEs as shown in  \Cref{eq:heat_equation}. Due to MLP's poor performance in handling highly heterogeneous problems, we only focus on the performance of KAN in solving PDE inverse problems and exclude MLP. The optimization process of the algorithm is as shown in \Cref{eq:k_non_heat_equation_painting}.  \Cref{fig:painting_pdes} shows the relative error $\mathcal{L}_{2}$ of KINN in solving the thermal conductivity inverse problem for the three famous paintings. We also present the prediction results of KINN at different iteration numbers. \Cref{tab:MLP_KAN_heterogenous} displays the absolute errors of KINN. Since the thermal conductivity represented by famous paintings is very non-smooth, there is considerable noise. Surprisingly, KAN fits the data very well. This not only indicates that KAN has a stronger fitting capability than MLP but also suggests that careful denoising is crucial when using KAN. In real-world inverse problems, experimental data often contain noise, and KAN might overfit this noise, reducing its performance. Thus, exploring methods to detect noise in advance and automatically adjust the size of the KAN network is essential for its application in inverse problems.

Next, we consider a more physically realistic PDE with a relatively smoother heterogeneous field $k(x,y)$:
\begin{equation}
	\begin{aligned}
		-\nabla\cdot[k(x,y)\nabla T(x,y)] &= f(x,y), \{x,y\}\in[0,1]^{2} \\
		T(x,y) &= 0, \{x,y\}\in\Gamma
	\end{aligned}
	\label{eq:Darcy}
\end{equation}
where $f(x,y)=1$, which is the well-known Darcy Flow equation. We generate the heterogeneous field $k(x,y)$ using a Gaussian random field and obtain $T$ using a traditional finite difference method. Note that  \Cref{eq:Darcy} is different from \Cref{eq:heat_equation} because in \Cref{eq:Darcy}, we need to consider the derivative of $k(x,y)$:
\begin{equation}
	\begin{aligned}
		-\nabla\cdot[k(x,y)\nabla T(x,y)] &= f(x,y) \\
		-(k_{,i}T_{,i}+kT_{,ii}) &= f(x,y)
	\end{aligned}
	\label{eq:Darcy-1}
\end{equation}

The approach to solving the Darcy flow inverse problem is similar to that for \Cref{eq:heat_equation}, with the loss function given by:
\begin{equation}
	\boldsymbol{\theta}_{k} = \arg\min_{\boldsymbol{\theta}_{k}} \sum_{i=1}^{N} |K_{i}(\boldsymbol{\theta}_{k})T_{,i} + K_{i}(\boldsymbol{\theta}_{k})\triangle T_{i} + F_{i}|^{2}.
	\label{eq:Darcy_heterogeneous}
\end{equation}

We construct the differential operator using finite differences. \Cref{fig:KINN_darcy_inverse} shows the performance of KINN in the Darcy Flow inverse problem, with KINN achieving convergence at 1000 epochs and a relative error $\mathcal{L}_{2}$ of $1.69\%$.

In summary, in inverse problems, KAN significantly outperforms MLP in terms of accuracy for highly complex inverse problem fields due to its superior fitting capability.

Please note that we solve inverse problems solely using the strong form of PDEs and do not employ the energy form. This is because the energy form encounters mathematical issues when solving inverse problems, as proved in \ref{sec:DEM_difficult_inverse}. For the inverse form by BINN, due to the limitations imposed by fundamental solutions as shown in \Cref{eq:fundamental_solution}, the range of solvable PDEs using BINN is not as general as the strong form PINNs. Therefore, we do not delve further into solving inverse problems with BINN here. However, exploring BINN for solving inverse problems remains a promising research direction for the future.

\subsection{KAN does not work well on complex geometries}

\begin{figure}
	\begin{centering}
		\includegraphics[scale=0.8]{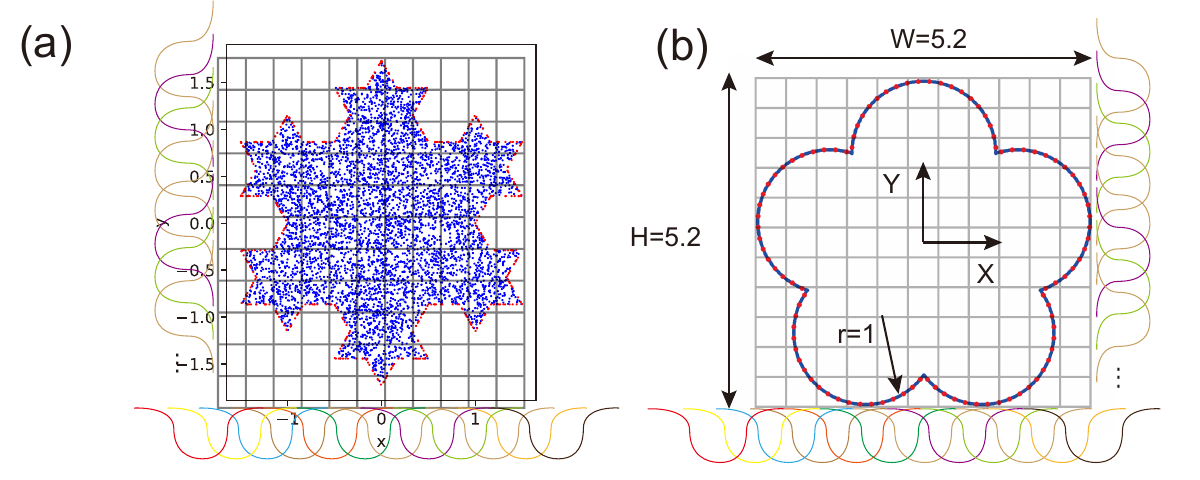}
		\par\end{centering}
	\caption{The introduction of complex geometry problem: (a) Koch: Blue points are interior points, red points are Dirichlet boundary conditions. (b) Flower: Red points are source points in BINN, and blue points are integration points.
	Because there are too many blue points, they are presented like blue lines. The radius of each flower petal is 1. \label{fig:The_intro_koch_flower}}
\end{figure}

In this example, we demonstrate the performance of KINN on complex geometries, as shown in \Cref{fig:The_intro_koch_flower}. We solve the Laplace equation, with all boundary conditions being Dirichlet boundary conditions:
\begin{equation}
	\begin{aligned}
		\triangle u(\boldsymbol{x})=0 & ,x\in\Omega\\
		u(\boldsymbol{x})=\bar{u}(\boldsymbol{x}) & ,x\in\Gamma
	\end{aligned}
	\label{eq:complex_pdes}
\end{equation}
The analytical solution is:
\begin{equation}
	u(\boldsymbol{x}) = \sin(x)\sinh(y) + \cos(x)\cosh(y)
	\label{eq:complex_exact}
\end{equation}
Note that the boundary condition $\bar{u}(\boldsymbol{x})$ is derived from the analytical solution. We solve the Koch snowflake and flower problems, as shown in \Cref{fig:The_intro_koch_flower}. The analytical solutions in \Cref{eq:complex_exact} and PDEs for the Koch and flower are consistent.

\begin{figure}
	\begin{centering}
		\includegraphics[scale=0.6]{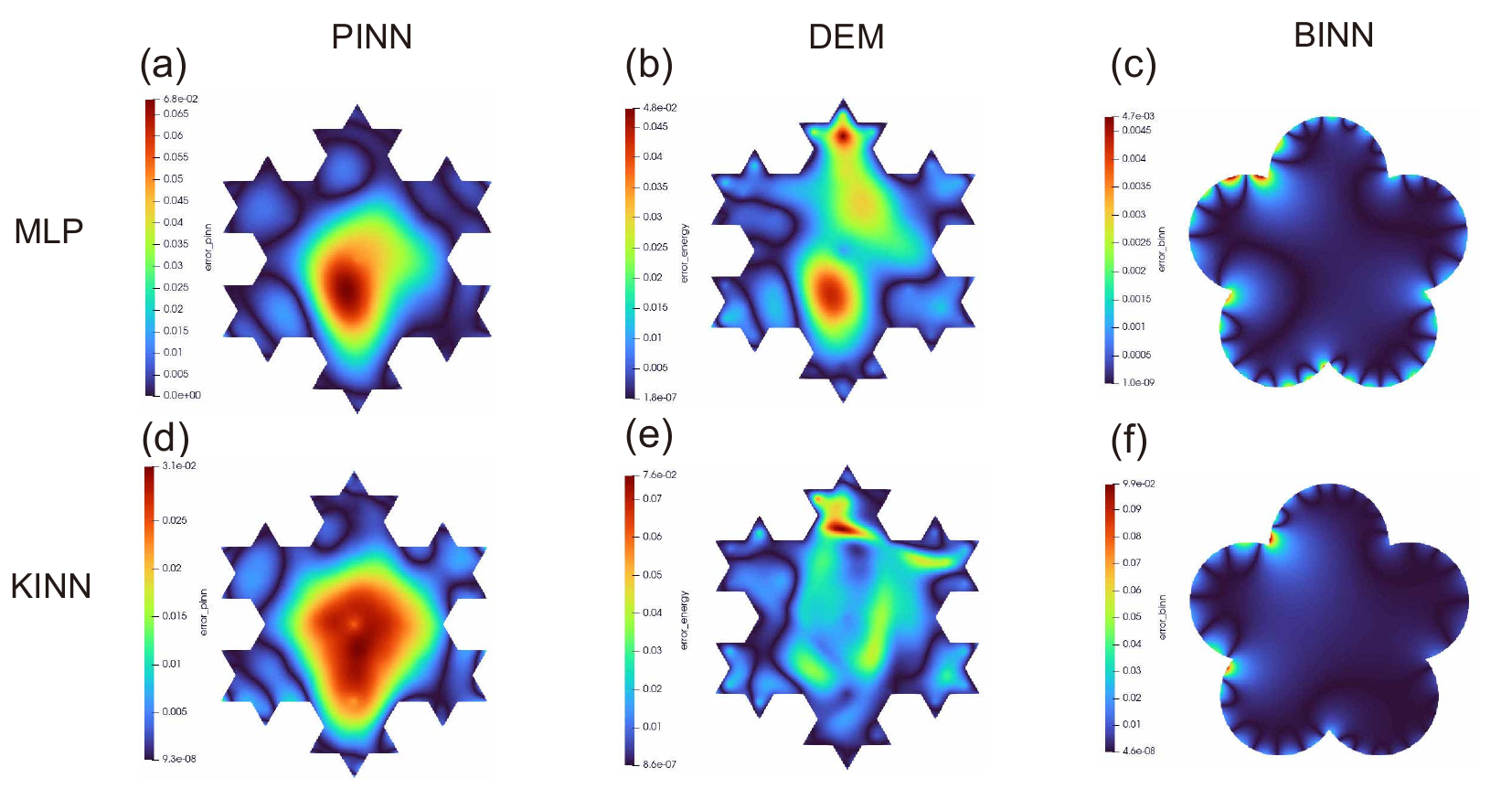}
		\par\end{centering}
	\caption{The absolute error by PINN, DEM, BINN with MLP and KINN in complex geometry problem: PINN, DEM, BINN with all MLP (first row), PINN, DEM, BINN with all KAN (second row). \label{fig:The-absolute-error_complex_geo}}
\end{figure}

\begin{figure}
	\begin{centering}
		\includegraphics[scale=0.6]{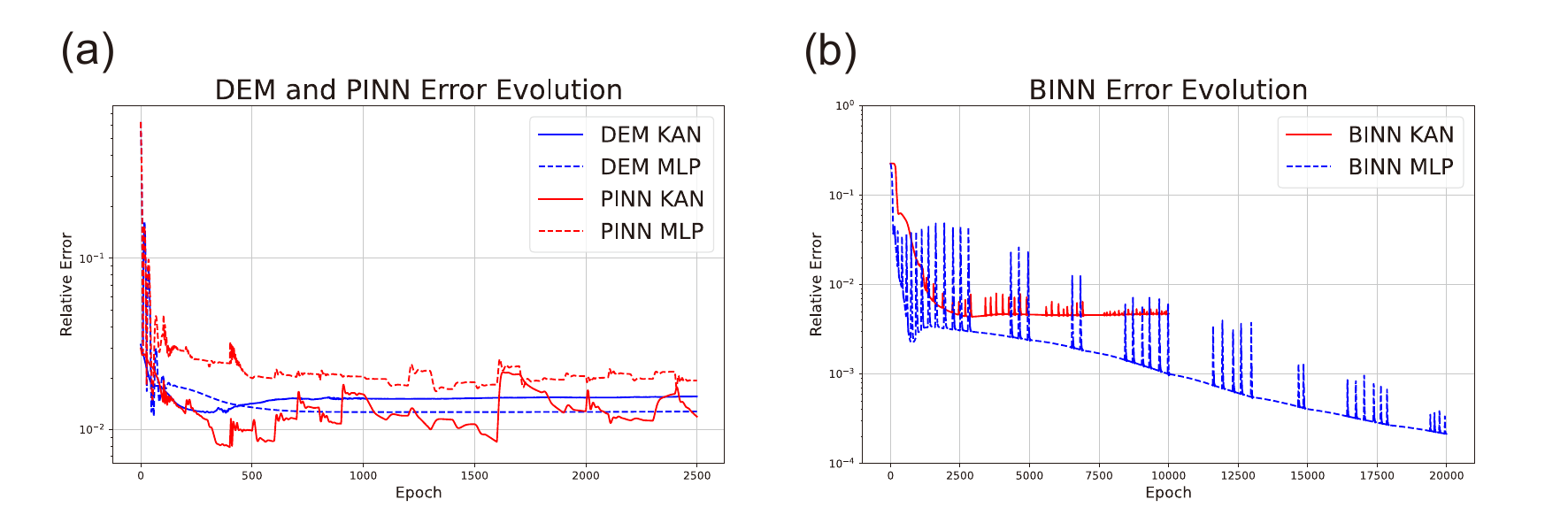}
		\par\end{centering}
	\caption{Error evolution of KINN and corresponding MLP algorithms in complex geometry: (a) PINNs and DEM $\mathcal{L}_{2}$ relative error evolution trend. (b) BINN $\mathcal{L}_{2}$ relative error evolution trend. \label{fig:error_evo_complex}}
\end{figure}

In the Koch problem, we use the strong form of PINNs and the energy form DEM. In DEM, we use Monte Carlo integration for energy. The $u(\boldsymbol{x})$ in PINNs and DEM is constructed using the distance function from \Cref{eq:admissible}:
\begin{equation}
	u(\boldsymbol{x}) = u_{p}(\boldsymbol{x}) + D(\boldsymbol{x}) * u_{g}(\boldsymbol{x}).
\end{equation}
In the flower problem, we use the boundary element method (BINN). Note that we solve all forms using both MLP and KAN. \Cref{fig:The-absolute-error_complex_geo} shows the relative error contour plots, indicating that KAN does not perform well in solving complex geometric problems. \Cref{fig:error_evo_complex} shows the relative error evolution trends for PINNs, DEM, and BINN. The results indicate that KAN performs similarly or even worse than MLP in complex boundaries, particularly in BINN.  \Cref{tab:The-comparison-between_complex_geo} summarizes the network structure, error, and time for each algorithm.

In conclusion, KINN does not show a significant improvement over MLP in solving complex geometries compared to regular geometries. The reason is that the grid range of KAN is rectangular in high dimensions, making it inherently more suitable for regular geometries. Additionally, there is a correlation between grid size and geometric complexity; the more complex the geometry, the larger the grid size should be. This is similar to the complexity of the fitted function, as shown in  \Cref{fig:KAN_MLP_heat}, where higher frequencies require larger grid sizes. However, increasing the grid size beyond a certain point can suddenly increase errors, as demonstrated in the original KAN paper \cite{liu2024kan} and our results in  \Cref{fig:crack_scaling_law}. Therefore, the grid size cannot be increased indefinitely and must correspond to the complexity of the fitted function to select an optimal grid size to prevent overfitting. 

Thus, in complex geometries there is a conflict between the simplicity of the function and the complexity of the geometry, making it difficult to find a balance between grid size and the complexity of the geometries. This is a potential area for future exploration to improve KAN's accuracy in solving PDEs in complex geometries.

To increase the performance of KINN in complex geometries, mesh adaptation techniques from finite element methods, such as h-p-refinements \cite{zienkiewicz1989effective}, especially h-refinement, can be incorporated into KINN. H-refinement can be applied to modify the grid size in KAN. Another approach is to use the concept of isoparametric transformation from finite element methods \cite{paradeepenergy} to transform complex geometries into a simpler reference domain. Alternatively, the idea of conformal transformations from mathematics can be used to handle complex geometries by mapping them into another easier-to-manage spatial domain to leverage the capabilities of KAN. 

\begin{table}
	\caption{The comparison between different PINNs, DEM, and BINN algorithms based on MLP or KAN in complex geometry. \label{tab:The-comparison-between_complex_geo}}
	\centering{}
	\begin{adjustbox}{max width=\textwidth}
	\begin{tabular}{cccccccc}
		\toprule 
		\multirow{1}{*}{Algorithms} & Relative error & Grid size & grid\_range & order & Architecture of NNs & Parameters & Time (Second, 1000 Epoch)\\
		\midrule 
		PINNs\_MLP & 0.01939 &  &  &  & {[}2, 30,30,30,30,1{]} & 2911 & 13.32\\
		KINN: PINNs & 0.01188 & 10 & {[}-1,1{]} & 2 & {[}2, 5,5,5,1{]} & 910 & 30.18\\
		DEM\_MLP & 0.01563 &  &  &  & {[}2, 30,30,30,30,1{]} & 2911 & 11.46\\
		KINN: DEM & 0.01565 & 3 & {[}-1,1{]} & 2 & {[}2, 5,5,1{]} & 280 & 11.38\\
		BINN\_MLP & 0.0002061 &  &  &  & {[}2, 30,30,30,30,1{]} & 2911 & 3.99\\
		KINN: BINN & 0.004729 & 10 & {[}-1,1{]} & 3 & {[}2, 5,5,5,1{]} & 975 & 14.95\\
		\bottomrule
	\end{tabular}
\end{adjustbox}
\end{table}

\section{Conclusion\label{sec:Conclusion}}
We compared KAN and MLP in strong, energy, and inverse forms of PDEs, proposing the KINN algorithm for solving forward and inverse problems. In forward problems, we conducted systematic numerical experiments and validations on common benchmarks in engineering mechanics. The results show that KAN has higher accuracy and convergence speed than MLP in most PDE problems, particularly in singularity problems, stress concentration problems, nonlinear hyperelastic problems, and heterogeneous problems. However, KAN's efficiency is currently lower than MLP under the same epoch due to the lack of specific optimizations for the KAN algorithm. With optimization, the efficiency of KINN can be significantly improved. Additionally, we systematically analyzed KAN from the NTK perspective and found that it has a much smaller spectral bias than MLP, making it more suitable for solving high and low-frequency mixed problems. Finally, we discovered that KAN does not perform as well on complex geometric PDE problems, primarily due to the conflict between grid size and geometric complexity. In inverse problems, validation on heterogeneous problems reveals that KAN has a stronger fitting capability compared to MLP, demonstrating significant advantages for very complex inverse problems.

Nevertheless, KINN currently has limitations and areas for expansion. During our experiments, we found that an overly large grid size can cause KAN to fail, i.e., the error increases due to overfitting. Therefore, when using KINN, it is important to determine the grid size according to the problem's complexity. In the future, incorporating mesh adjustment techniques and concepts from traditional FEM can help KAN adjust the appropriate grid size. Furthermore, pure data can be used to analyze function fields obtained from traditional algorithms, and we can employ KAN to obtain simple symbolic representations of functions. Moreover, mesh adjustment methods from FEM can also be integrated into KINN, such as h-p-refinements \cite{zienkiewicz1989effective}, particularly h-refinement. The concept of h-refinement can be used to adjust the grid size in KAN. 
Alternatively, the idea of isoparametric transformation in finite element methods \cite{paradeepenergy} can be used to map complex geometries to a simple reference domain. The mesh generation techniques in finite element analysis (h-p-refinements) and the concept of isoparametric transformation could both potentially address the issue of low accuracy in solving complex geometries with KINN  in the future.
Additionally, we have not yet explored the weak form. This is because the weak form requires consideration of different weight functions, making the research more complex compared to other forms. Moreover, the energy form is a special case of the weak form. Nonetheless, it is undeniable that studying the performance of KINN  in the weak form will be an important direction for future research. Interested readers can refer to \cite{hp-VPINN,kharazmi2019variational} on the weak form.
As for efficiency, although KINN has fewer trainable parameters, its computation time per epoch is longer than that of the corresponding MLP. This is because KAN is self-implemented and lacks the optimized AI libraries available for MLP. This suggests that there is potential for future optimization of KAN within KINN. The majority of the computation time in KAN is spent on calculating B-spline parameters. In the future, more efficient basis functions instead of B-splines or optimized B-splines calculation methods could be used in KINN to increase the efficiency of KINN. For example, Shukla et al.  \cite{shukla2024comprehensive} proposed using Chebyshev orthogonal polynomials to significantly improve the speed of KAN in solving PDEs.

In conclusion, we believe that KAN will become the cornerstone of AI for PDEs in the future, replacing MLP, because KAN is more in line with the essence of solving PDEs compared to MLP. KAN will be the "engine" in AI for PDEs.

\section*{Declaration of competing interest}
The authors declare that they have no known competing financial interests or personal relationships that could
have appeared to influence the work reported in this paper.

\section*{Acknowledgement}
The study was supported by the Key Project of the National Natural Science Foundation of China (12332005) and scholarship from Bauhaus University in Weimar. Thanks for the helpful discussions with the authors of the original KAN paper \cite{liu2024kan}, especially Ziming Liu, Yixuan Wang, and Thomas Y. Hou.

\appendix

\section{The requirement of PINNs to DEM\label{sec:PINN_DEM}}

Consider a general form of PDEs as in \Cref{eq:original_form}, written in the Galerkin form:
\begin{equation}
	\int_{\Omega}[\boldsymbol{P}(\boldsymbol{u})-\boldsymbol{f}]\cdot\delta\boldsymbol{u}d\Omega-\int_{\Gamma}[\boldsymbol{B}(\boldsymbol{u})-\boldsymbol{g}]\cdot\delta\boldsymbol{u}d\Gamma=0\label{eq:Garerkin_form_PDEs}
\end{equation}
If the differential operator $\boldsymbol{P}$ is linear and self-adjoint, the linear operator is:
\begin{equation}
\boldsymbol{P}(\sum_{i=1}^{N}\alpha_{i}\boldsymbol{u}_{i})=\sum_{i=1}^{N}\alpha_{i}\boldsymbol{P}(\boldsymbol{u}_{i})
\end{equation}
The self-adjoint operator must satisfy:
\[
\int_{\Omega}\boldsymbol{P}(\boldsymbol{u})\cdot\boldsymbol{v}d\Omega=\int_{\Omega}\boldsymbol{u}\cdot\boldsymbol{P}^{*}(\boldsymbol{v})d\Omega+b(\boldsymbol{u},\boldsymbol{v})
\]
where, after integration by parts, $\boldsymbol{P}=\boldsymbol{P}^{*}$ and $b(\boldsymbol{u},\boldsymbol{v})$ refers to the boundary integral term regarding $\boldsymbol{u}$ and $\boldsymbol{v}$.

If the operator $\boldsymbol{P}$ in \Cref{eq:Garerkin_form_PDEs} is linear and self-adjoint, we use the self-adjoint property of $\boldsymbol{P}$:
\begin{equation}
	\begin{aligned}
		\int_{\Omega}\boldsymbol{P}(\boldsymbol{u})\cdot\delta\boldsymbol{u}d\Omega &= \int_{\Omega}\frac{1}{2}\boldsymbol{P}(\boldsymbol{u})\cdot\delta\boldsymbol{u}d\Omega+\int_{\Omega}\frac{1}{2}\boldsymbol{P}(\boldsymbol{u})\cdot\delta\boldsymbol{u}d\Omega \\
		&= \int_{\Omega}\frac{1}{2}\boldsymbol{P}(\boldsymbol{u})\cdot\delta\boldsymbol{u}d\Omega+\int_{\Omega}\frac{1}{2}\boldsymbol{u}\cdot\boldsymbol{P}(\delta\boldsymbol{u})d\Omega+b(\boldsymbol{u},\delta\boldsymbol{u})
	\end{aligned}
\end{equation}
Then, using the linear property of $\boldsymbol{P}$:
\begin{equation}
	\begin{aligned}
		\int_{\Omega}\boldsymbol{P}(\boldsymbol{u})\cdot\delta\boldsymbol{u}d\Omega &= \int_{\Omega}\frac{1}{2}\boldsymbol{P}(\boldsymbol{u})\cdot\delta\boldsymbol{u}d\Omega+\int_{\Omega}\frac{1}{2}\boldsymbol{u}\cdot\delta\boldsymbol{P}(\boldsymbol{u})d\Omega+b(\boldsymbol{u},\delta\boldsymbol{u}) \\
		&= \delta\left(\int_{\Omega}\frac{1}{2}\boldsymbol{P}(\boldsymbol{u})\cdot\boldsymbol{u}d\Omega\right)+b(\boldsymbol{u},\delta\boldsymbol{u})
	\end{aligned}
	\label{eq:linear_self_adjoint}
\end{equation}
Substituting \Cref{eq:linear_self_adjoint} into \Cref{eq:Garerkin_form_PDEs}:
\begin{equation}
	\int_{\Omega}[\boldsymbol{P}(\boldsymbol{u})-\boldsymbol{f}]\cdot\delta\boldsymbol{u}d\Omega-\int_{\Gamma}[\boldsymbol{B}(\boldsymbol{u})-\boldsymbol{g}]\cdot\delta\boldsymbol{u}d\Gamma = \delta\left(\int_{\Omega}[\frac{1}{2}\boldsymbol{P}(\boldsymbol{u})-\boldsymbol{f}]\cdot\boldsymbol{u}d\Omega\right)+b(\boldsymbol{u},\delta\boldsymbol{u})-\int_{\Gamma}[\boldsymbol{B}(\boldsymbol{u})-\boldsymbol{g}]\cdot\delta\boldsymbol{u}d\Gamma
	\label{eq:functional_PDEs}
\end{equation}
Usually, the two distribution integral terms can be written as a boundary integral functional, i.e.,
\begin{equation}
\delta(\Gamma(\boldsymbol{u}))=b(\boldsymbol{u},\delta\boldsymbol{u})-\int_{\Gamma}[\boldsymbol{B}(\boldsymbol{u})-\boldsymbol{g}]\cdot\delta\boldsymbol{u}d\Gamma
\end{equation}
Therefore, \Cref{eq:functional_PDEs} can be transformed into a variational form of a functional,
\begin{equation}
	\begin{aligned}
		\delta\Pi &= \int_{\Omega}[\boldsymbol{P}(\boldsymbol{u})-\boldsymbol{f}]\cdot\delta\boldsymbol{u}d\Omega-\int_{\Gamma}[\boldsymbol{B}(\boldsymbol{u})-\boldsymbol{g}]\cdot\delta\boldsymbol{u}d\Gamma \\
		\Pi &= \int_{\Omega}[\frac{1}{2}\boldsymbol{P}(\boldsymbol{u})-\boldsymbol{f}]\cdot\boldsymbol{u}d\Omega+\Gamma(\boldsymbol{u})
	\end{aligned}
\end{equation}
Thus, solving the strong form of PDEs (PINNs) is equivalent to solving the stationary value problem of the functional. 
If the operator $\boldsymbol{P}$ is of 2m even order and is linear and self-adjoint, the stationary value problem can be further transformed into an extremum problem (DEM), which means $\delta^{2}\Pi>0$ to guarantee a positive definite matrix.

\section{Integration strategies in BINN} \label{sec:app_integral}
To evaluate the residuals in \Cref{eq:BINN_loss_funciton}, we should determine how to allocate the source points $\boldsymbol{y}_i$ on the boundary and evaluate the integrals with singularity. Following \cite{sun2023binn}, a piece-wise integration strategy with regularization techniques is applied: 
The whole boundary $\Gamma$ will be divided into $N_s$ segments: $\Gamma = \bigcup\limits_{i=1}\limits^{N_s}\Gamma_i $, with $N_s$ source points located on the center of each segment. Let  $\boldsymbol{y}_i$ denote the source point on $\Gamma_i $, then integrals will be evaluated piece-wisely on each $\Gamma_i$.

For source point $\boldsymbol{y}_i$, the integrals on $\Gamma_j  (i \neq j)$ are always regular. In the present work, those integrals are computed with the Gaussian quadrature rule. 

Weakly-singular integrals that involve $u^f(\boldsymbol{x},\boldsymbol{y}_i)$ in \Cref{eq:fundamental_solution} on $\Gamma_i$ can be written into the following form (Note that the source point $\boldsymbol{y}_i$ is located exactly at the center of $\Gamma_i$):
\begin{equation}
	\int_{-a}^{a} \ln{|t|}f(t) dt
	\label{eq:weak}
\end{equation}
where $t$ is the parametric coordinate on $\Gamma_i$, $f(t)$ is a regular term containing boundary conditions ($\bar{u}(\boldsymbol{x})$ or $\bar{t}(\boldsymbol{x})$), neural networks $\phi(\boldsymbol{x}; \boldsymbol{\theta})$, Jacobians. $a$ is the half-length of the segment $\Gamma_i$, and the length is in the parametric coordinate system. 

We use integration by parts to transform \Cref{eq:weak} into:
\begin{equation}
	\begin{aligned}
		\int_{-a}^{a}\ln|t|f(t) \, dt & = \int_{-a}^{a}\ln(|t|)[f(t)-f(0)] \, dt + f(0)\int_{-a}^{a}\ln(|t|) \, dt\\
		& = \int_{-a}^{a}\ln(|t|)[f(t)-f(0)] \, dt + 2f(0)\int_{0}^{a}\ln(t) \, dt\\
		& = \int_{-a}^{a}\ln(|t|)[f(t)-f(0)] \, dt + 2f(0)[a\ln(a)-a]
	\end{aligned}
	\label{eq:weak_integration}
\end{equation}

Considering the first term on the right-hand side of \Cref{eq:weak_integration}, we expand \(f(t) - f(0)\) using the Taylor series:

\begin{equation}
	\begin{aligned}
		\int_{-a}^{a}\ln(|t|)[f(t)-f(0)] \, dt & = \int_{-a}^{a}\ln(|t|)[f'(0)t+\frac{f''(0)}{2}t^2 + O(t^3)] \, dt\\
		& = \int_{-a}^{a}\ln(|t|)t\left[f'(0)+\frac{f''(0)}{2}t + O(t^2)\right] \, dt
	\end{aligned}
	\label{eq:weak_integration_taylor}
\end{equation}

where \(O(t^3)\) includes the higher-order terms of \(t^3\) and above. We can see that \(\ln(|t|)t\) approaches zero as \(t\) approaches zero. Clearly, this eliminates the singularity in the integral \(\ln|t|f(t)\) in \Cref{eq:weak_integration}. Therefore, we only need to use Gaussian quadrature to handle \Cref{eq:weak_integration}.

Next, we analyze the Cauchy-principal integrals. Using the substitution \(t = -s\), we get:

\begin{equation}
	\begin{aligned}
		\int_{-a}^{a}\frac{1}{t}f(t) \, dt & = \int_{-a}^{0}\frac{1}{t}f(t) \, dt + \int_{0}^{a}\frac{1}{t}f(t) \, dt\\
		& = \int_{a}^{0}\frac{1}{-s}f(-s)(-ds) + \int_{0}^{a}\frac{1}{t}f(t) \, dt\\
		& = -\int_{0}^{a}\frac{1}{s}f(-s) \, ds + \int_{0}^{a}\frac{1}{t}f(t) \, dt\\
		& = \int_{0}^{a}\frac{1}{t}[f(t)-f(-t)] \, dt
	\end{aligned}
	\label{eq:Cauchy_singular}
\end{equation}

We can see that \Cref{eq:Cauchy_singular} eliminates the strong singularity. Using the Taylor expansion, we get:

\begin{equation}
	\begin{aligned}
		\int_{-a}^{a}\frac{1}{t}f(t) \, dt & = \int_{0}^{a}\frac{1}{t}[f(t)-f(-t)] \, dt\\
		& = \int_{0}^{a}\frac{1}{t}[2f'(0)t + t^2f''(0) + O(t^3)] \, dt\\
		& = \int_{0}^{a}[2f'(0) + tf''(0) + O(t^2)] \, dt
	\end{aligned}
	\label{eq:Cauchy_singular_prove}
\end{equation}

Therefore, we only need to apply Gaussian quadrature to the right-hand term in \Cref{eq:Cauchy_singular}.

In summary, \Cref{eq:weak_integration} and \Cref{eq:Cauchy_singular} are mathematical transformations for handling weakly-singular integrals and Cauchy-principal integrals, respectively in BINN to address singularities.

\section{Similarities between KAN and Finite Elements Method\label{sec:Similarities-KAN_FEM}}

For simplicity, let's first consider a 1D linear finite element, with \(N\) elements. The approximation function is
\begin{equation}
u^{fem}(x)=\sum_{i=1}^{N+1}N_{i}(x)u_{i}
\end{equation}
where \(N(x)\) is the shape function in finite elements, and \(u_{i}\) is the nodal displacement. On the other hand, consider KAN with structure \([1,1]\) and grid size=N. The mathematical structure of KAN is:
\begin{equation}
u^{kan}(x)=K(x)=\sum_{i=1}^{N+k}B_{i}(x)c_{i}
\end{equation}
where \(B_{i}(x)\) is the basis function of B-splines. For simplicity, we first disregard the additional residual function \(\boldsymbol{W} \cdot \sigma(\boldsymbol{X})\) and the scaling factors \(\boldsymbol{S}\) for B-splines, as shown in \Cref{eq:function_each_kan}. If we choose first-order splines, i.e., \(k=1\), then we find that
\begin{equation}
B_{i}(x)=N_{i}(x)
\end{equation}
This means that KAN, in this special structure, is equivalent to a linear 1D finite element. In higher orders, KAN and FEM differ. For example, consider a quadratic element with shape functions:
\begin{equation}
	\begin{aligned}
		N_{1} &= \frac{x(x-1)}{2} \\
		N_{2} &= (x+1)(1-x) \\
		N_{3} &= \frac{(x+1)x}{2}
	\end{aligned}
\end{equation}
On the other hand, if we consider \(k=2\) quadratic B-splines, and control the number of nodes to be the same (grid size is 2), for \(-1 \leq x < 0\):
\begin{equation}
	\begin{aligned}
		B_{1} &= \frac{x^{2}}{2} \\
		B_{2} &= -x^{2}-x+\frac{1}{2} \\
		B_{3} &= \frac{1}{2}x^{2}+x+\frac{1}{2} \\
		B_{4} &= 0
	\end{aligned}
\end{equation}
and for \(0 \leq x < 1\):
\begin{equation}
	\begin{aligned}
		B_{1} &= 0 \\
		B_{2} &= \frac{1}{2}x^{2}-x+\frac{1}{2} \\
		B_{3} &= -x^{2}+x+\frac{1}{2} \\
		B_{4} &= \frac{x^{2}}{2}
	\end{aligned}
\end{equation}

Although it seems that KAN and quadratic FEM differ because order=2, grid size=2 KAN has four coefficients to determine while FEM has only three, we prove that KAN can degenerate to FEM. Setting \(u^{fem}(x)=u^{kan}(x)\) for \(-1 \leq x < 0\):
\begin{equation}
	\begin{aligned}
		\sum_{i=1}^{3}N_{i}(x)u_{i} &= \sum_{i=1}^{4}B_{i}(x)c_{i} \\
		\frac{x(x-1)}{2}u_{1}+(x+1)(1-x)u_{2}+\frac{(x+1)x}{2}u_{3} &= \frac{x^{2}}{2}c_{1}+(-x^{2}-x+\frac{1}{2})c_{2}+(\frac{1}{2}x^{2}+x+\frac{1}{2})c_{3} \\
		(\frac{1}{2}u_{1}-u_{2}+\frac{1}{2}u_{3})x^{2}+(-\frac{1}{2}u_{1}+\frac{1}{2}u_{3})x+u_{2} &= (\frac{1}{2}c_{1}-c_{2}+\frac{1}{2}c_{3})x^{2}+(-c_{2}+c_{3})x+\frac{1}{2}c_{2}+\frac{1}{2}c_{3}
	\end{aligned}
\end{equation}
We obtain the relationship:
\begin{equation}
\left[\begin{array}{ccc}
	\frac{1}{2} & -1 & \frac{1}{2} \\
	0 & -1 & 1 \\
	0 & \frac{1}{2} & \frac{1}{2}
\end{array}\right]\left[\begin{array}{c}
	c_{1} \\
	c_{2} \\
	c_{3}
\end{array}\right] = \left[\begin{array}{ccc}
	\frac{1}{2} & -1 & \frac{1}{2} \\
	-\frac{1}{2} & 0 & \frac{1}{2} \\
	0 & 1 & 0
\end{array}\right]\left[\begin{array}{c}
	u_{1} \\
	u_{2} \\
	u_{3}
\end{array}\right]
\end{equation}

\begin{equation}
	\left[\begin{array}{c}
		c_{1}\\
		c_{2}\\
		c_{3}
	\end{array}\right]=\left[\begin{array}{ccc}
		\frac{1}{2} & -1 & \frac{1}{2}\\
		0 & -1 & 1\\
		0 & \frac{1}{2} & \frac{1}{2}
	\end{array}\right]^{-1}\left[\begin{array}{ccc}
		\frac{1}{2} & -1 & \frac{1}{2}\\
		-\frac{1}{2} & 0 & \frac{1}{2}\\
		0 & 1 & 0
	\end{array}\right]\left[\begin{array}{c}
		u_{1}\\
		u_{2}\\
		u_{3}
	\end{array}\right]=\left[\begin{array}{ccc}
		1.75 & -1 & 0.25\\
		0.25 & 1 & -0.25\\
		-0.25 & 1 & 0.25
	\end{array}\right]\left[\begin{array}{c}
		u_{1}\\
		u_{2}\\
		u_{3}
	\end{array}\right]\label{eq:left_domain_order2}
\end{equation}

Similarly, for \(0 \leq x < 1\):
\begin{equation}
	\begin{aligned}
		\sum_{i=1}^{3}N_{i}(x)u_{i} &= \sum_{i=1}^{4}B_{i}(x)c_{i} \\
		\frac{x(x-1)}{2}u_{1}+(x+1)(1-x)u_{2}+\frac{(x+1)x}{2}u_{3} &= (\frac{1}{2}x^{2}-x+\frac{1}{2})c_{2}+(-x^{2}+x+\frac{1}{2})c_{3}+(\frac{x^{2}}{2})c_{4} \\
		(\frac{1}{2}u_{1}-u_{2}+\frac{1}{2}u_{3})x^{2}+(-\frac{1}{2}u_{1}+\frac{1}{2}u_{3})x+u_{2} &= (\frac{1}{2}c_{2}-c_{3}+\frac{1}{2}c_{4})x^{2}+(-c_{2}+c_{3})x+\frac{1}{2}c_{2}+\frac{1}{2}c_{3}
	\end{aligned}
\end{equation}
We obtain the relationship:
\begin{equation}
\left[\begin{array}{ccc}
	\frac{1}{2} & -1 & \frac{1}{2} \\
	-1 & 1 & 0 \\
	\frac{1}{2} & \frac{1}{2} & 0
\end{array}\right]\left[\begin{array}{c}
	c_{2} \\
	c_{3} \\
	c_{4}
\end{array}\right] = \left[\begin{array}{ccc}
	\frac{1}{2} & -1 & \frac{1}{2} \\
	-\frac{1}{2} & 0 & \frac{1}{2} \\
	0 & 1 & 0
\end{array}\right]\left[\begin{array}{c}
	u_{1} \\
	u_{2} \\
	u_{3}
\end{array}\right]
\end{equation}
\begin{equation}
	\left[\begin{array}{c}
		c_{2} \\
		c_{3} \\
		c_{4}
	\end{array}\right] = \left[\begin{array}{ccc}
		0.25 & 1 & -0.25 \\
		-0.25 & 1 & 0.25 \\
		0.25 & -1 & 1.75
	\end{array}\right]\left[\begin{array}{c}
		u_{1} \\
		u_{2} \\
		u_{3}
	\end{array}\right]\label{eq:right_domain_order2}
\end{equation}

Comparing \Cref{eq:left_domain_order2} and \Cref{eq:right_domain_order2}, we find that \(\{c_{1}, c_{2}, c_{3}, c_{4}\}\) can be completely determined by \(\{u_{1}, u_{2}, u_{3}\}\), i.e.,
\begin{equation}
\left[\begin{array}{c}
	c_{1} \\
	c_{2} \\
	c_{3} \\
	c_{4}
\end{array}\right] = \left[\begin{array}{ccc}
	1.75 & -1 & 0.25 \\
	0.25 & 1 & -0.25 \\
	-0.25 & 1 & 0.25 \\
	0.25 & -1 & 1.75
\end{array}\right]\left[\begin{array}{c}
	u_{1} \\
	u_{2} \\
	u_{3}
\end{array}\right]
\end{equation}
This demonstrates that  KAN can completely encompass quadratic finite elements when order=2, and grid size=2. For higher dimensions instead of 1D, this conclusion holds to some extent. For example, consider a simple 2D linear element. With uniform N grids in both x and y directions, the FEM approximation is:
\begin{equation}
	u(\boldsymbol{x})=\sum_{i=1}^{(N+1)^{2}}N_{i}(\boldsymbol{x})u_{i}\label{eq:2d_fem}
\end{equation}
where \(N_{i}(\boldsymbol{x})\) is the shape function:
\begin{equation}
N_{i}(\boldsymbol{x})=N_{i}^{x}(x)\cdot N_{i}^{y}(y)
\end{equation}
For KAN with structure \([2,1]\), grid size=N, order=1, the KAN function is:
\begin{equation}
	K(\boldsymbol{x})=\sum_{i=1}^{N+1}B_{i}(x)c_{i}^{x}+\sum_{i=1}^{N+1}B_{i}(y)c_{i}^{y}=\sum_{i=1}^{N+1}N_{i}^{x}(x)c_{i}^{x}+\sum_{i=1}^{N+1}N_{i}^{y}(x)c_{i}^{y}\label{eq:2d_kan}
\end{equation}
Comparing \Cref{eq:2d_fem} and \Cref{eq:2d_kan}, we find significant differences in the forms of FEM and KAN. However, if we replace the additive combination of KAN activation functions in \Cref{eq:2d_kan} with a multiplicative combination, it would exactly match the 2D function representation of FEM:

\begin{equation}
	K_{modified}(\boldsymbol{x})=\sum_{i=1}^{N+1}B_{i}(x)c_{i}^{x}*\sum_{i=1}^{N+1}B_{i}(y)c_{i}^{y}=\sum_{i=1}^{N+1}N_{i}^{x}(x)c_{i}^{x}*\sum_{i=1}^{N+1}N_{i}^{y}(x)c_{i}^{y} = \sum_{i=1}^{(N+1)^{2}}N_{i}(\boldsymbol{x})c_{i}^{x}c_{i}^{y}.
\end{equation}

This suggests that KAN could be transformed into a Deep Finite Element Method by adjusting the activation functions from additive to multiplicative combinations. Note that the above proof is based on the very shallow layers of KAN. With deeper KAN structures, we believe KAN will exhibit much stronger fitting capabilities compared to FEM. This proof mainly illustrates the similarity between FEM and KAN, with the fundamental reason being that FEM's fitting functions are spline interpolations. For example, linear elements correspond to first-order splines, making FEM and KAN similar. Since KAN is composed of nested B-spline functions, KAN is mathematically a nested shape function in  FEM:
\begin{equation}
S(S(\cdots S(\boldsymbol{x})))
\end{equation}
where \(S\) is the shape function in FEM.

Considering that B-splines are essentially a degenerate form of NURBS in IGA, where B-splines are modified to be non-uniform and combined with weight functions to form NURBS, mathematically, KAN  is quite similar to nested NURBS in IGA. Currently, KAN still uses B-splines, but in the future, it could be replaced by NURBS instead of B-splines.

As a result, KAN is similar to nested functions of shape functions in FEM and NURBS in IGA.

\section{The highest order of PDEs that KINN can solve\label{sec:The-highest-order}}

The key to solving high-order PDEs using neural networks is ensuring that higher-order derivatives are non-zero. We analyze the applicability conditions of KINN for high-order PDEs based on the structure of KAN.

The highest order of PDEs that KINN can solve is related to the order of the B-splines and the activation function. We consider the following single-layer KINN computation structure:
\begin{equation}
	\boldsymbol{X}^{(\text{new})}=K(\boldsymbol{X}^{(\text{old})})=\tanh\left\{[\sum_{\text{column}}\boldsymbol{\phi}(\boldsymbol{X}^{(\text{old})})\odot\boldsymbol{S}]+\boldsymbol{W}\cdot\sigma(\boldsymbol{X}^{(\text{old})})\right\}.
\end{equation}
If KINN has an $N+1$ layer network structure $[n_{0}, n_{1}, n_{2}, \cdots, n_{N}]$, then the overall computation is
\begin{equation}
	\boldsymbol{X}^{(N)}=K^{(N)}\circ\cdots\circ K^{(2)}\circ K^{(1)}(\boldsymbol{X}^{(0)}).
\end{equation}
Using the chain rule, we can derive the derivative of the output with respect to the input, which approximates the differential operator of PDEs:
\begin{equation}
	\frac{\partial \boldsymbol{X}^{(N)}}{\partial \boldsymbol{X}^{(0)}}=\frac{\partial K^{(N)}(\boldsymbol{X}^{(N-1)})}{\partial \boldsymbol{X}^{(N-1)}}\frac{\partial K^{(N-1)}(\boldsymbol{X}^{(N-2)})}{\partial \boldsymbol{X}^{(N-2)}}\cdots\frac{\partial K^{(2)}(\boldsymbol{X}^{(1)})}{\partial \boldsymbol{X}^{(1)}}\frac{\partial K^{(1)}(\boldsymbol{X}^{(0)})}{\partial \boldsymbol{X}^{(0)}}.\label{eq:chain_rule}
\end{equation}
It is clear from \Cref{eq:chain_rule} that it involves the multiplication of similar tensors. We analyze one term:
\begin{align}
	\frac{\partial K^{(I+1)}(\boldsymbol{X}^{(I)})}{\partial \boldsymbol{X}^{(I)}} & =\frac{\partial \tanh(\boldsymbol{Y}^{(I)})}{\partial \boldsymbol{Y}^{(I)}}\frac{\partial \boldsymbol{Y}^{(I)}}{\partial \boldsymbol{X}^{(I)}}\label{eq:every_chain_rule}\\
	\boldsymbol{Y}^{(I)} & =[\sum_{\text{column}}\boldsymbol{\phi}(\boldsymbol{X}^{(I)})\odot\boldsymbol{S}]+\boldsymbol{W}\cdot\sigma(\boldsymbol{X}^{(I)})\nonumber 
\end{align}
For convenience, we use tensor operations, obtaining:
\begin{align}
	\frac{\partial \boldsymbol{Y}^{(I)}}{\partial \boldsymbol{X}^{(I)}} & =\frac{\partial [S_{i^{*}j}\phi_{ij}(\boldsymbol{X}^{(I)})+W_{ij}\sigma(X_{j}^{(I)})]}{\partial \boldsymbol{X}^{(I)}}=S_{i^{*}j}\frac{\partial \phi_{ij}(\boldsymbol{X}^{(I)})}{\partial \boldsymbol{X}^{(I)}}+W_{ij}\cdot\frac{\partial \sigma(X_{j}^{(I)})}{\partial \boldsymbol{X}^{(I)}} \label{eq:derivative_y_x}
\end{align}
where the {*} denotes terms not included in the summation. Substituting the B-spline formula into \Cref{eq:derivative_y_x}:
\begin{align}
	\frac{\partial \boldsymbol{Y}^{(I)}}{\partial \boldsymbol{X}^{(I)}} & =S_{i^{*}j}\frac{\partial c_{m}^{(i,j)}B_{m}(X_{j^{*}})}{\partial \boldsymbol{X}^{(I)}}+W_{ij}\cdot\frac{\partial \sigma(X_{j}^{(I)})}{\partial \boldsymbol{X}^{(I)}}=S_{i^{*}j}[c_{m}^{(i,j)}\frac{\partial B_{m}(X_{j^{*}})}{\partial \boldsymbol{X}^{(I)}}]+W_{ij}\frac{\partial \sigma(X_{j}^{(I)})}{\partial \boldsymbol{X}^{(I)}}\nonumber \\
	\frac{\partial Y_{i}^{(I)}}{\partial X_{j}^{(I)}} & =S_{i^{*}k}[c_{m}^{(i,k)}\frac{\partial B_{m}(X_{k^{*}})}{\partial X_{j}^{(I)}}]+W_{ik}\frac{\partial \sigma(X_{k}^{(I)})}{\partial X_{j}^{(I)}}\label{eq:Y_X}
\end{align}
Substituting \Cref{eq:Y_X} into \Cref{eq:every_chain_rule}, we find:
\begin{align}
	\frac{\partial K_{i}^{(I+1)}(\boldsymbol{X}^{(I)})}{\partial X_{j}^{(I)}} & =\frac{\partial \tanh(Y_{i}^{(I)})}{\partial Y_{m}^{(I)}}\frac{\partial Y_{m}^{(I)}}{\partial X_{j}^{(I)}}=\frac{\partial \tanh(Y_{i}^{(I)})}{\partial Y_{m}^{(I)}}S_{m^{*}k}[c_{q}^{(m,k)}\frac{\partial B_{q}(X_{k^{*}})}{\partial X_{j}^{(I)}}]+\frac{\partial \tanh(Y_{i}^{(I)})}{\partial Y_{m}^{(I)}}W_{mk}\frac{\partial \sigma(X_{k}^{(I)})}{\partial X_{j}^{(I)}}\label{eq:tensor_ana_chain_rule}
\end{align}
From \Cref{eq:tensor_ana_chain_rule}, we see that the activation function $\tanh$ ensures that the first-order derivative of the single-layer KINN chain rule is non-zero. Since \Cref{eq:chain_rule} involves multiple tensor multiplications from \Cref{eq:tensor_ana_chain_rule}, the product rule of differentiation ensures that higher-order derivatives remain non-zero if each term in \Cref{eq:tensor_ana_chain_rule} is non-zero. The smoothness of $\tanh$ guarantees that the higher-order tensor derivatives are non-zero.

If we remove the $\tanh$ activation function, we get:
\begin{align}
	\frac{\partial K_{i}^{(I+1)}(\boldsymbol{X}^{(I)})}{\partial X_{j}^{(I)}} & =\delta_{im}S_{m^{*}k}[c_{q}^{(m,k)}\frac{\partial B_{q}(X_{k^{*}})}{\partial X_{j}^{(I)}}]+\delta_{im}W_{mk}\frac{\partial \sigma(X_{k}^{(I)})}{\partial X_{j}^{(I)}}\label{eq:tensor_ana_chain_rule_non_tanh}
\end{align}
We find that \Cref{eq:tensor_ana_chain_rule_non_tanh} can still be non-zero due to $\partial \sigma(X_{k}^{(I)})/\partial X_{j}^{(I)}$ if $\sigma(X_{k}^{(I)})$ is infinitely smooth. However, this might compromise the fitting ability of the B-splines because $\partial B_{q}(X_{k^{*}})/\partial X_{j}^{(I)}$ could be zero, depending on the B-spline order, the number of KAN layers, and the highest order of the PDEs.

If we use only B-spline functions to construct KAN, such as:
\begin{equation}
	\boldsymbol{X}^{(\text{new})}=K(\boldsymbol{X}^{(\text{old})})=\sum_{\text{column}}[\boldsymbol{\phi}(\boldsymbol{X}^{(\text{old})})\odot \boldsymbol{S}],
\end{equation}
we might easily get zero for higher-order derivatives. For example, if the B-spline order is 3 and the KAN network structure is $[2,5,1]$, the maximum order of KAN is $3*(3-1)=6$. Therefore, solving derivatives higher than the 6th order would result in zero derivatives, rendering the algorithm ineffective. Hence, each layer's $\tanh$ activation function not only pulls the output back within the grid size $[-1,1]$ to better utilize the B-splines but also ensures there is no limit to the order of PDEs that can be solved.

Based on the above analysis, \Cref{tab:Ana_KINN} summarizes the core functions of each component in KINN.

\begin{figure}
	\centering
	\includegraphics[scale=0.8]{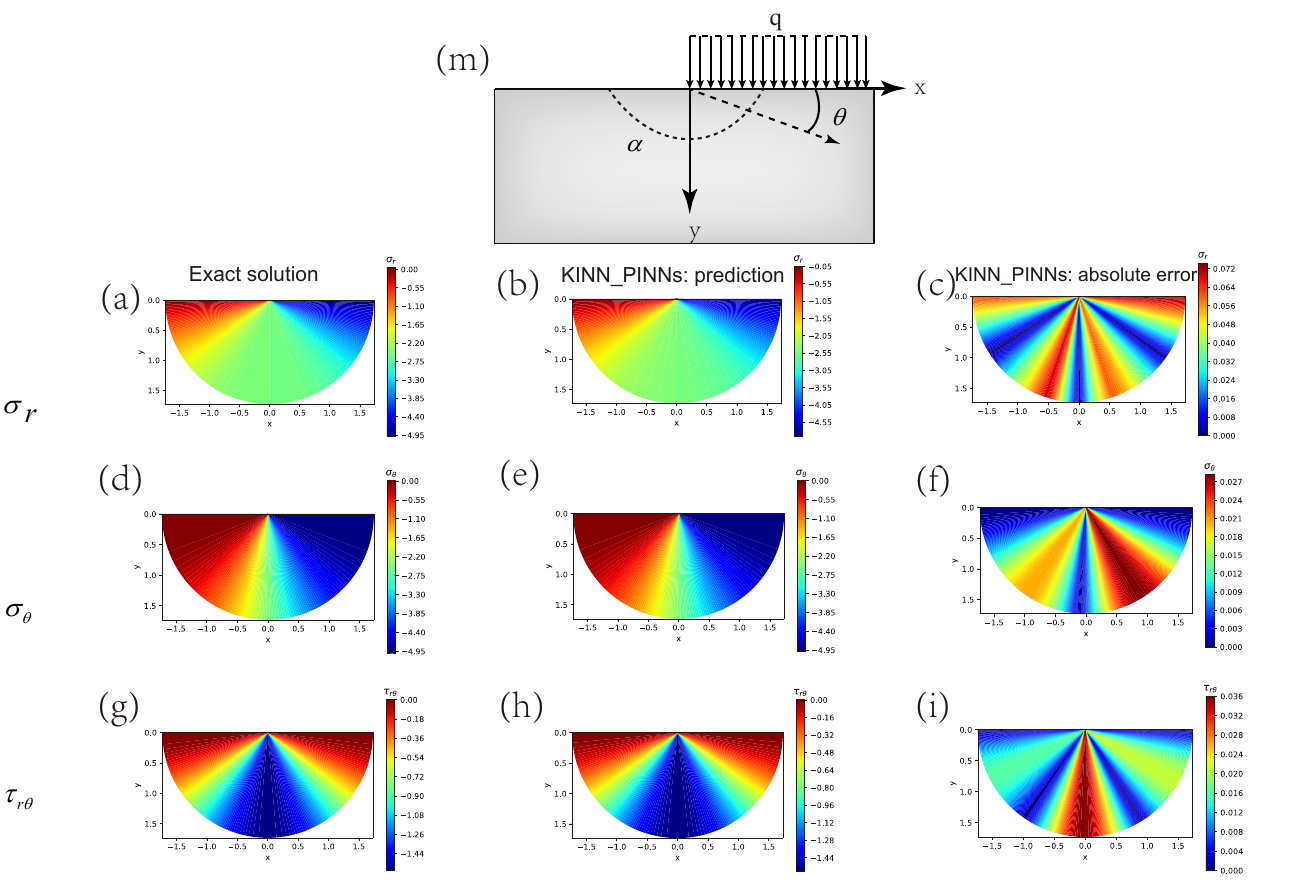}
	\caption{The infinite wedge problem and the performance of the KINN algorithm: (m) Schematic of the infinite wedge problem. (a,b,c) Analytical solution, KINN prediction, and absolute error of $\sigma_{r}$. (d,e,f) Analytical solution, KINN prediction, and absolute error of $\sigma_{\theta}$. (g,h,i) Analytical solution, KINN prediction, and absolute error of $\tau_{r\theta}$.}
	\label{fig:wedge}
\end{figure}

To verify our theory, we solve a one-dimensional fourth-order PDE problem:
\begin{equation}
	\begin{cases}
		\nabla^{4}\phi=0 & \theta\in[0,\pi]\\
		\sigma_{\theta}=\frac{\partial^{2}\phi}{\partial r^{2}}=-q & \theta=0\\
		\sigma_{\theta}=\frac{\partial^{2}\phi}{\partial r^{2}}=0 & \theta=\pi\\
		\tau_{r\theta}=-\frac{\partial}{\partial r}\left(\frac{1}{r}\frac{\partial\phi}{\partial\theta}\right)=0 & \theta=0\\
		\tau_{r\theta}=-\frac{\partial}{\partial r}\left(\frac{1}{r}\frac{\partial\phi}{\partial\theta}\right)=0 & \theta=\pi
	\end{cases}
\end{equation}
This is the problem of an infinite wedge \cite{wang2023dcm}, as shown in \Cref{fig:wedge}m, where $\phi$ is the Airy stress function. The analytical solution to this problem is
\begin{equation}
	\phi=cr^{2}\left[\alpha - \theta + \sin(\theta)\cos(\theta) - \cos^{2}(\theta)\tan(\alpha)\right]
\end{equation}
where $c=\frac{q}{2(\tan(\alpha) - \alpha)}$, $\alpha$ is the angle of the infinite wedge, here $\alpha=\pi$, $q=5$, $E=1000$, and $\mu=0.3$.

We solve this problem using KINN-PINNs. \Cref{fig:wedge} shows the comparison between the KINN-PINNs predictions and the analytical solution. From the absolute error, we can see that KINN can solve this problem well. However, if we modify the structure of KAN by removing the activation function $\tanh$ and the residual term $\boldsymbol{W} \cdot \sigma(\boldsymbol{X})$:
\begin{equation}
	\boldsymbol{Y}^{\text{new}} = \tanh(\boldsymbol{Y}) = \sum_{\text{column}}[\boldsymbol{\phi}(\boldsymbol{X}) \odot \boldsymbol{S}]
\end{equation}
and set the structure of KAN to {[}1,1{]} with a spline order of 3, we find that KINN cannot compute the fourth-order derivative. This result is consistent with our theoretical derivation. Therefore, when using KAN to solve PDEs, it is crucial to include the activation function $\tanh$ and the residual term $\boldsymbol{W} \cdot \sigma(\boldsymbol{X})$ to ensure the feasibility of solving high-order PDEs.

\begin{table}
	\caption{The role of each function in KINN: $\boldsymbol{X}^{(\text{new})}=K(\boldsymbol{X}^{(\text{old})})=\tanh\{\sum_{\text{column}}[\boldsymbol{\phi}(\boldsymbol{X}^{(\text{old})})\odot \boldsymbol{S}]+\boldsymbol{W}\cdot \sigma(\boldsymbol{X}^{(\text{old})})\}$\label{tab:Ana_KINN}}
	
	\centering{}%
	\begin{adjustbox}{max width=\textwidth}
	\begin{tabular}{cc}
		\toprule 
		\multirow{1}{*}{Function in KINN} & Description\tabularnewline
		\midrule 
		$\boldsymbol{\phi}$ & Enhances KAN's ability to fit the activation function by adding B-splines\tabularnewline
		$\sigma$ & Residual term in KAN, similar to the ResNet\tabularnewline
		$\boldsymbol{S}$ & Scaling factor for $\boldsymbol{\phi}$\tabularnewline
		$\boldsymbol{W}$ & Scaling factor for $\sigma$\tabularnewline
		$\tanh$ & Improves function smoothness, scales output within the grid size, enhancing the effectiveness of B-splines\tabularnewline
		\bottomrule
	\end{tabular}
	\end{adjustbox}
\end{table}

\section{Scale normalization in Deep Energy Method}\label{sec:scale_normal}

Scale normalization is a crucial operation in KINN because it ensures that the input remains within the range of the B-spline basis functions, while we keep the grid size of the B-splines fixed at $[-1,1]$. However, special attention needs to be given to  DEM. When the characteristic size is particularly large, we consider \Cref{fig:plate_hole_intro} to illustrate this issue. Without scale normalization, this problem often uses the following admissible functions:
\begin{align}
	u_{x} & =x\phi_{x}(\boldsymbol{x};\boldsymbol{\theta})\label{eq:u_x}\\
	u_{y} & =y\phi_{y}(\boldsymbol{x};\boldsymbol{\theta})\label{eq:u_y}
\end{align}
Without scale normalization, large coordinate gradients may arise during differentiation. For instance, consider the normal strain $\varepsilon_{xx}$ obtained by differentiating \Cref{eq:u_x} with respect to $x$:
\begin{equation}
	\varepsilon_{xx}=\frac{\partial u_{x}}{\partial x}=\phi_{x}(\boldsymbol{x};\boldsymbol{\theta})+x\frac{\partial\phi_{x}(\boldsymbol{x};\boldsymbol{\theta})}{\partial x}\label{eq:derivative}
\end{equation}
In \Cref{eq:derivative}, the term $x\frac{\partial\phi_{x}(\boldsymbol{x};\boldsymbol{\theta})}{\partial x}$ can cause the normal strain $\varepsilon_{xx}$ to become excessively large. Note that $\phi_{x}(\boldsymbol{x};\boldsymbol{\theta})$ and $\frac{\partial\phi_{x}(\boldsymbol{x};\boldsymbol{\theta})}{\partial x}$ do not typically increase significantly with size. This is because $\phi_{x}(\boldsymbol{x};\boldsymbol{\theta})$ often includes a $\tanh$ activation function in the hidden layers, which scales the output to $[-1,1]$. The derivative $\frac{\partial\phi_{x}(\boldsymbol{x};\boldsymbol{\theta})}{\partial x}$ is related to the weights of the initial linear transformation in a simple fully connected neural network:
\begin{equation}
	\begin{aligned}\frac{\partial\phi_{x}(\boldsymbol{x};\boldsymbol{\theta})}{\partial x} & =\frac{\partial y^{(N)}(a^{(N-1)})}{\partial\boldsymbol{a}^{(N-1)}}\frac{\partial\boldsymbol{a}^{(N-1)}(\boldsymbol{y}^{(N-1)})}{\partial\boldsymbol{y}^{(N-1)}}\frac{\partial\boldsymbol{y}^{(N-1)}(\boldsymbol{a}^{(N-2)})}{\partial\boldsymbol{a}^{(N-2)}}\cdots\frac{\partial\boldsymbol{a}^{(1)}(\boldsymbol{y}^{(1)})}{\partial\boldsymbol{y}^{(1)}}\frac{\partial\boldsymbol{y}^{(1)}(x)}{\partial x}\\
		& =\boldsymbol{W}^{(N)}\sigma^{'}(\boldsymbol{y}^{(N-1)})\boldsymbol{W}^{(N-1)}\cdots\sigma^{'}(\boldsymbol{y}^{(1)})\boldsymbol{W}^{(1)}
	\end{aligned}
	\label{eq:nn_deri}
\end{equation}
where $\sigma^{'}(y^{(N-1)})$ is the derivative of the activation function with respect to $y^{(N-1)}$. If the activation function is $\tanh$, this term will not exceed 1. Hence, the range of \Cref{eq:nn_deri} is:
\begin{align}
	\left|\frac{\partial\phi_{x}(\boldsymbol{x};\boldsymbol{\theta})}{\partial x}\right| & \leq\left|\boldsymbol{W}^{(N)}\boldsymbol{W}^{(N-1)}\cdots\boldsymbol{W}^{(1)}\right|\label{eq:nn_deri-1}
\end{align}
The derivative $\frac{\partial\phi_{x}(\boldsymbol{x};\boldsymbol{\theta})}{\partial x}$ is related to the weights of the linear transformation and does not exhibit a size effect. Therefore, the magnitude of \Cref{eq:derivative} is mainly influenced by the term $x\frac{\partial\phi_{x}(\boldsymbol{x};\boldsymbol{\theta})}{\partial x}$, which depends on the size of $x$. To address this issue, we introduce scale normalization, modifying \Cref{eq:u_x} and \Cref{eq:u_y} as follows:
\begin{align}
	u_{x} & =x\phi_{x}\left(\frac{\boldsymbol{x}}{L};\boldsymbol{\theta}\right)\label{eq:u_x-scale}\\
	u_{y} & =y\phi_{y}\left(\frac{\boldsymbol{x}}{L};\boldsymbol{\theta}\right)\label{eq:u_y-scale}
\end{align}
where $L$ is the characteristic size of the object being simulated. Differentiating these expressions yields:
\begin{equation}
	\varepsilon_{xx}=\frac{\partial u_{x}}{\partial x}=\phi_{x}\left(\frac{\boldsymbol{x}}{L};\boldsymbol{\theta}\right)+\frac{x}{L}\frac{\partial\phi_{x}\left(\frac{\boldsymbol{x}}{L};\boldsymbol{\theta}\right)}{\partial x}\label{eq:derivative-scale}
\end{equation}
The benefit of scale normalization is that it eliminates the size effect on coordinate derivatives and maps the input to $\frac{\boldsymbol{x}}{L}$ (within $[-1,1]$), enhancing the fitting ability of the neural network. This is applicable not only in MLP but also in KAN, where scale normalization is even more essential than MLP. 

In this section, we mathematically explain why scale normalization is a critical trick in DEM.
Furthermore, \Cref{eq:u_x-scale} and \Cref{eq:u_y-scale} have another advantage: the linear scaling of the solution is directly related to the size. If the size of the PDE simulation increases by a factor of $L$, the solution will scale by the same factor. The involvement of the coordinate $x$ in \Cref{eq:u_x-scale} and \Cref{eq:u_y-scale} ensures that the solution scales appropriately with size, maintaining the scale relationship.

\section{The Challenge to Solve Inverse Problems in Deep Energy Methods\label{sec:DEM_difficult_inverse}}

We consider the DEM loss function for elastic mechanics problems:

\begin{equation}
	\begin{split}
		\mathcal{L}_{DEM} & =\int_{\Omega}\varPsi \, d\Omega - \int_{\Gamma^{\boldsymbol{t}}}\bar{\boldsymbol{t}}\cdot\boldsymbol{u} \, d\Gamma - \int_{\Omega}\boldsymbol{f}\cdot\boldsymbol{u} \, d\Omega \\
		\varPsi & = \frac{1}{2}\sigma_{ij}\varepsilon_{ij} \\
		\sigma_{ij} & = 2G\varepsilon_{ij} + \lambda\varepsilon_{kk}\delta_{ij} \\
		\varepsilon_{ij} & = \frac{1}{2}(u_{i,j} + u_{j,i}) \\
		G & = \frac{E}{2(1+\upsilon)} \\
		\lambda & = \frac{E\upsilon}{(1+\upsilon)(1-2\upsilon)} \\
		\text{s.t.} & \quad \boldsymbol{u} = \bar{\boldsymbol{u}}(\boldsymbol{x}), \, \boldsymbol{x} \in \Gamma^{\boldsymbol{u}}
	\end{split}
	,
\end{equation}
where $E$ and $\upsilon$ are the elastic modulus and Poisson's ratio of interest in the inverse problem. $\bar{\boldsymbol{u}}(\boldsymbol{x})$ represents the prescribed displacement field on the boundary $\Gamma^{\boldsymbol{u}}$.

Assuming that the displacement field $\boldsymbol{u}$, the force field $\boldsymbol{f}$, and the boundary force condition $\bar{\boldsymbol{t}}$ are known in advance, the mathematical form of the inverse problem is:

\begin{equation}
	\{E, \upsilon\} = \arg\min_{E, \upsilon} \mathcal{L}_{DEM} \label{eq:DEM_optimization}
\end{equation}

We now analyze the DEM loss function and expand it as follows:

\begin{align}
	\mathcal{L}_{DEM} & = \int_{\Omega} \frac{1}{2} [2G\varepsilon_{ij} + \lambda\varepsilon_{kk}\delta_{ij}] \varepsilon_{ij} \, d\Omega - \int_{\Gamma^{\boldsymbol{t}}} \bar{t}_{i}u_{i} \, d\Gamma - \int_{\Omega} f_{i}u_{i} \, d\Omega \label{eq:DEM_LOSS_inverse}
\end{align}
Considering that $\boldsymbol{u}$, $\boldsymbol{f}$, and $\bar{\boldsymbol{t}}$ are pre-fitted, the last two terms in \Cref{eq:DEM_LOSS_inverse} are constants and do not affect the optimization process. Thus,  \Cref{eq:DEM_optimization} is equivalent to:

\begin{equation}
	\begin{aligned}\{E,\upsilon\} & =\arg\min_{E,\upsilon}\{\int_{\Omega}\frac{1}{2}[2G\varepsilon_{ij}+\lambda\varepsilon_{kk}\delta_{ij}]\varepsilon_{ij}d\Omega\}\\
		& =\arg\min_{E,\upsilon}\{\int_{\Omega}(\varPsi_{V}+\varPsi_{S})d\Omega\}\\
		\varPsi_{V} & =\frac{1}{2}(\frac{2}{3}G+\lambda)\varepsilon_{kk}\varepsilon_{kk}\\
		\varPsi_{S} & =G\varepsilon_{ij}^{'}\varepsilon_{ij}^{'}\\
		\varepsilon_{ij}^{'} & =\varepsilon_{ij}-\frac{1}{3}\varepsilon_{kk}\delta_{ij}
	\end{aligned}
	\label{eq:DEM_optimization_reduced}
\end{equation}

Since the displacement field is pre-fitted, the strain $\varepsilon_{ij}$ is fixed during optimization. Given that the strain energy density $\varPsi \geq 0$, i.e., the strain energy consists of the volumetric strain energy $\varPsi_{V}$ and the shear strain energy $\varPsi_{S}$, both of which must be non-negative, the optimal solution is necessarily $E = 0$ and any Poisson's ratio $\upsilon$, which results in the minimum strain energy of zero.

However, it is evident that the elastic modulus in the inverse problem cannot be zero. Therefore, this analysis reveals that solving inverse problems using the DEM energy form involves a mathematical optimization error. The core issue is that the work of external forces is ignored in the optimization process of the variational principle. Fortunately, strong-form PINNs do not encounter this problem, which will not be proven here.

If we want to use DEM for inverse problems, we can adopt the idea from FEM instead of minimizing the energy functional directly as described in \Cref{eq:DEM_optimization_reduced}. For example, we can set the inverse variables \(E\) and \(\upsilon\) as initial guesses and use DEM to solve the PDEs to obtain the displacement. The computed and given displacements can be compared to measure the difference error. Based on the error, we can optimize the initial guesses for \(E\) and \(\upsilon\). This approach is very similar to the method used in topology optimization problems.

\section{Supplementary code}
The code of this work will be available at \url{https://github.com/yizheng-wang/Research-on-Solving-Partial-Differential-Equations-of-Solid-Mechanics-Based-on-PINN} after accepted.

\bibliographystyle{elsarticle-num}
\addcontentsline{toc}{section}{\refname}\bibliography{KINN.bib}

\end{document}